%% file: main_arxiv.tex
\documentclass[12pt]{article}
\usepackage[utf8]{inputenc}
\usepackage{booktabs}
\usepackage{array}
\usepackage{multirow}
\usepackage{multicol}

\usepackage[margin=1in]{geometry}

\usepackage[
    backend=biber,
    style=numeric-comp,
    sortcites,
    sorting=none,
    giveninits=true,
    natbib,
    hyperref=false,
    maxbibnames=99,
    doi=false,isbn=false,url=false,eprint=false
]{biblatex}

\usepackage[hidelinks]{hyperref}
\input{header.tex}

\newtheorem{theorem}{Theorem}
\newtheorem{corollary}{Corollary}
\newtheorem{proposition}{Proposition}
\newtheorem{assumption}{Assumption}
\newtheorem{lemma}{Lemma}

\newtheorem*{remark}{Remark}

\crefname{assumption}{assumption}{assumptions}
\Crefname{assumption}{Assumption}{Assumptions}

\usepackage{setspace}
\usepackage[colorinlistoftodos]{todonotes}

\title{Meta Fusion: A Unified Framework For Multimodality Fusion with Mutual Learning}
\author[1]{Ziyi Liang}
\author[2]{Annie Qu}
\author[1]{Babak Shahbaba\thanks{Corresponding author: \texttt{babaks@uci.edu}}}
\affil[1]{Department of Statistics, University of California, Irvine, CA, USA}
\affil[2]{Department of Statistics and Applied Probability, University of California, Santa Barbara, CA, USA}

\date{\today}

\begin{document}

\maketitle

\input{abstract}

\input{body}

\clearpage
\printbibliography 

\clearpage

\input{appendix}

\clearpage

\end{document}

%% file: header.tex
\usepackage{placeins}
\usepackage{tikz}
\usepackage{parskip}
\usepackage{graphicx, color, url}
\usepackage{xargs}
\usepackage[colorinlistoftodos,prependcaption,textsize=tiny]{todonotes}
\newcommandx{\note}[2][1=]{\todo[linecolor=blue,backgroundcolor=blue!25,bordercolor=blue,#1]{#2}}

\usepackage{authblk}

\usepackage{caption}
\usepackage{subcaption}
\usepackage{algorithm}
\usepackage{algorithmic}

\usepackage{xcolor}
\usepackage{graphicx,amsfonts,amsthm,amssymb}
\usepackage{bm}
\usepackage[hidelinks]{hyperref}
\usepackage{amsmath,mathrsfs}
\usepackage{mathtools}
\usepackage{multicol}
\usepackage{cleveref}

\newcommand{\argmin}{\mathop{\mathrm{argmin}}}
\newcommand{\argmax}{\mathop{\mathrm{argmax}}}

\newcommand{\mat}[1]{\bm{#1}}
\DeclarePairedDelimiter\norm{\lVert}{\rVert}
\DeclarePairedDelimiter\abs{\lvert}{\rvert}
\DeclarePairedDelimiter\paren{(}{)}
\DeclarePairedDelimiter\brac{[}{]}
\DeclarePairedDelimiter\cbrac{\{}{\}}

\newcommand{\Bigmid}{\mathrel{}\!\!\Big\vert\!\!\mathrel{}}
\newcommand{\bigmid}{\mathrel{}\!\!\big\vert\!\!\mathrel{}}

\def\E{\mathbb{E}}
\def\P{\mathbb{P}}
\def\I{\mathbb{I}}

\def\Var{\mathrm{Var}}

\def\cD{\mathcal{D}}
\def\cE{\mathcal{E}}

\def\cL{\mathcal{L}}

\def\cN{\mathcal{N}}
\def\cO{\mathcal{O}}
\def\cP{\mathcal{P}}
\def\cS{\mathcal{S}}

\def\cC{\mathcal{C}}

%% file: abstract.tex
\begin{abstract}
Developing effective multimodal data fusion strategies has become increasingly essential for improving the predictive power of statistical machine learning methods across a wide range of applications, from autonomous driving to medical diagnosis. Traditional fusion methods, including early, intermediate, and late fusion, integrate data at different stages, each offering distinct advantages and limitations. In this paper, we introduce \textbf{Meta Fusion}, a flexible and principled framework that unifies these existing strategies as special cases. Motivated by deep mutual learning and ensemble learning, Meta Fusion constructs a cohort of models based on various combinations of latent representations across modalities, and further boosts predictive performance through \emph{soft information sharing} within the cohort. Our approach is model-agnostic in learning the latent representations, allowing it to flexibly adapt to the unique characteristics of each modality. Theoretically, our soft information sharing mechanism reduces the generalization error. Empirically, Meta Fusion consistently outperforms conventional fusion strategies in extensive simulation studies. We further validate our approach on real-world applications, including Alzheimer's disease detection and neural decoding.

\vspace{0.5em}

\end{abstract}

%% file: body.tex
\section{Introduction}
Modern scientific research often involves processing and analyzing large amounts of data from diverse sources to extract information, make discoveries, and support decision-making. Effectively integrating these complex, multimodal data can lead to more holistic and rigorous solutions to scientific problems.
For example, autonomous driving combines multiple sensors for safe navigation~\citep{Rashed2019lidar&camera, Jelena_2018_sensor_fusion}. Sentiment analysis uses text, video, and audio to interpret emotions like humans do with multiple senses~\citep{bagher-zadeh-etal-2018-multimodal, Das_2023_sentiment_survery, gandhi-2023-multimodal-sentiment}. Medical diagnosis often integrates demographics, medical history, functional assessment, and imaging data~\citep{zhang-2011-ad-multimodal, Qiu-2022-multimodal-alzheimer}. 
Consequently, developing effective multimodal data fusion strategies has become crucial to enhancing our understanding of scientific mechanisms and improving the predictive accuracy of statistical machine learning models.

\subsection{Key questions: how to fuse, and what to fuse}\label{sec:two-questions}
Most existing research in multimodality fusion centers around two pivotal questions: how to fuse and what to fuse. The first addresses the encompassing fusion structure, while the second focuses on representation learning of multimodal data.

\begin{figure}[!t]
    \centering
    \includegraphics[width=\linewidth]{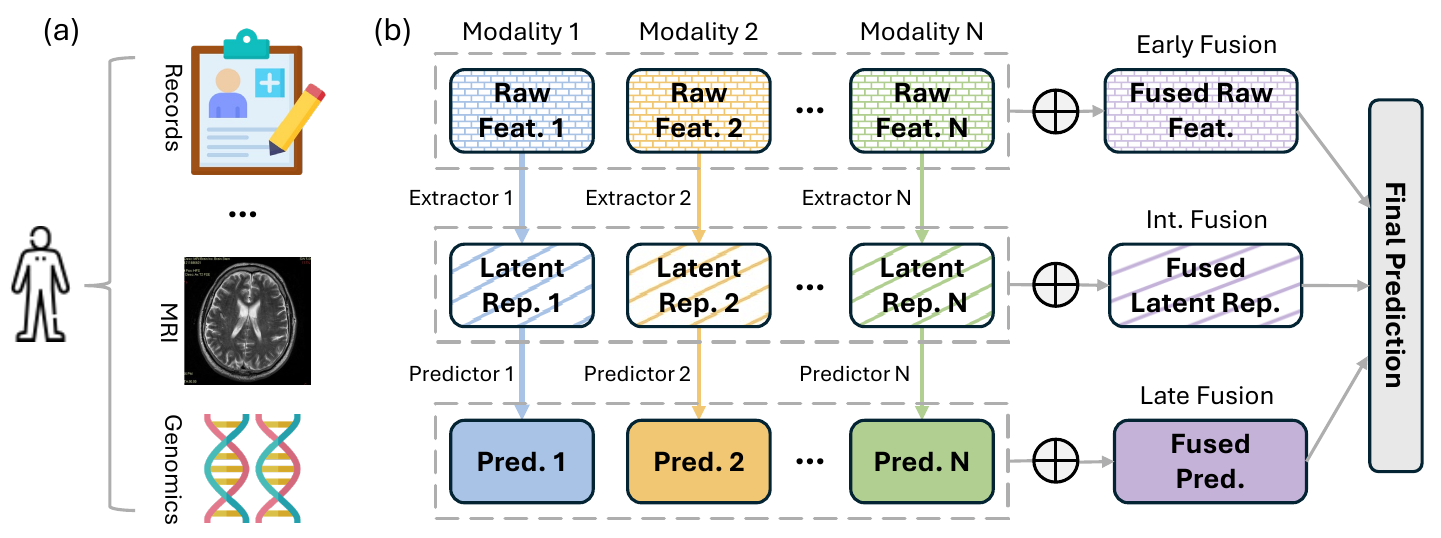}
    \caption{Overview of classical multimodal fusion strategies. (a) Example of multimodal data sources used in medical diagnosis, including patient records, MRI scans, and genomic data. (b) Illustration of the three broad fusion categories: early fusion, late fusion and intermediate fusion.}
    \label{fig:fusion-taxonomy}
\end{figure}

\paragraph{How to fuse?} Existing data fusion frameworks can be broadly categorized into early, late, and intermediate fusion (\Cref{fig:fusion-taxonomy}), based on the stage at which fusion occurs~\citep{Huang2020-fusion-review, zhang2021-imagefusion-survey}. Therefore, this question can also be framed as: ``When to fuse?'' \textit{Early fusion}~\citep{Couprie2013IndoorSS, yi2022-multimodal-autunomous} combines raw modalities before feature extraction, retaining all pertinent information and capturing rich cross-modal interactions. However, it is prone to overfitting in high-dimensional feature spaces and may perform poorly with highly heterogeneous and noisy modalities. \textit{Late fusion}~\citep{Qiu2018-latefusoin-alzheimer, Reda2018-latefusion-cancer, Yoo2019-latefusion-lesion}, or decision-level fusion, applies separate models to each modality and then aggregates their predictions. This approach is simple, robust, but suboptimal when the prediction task strongly depends on interactions or complementary information across modalities~\citep{Huang2020-fusion-review, ding2022-coop}. \textit{Intermediate fusion}~\citep{Yala2019-jointfusion-breastcancer, Nie2019-jointfusion-braintumor, Tan2022-jointfusion-socialmedia} finds a common ground by fusing latent representation extracted from each modality, but its effectiveness largely depends on the quality of the learned latent representations.

\paragraph{What to fuse?} A crucial step in data fusion is feature extraction, which involves learning low-dimensional latent representations to reduce noise and capture essential information from raw data. Early fusion seeks to learn a joint latent representation for all modalities simultaneously, as in multimodal Deep Boltzmann Machines~\citep{SrivastavaS2012_multimodal-DBM} or multimodal encoder-based methods~\citep{Wu2018-multimodalVAE, sutter2024-unity}. Intermediate fusion typically employs separate unimodal feature extractors. 
However, a key challenge lies in selecting appropriate latent dimensions for each modality since 
mis-specifying these dimensions can severely hinder the overall performance of the fusion process. 


\subsection{Main contribution}
Although many solutions have been proposed for specific applications~\citep{Yala2019-jointfusion-breastcancer, Nie2019-jointfusion-braintumor, yi2022-multimodal-autunomous}, comprehensive and reliable answers to these central questions remain elusive.
In this work, we aim to address this gap by proposing a data-driven framework, called \emph{Meta Fusion}, that automatically determines when to fuse and what to fuse.
Our approach starts with constructing a cohort of heterogeneous models, each focusing on different combinations of latent representations across modalities. These models are trained using a novel soft information sharing mechanism inspired by deep mutual learning~\citep{Zhang2017DeepML}. Rather than learning their assigned tasks independently, models in our framework are encouraged to align their outputs with those of the top performers in the cohort. Importantly, only outputs, rather than model parameters or latent representations, are shared during updates. This preserves cohort diversity, a reason we name it ``soft information sharing.'' Finally, we form a decision-making committee among the cohort through ensemble learning techniques~\citep{Breiman1996-bagging, caruana2004EnsembleSelection} and aggregate the committee’s predictions. 

Our key contributions include: First, we develop a task-agnostic and model-agnostic framework that provides a unified solution for multimodal fusion problems. Most existing benchmarks, as shown in \Cref{sec:relation}, can be viewed as special cases of this unifying framework. Second, we introduce a soft information sharing strategy through adaptive mutual learning that dynamically determines which information to share within the cohort. Specifically, we screen for top performers in the cohort and let others learn from them but not vise versa to avoid negative knowledge transfer. Our theoretical analysis shows that this strategy reduces individual model generalization error under appropriate conditions. To our knowledge, this is the first theoretical analysis of deep mutual learning and may be of independent interest in a broad context as it offers insights into the optimization landscape. Third, we demonstrate Meta Fusion's effectiveness through extensive numerical studies and applications to early detection of Alzheimer's disease and decoding of neuronal activities in the hippocampus.

\subsection{Organization}
The paper is organized as follows. \Cref{sec:related-work} reviews related work. \Cref{sec:methodology} presents Meta Fusion, with methodological details in \Cref{sec:main-method} and connections to existing benchmarks in \Cref{sec:relation} (with additional details in \Cref{app:connection-to-coop}). \Cref{sec:theory} provides the theoretical results. \Cref{sec:synthetic-experiments} and \Cref{sec:real-experiments} present simulation experiments and real-world data analysis, respectively. \Cref{sec:discussion} concludes with a discussion and suggestions for future directions. Implementation details and additional numerical experiments are provided in \Cref{app:implementation} and \Cref{app:additional-num-exp} respectively. \Cref{app:proofs} contains all mathematical proofs and \Cref{app:supp-tabs-figs} includes supplementary tables and figures.

\section{Related work}\label{sec:related-work}
\subsection{Cooperative learning for multiview analysis}\label{sec:related-work:coop}
Among recent fusion strategies, the cooperative learning approach proposed by \citet{ding2022-coop} is particularly relevant to our work, as it also aims to develop a framework that automatically determines when to fuse. Consider two modalities as random variables $X \in \mathbb{R}^{p_x}$ and $Z \in \mathbb{R}^{p_z}$ with ground truth label $Y\in \mathbb{R}$, where $p_x,p_z \in \mathbb{Z}$ are feature dimensions. Cooperative learning minimizes the following population quantity for a given hyperparameter $\rho \in [0,1)$:
\begin{equation}\label{eq:coop-objective}
\begin{aligned}
    \min_{f_X, f_Z, f_{XZ}} \mathbb{E}\Big\{&\frac{1}{2}\paren*{Y-f_{X}(X)-f_{Z}(Z)-f_{XZ}(X, Z)}^2 \\
    &+ \frac{\rho}{2}\paren*{f_{X}(X)-f_{Z}(Z)}^2 + \frac{\rho}{2(1-\rho)}f^2_{XZ}(X, Z)\Big\}.
\end{aligned}
\end{equation}
The expectation is taken over the randomness of $X$ and $Z$. Here, $f_{X}$, $f_{Z}$ are unimodal predictors, and $f_{XZ}$ is a joint predictor capturing feature interactions. The first term in \eqref{eq:coop-objective} is the mean squared error; the second is a penalty encouraging agreement between modalities; and the third controls the contribution of the interaction term. \citet{ding2022-coop} show that~\eqref{eq:coop-objective} admits fixed points:
\begin{equation}\label{eq:coop-solution}
\begin{aligned}    f_{X}(X) &= \mathbb{E}\cbrac*{\frac{Y}{1+\rho} - \frac{(1-\rho)f_{Z}(Z)}{(1+\rho)} - \frac{f_{XZ}(X, Z)}{1+\rho} \mid X},\\
    f_{Z}(Z) &= \mathbb{E}\cbrac*{\frac{Y}{1+\rho} - \frac{(1-\rho)f_{X}(X)}{(1+\rho)} - \frac{f_{XZ}(X, Z)}{1+\rho} \mid X},\\
    f_{XZ}(X, Z) &= \mathbb{E}\cbrac[\Big]{(1-\rho)(Y - f_{X}(X)-f_{Z}(Z)) \mid X, Z}.
\end{aligned}
\end{equation}
By adjusting
$\rho$, cooperative learning can interpolate between early and late fusion. When $\rho=0$, \eqref{eq:coop-objective} reduces to the additive model 
$f_{X}(X) + f_{Z}(Z) + f_{XZ}(X, Z)$, equivalent to early fusion if the functions are additive. As $\rho \rightarrow 1$, it is clear from \eqref{eq:coop-solution} that the joint function $ f_{XZ}$ vanishes, and the solution converges to $0.5\mathbb{E}(Y\mid X)+0.5\mathbb{E}(Y\mid Z)$, corresponding to late fusion via averaging the marginal predictions.
In practice, $\rho$ is selected through cross-validation, offering a data-driven way to determine when to fuse modalities. However, cooperative learning has limitations. First, it does not fully address what to fuse: the joint function $f_{XZ}$ is trained on the combined features of both modalities, which can lead to poor performance in high-dimensional or heterogeneous settings. While \citet{ding2022-coop} recommend using Lasso~\citep{tibshirani1996-lasso} for feature selection in linear settings, it remains unclear how to effectively reduce dimensionality for complex models in nonlinear contexts. This issue also affects the unimodal predictors. Since the final prediction is the sum of $f_{X},f_{Z} \text{, and } f_{XZ}$, overfitting in any component may degrade overall performance. Second, the framework is tailored to regression tasks using mean squared error, and its adaptation to classification remains an open problem.

\subsection{Deep mutual learning}\label{sec:related-work:mutual-learning}
Deep mutual learning extends knowledge distillation~\citep{bucilua2006model-compression, Hinton2015DistillingTK} to a cohort of \emph{student} models that learn collaboratively by sharing information with one another. In addition to optimizing their task-specific losses, these models also align their outputs during training. 
The original mutual learning framework~\citep{Zhang2017DeepML} is designed for single-modal data and homogeneous students, where models with identical architectures are trained in parallel with different initializations.
More recently, deep mutual learning has been extended to multimodal fusion~\citep{zhang-2021-ml-image-segmentation, wang-2022-amnet, zhang2023-contrastive-ml-image, li-2024-ml-multimodal-recommender}, typically assigning one model per modality and applying standard mutual learning, which treats all students equally. This approach might be suboptimal in the presence of noisy or heterogeneous modalities, as demonstrated by our theoretical and experimental findings (\Cref{sec:theory} and \Cref{app:comp-div-weights}). In contrast, Meta Fusion introduces adaptive mutual learning to selectively determine which models to learn from and effectively mitigate negative knowledge transfer.

\section{Methodology}\label{sec:methodology}

\subsection{Meta Fusion}\label{sec:main-method}

\begin{figure}[!t]
    \centering
    \includegraphics[width=\textwidth]{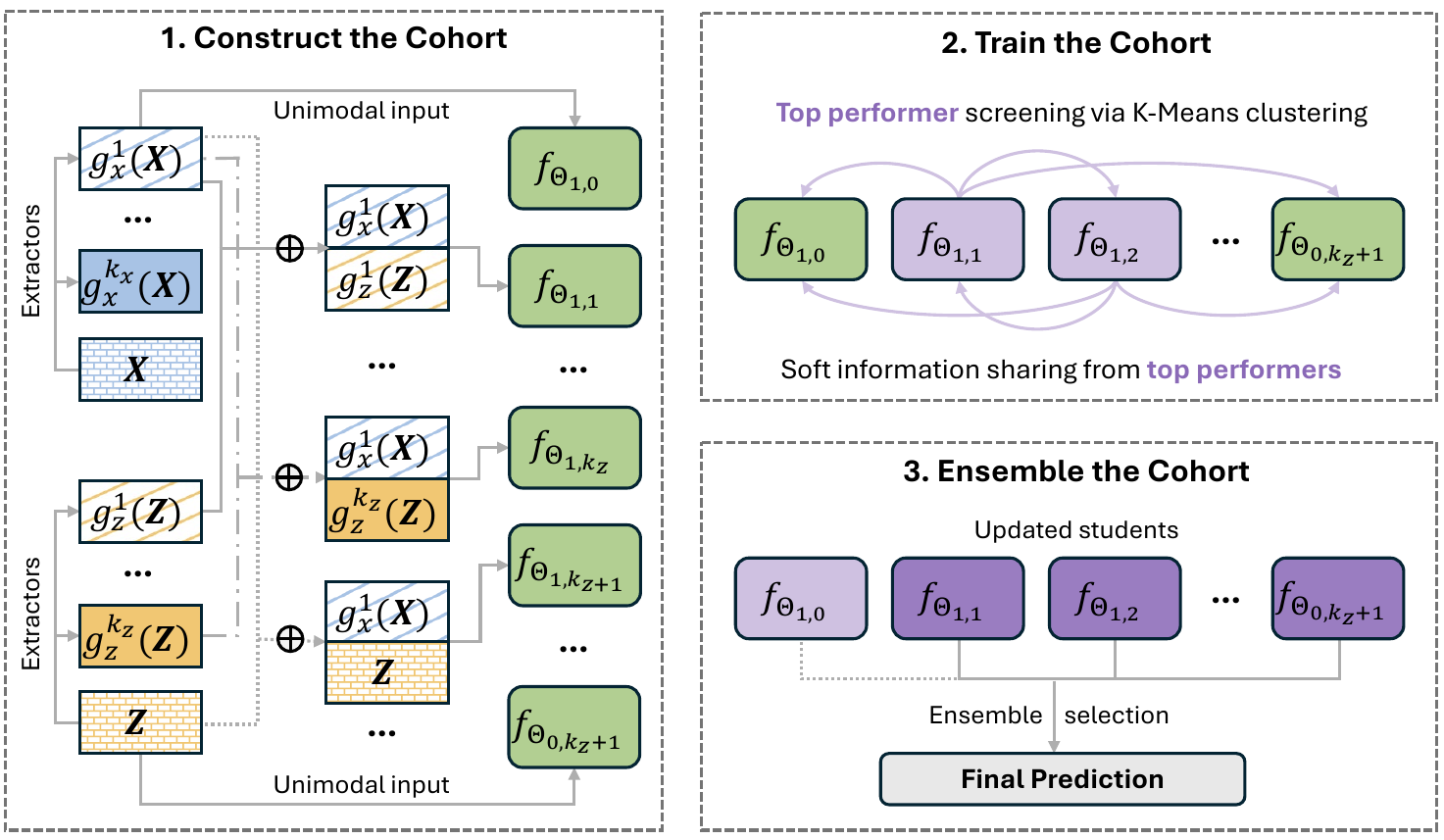}
    \caption{Overview of the Meta Fusion pipeline.}
    \label{fig:meta-fusion-overview}
\end{figure}

To address the above issues, we propose Meta Fusion, a novel mutual learning framework for multimodal data fusion that flexibly extracts and integrates information from diverse data sources. As illustrated in \Cref{fig:meta-fusion-overview}, Meta Fusion consists of \textit{three main steps}. First, we construct a cohort of student models (\Cref{sec:build-cohort}), each receiving different combinations of latent representations from the available modalities. These student models may have varying architectures to accommodate the heterogeneity of the input data. Second, as described in \Cref{sec:training}, the cohort undergoes adaptive mutual learning to facilitate soft information sharing. We use a data-driven approach to identify the top-performing student models and allow the others to learn from them, thereby reducing the risk of negative knowledge transfer. This approach contrasts with conventional mutual learning~\citep{Zhang2017DeepML}, which treats all student models equally. Finally, we aggregate the predictions of the cohort using ensemble techniques, such as ensemble selection (\Cref{sec:ensemble}), to produce the final output.

\subsubsection{Constructing the student cohort}\label{sec:build-cohort}

Rather than relying on a single model to capture and integrate information from all modalities, we construct a set of models called ``students". We refer to their collection as a ``student cohort"~\citep{Zhang2017DeepML}. Each student can have a different structure or focus on different modalities to promote diversity of feature learning. 
Given samples from two modalities, $\mat{X} \in \mathbb{R}^{n\times p_x}$, $\mat{Z} \in \mathbb{R}^{n\times p_z}$, where $n \in \mathbb{Z}$ is the sample size, we use $k_x > 1$ feature extractors for $\mat{X}$, denoted as $g^i_x(\mat{X})$ for $i \in [k_x]$, and $k_z>1$ extractors for $\mat{Z}$, denoted as $g^j_z(\mat{Z})$ for $j \in [k_z]$. Extractors are tailored to each modality. For example, convolutional networks or pre-trained models like ResNet~\citep{He2015DeepRL} for images, and BERT~\citep{Devlin2019BERTPO} for text. Different extractors, or outputs from various layers, yield diverse latent representations, capturing information at multiple resolution levels. We also introduce two dummy extractors for each modality: an identity mapping, $g^{k_x+1}_x(\mat{X}) \coloneqq \mat{X}$ and $g^{k_z+1}_x(\mat{Z}) \coloneqq \mat{Z}$, which preserves all raw features, and a null mapping, $g^{0}_x(\mat{X}) \coloneqq \emptyset$ and $g^{0}_x(\mat{X}) \coloneqq \emptyset$, which excludes the corresponding modality.

Given the extractors $\cbrac{g^i_x(\mat{X})}_{i \in \cbrac{0, \dots, k_{x}+1}}$ and $\cbrac{g^j_z(\mat{Z})}_{j \in \cbrac{0, \dots, k_{z}+1}}$, we fuse their outputs to construct student models by using a cross-modality pairing strategy, illustrated in the left panel of \Cref{fig:meta-fusion-overview}. That is, each latent representation of $\mat{X}$, $g^i_x(\mat{X})$ for $i \in \cbrac{0, 1, \dots, k_{x}+1}$, is paired with a latent representation of $\mat{Z}$, $g^i_z(\mat{Z})$ for $j \in \cbrac{0, 1, \dots,k_{z}+1}$ to form a student model. Pairings are combined as $g^i_x(\mat{X}) \oplus g^j_z(\mat{Z})$, where $\oplus$ denotes column-wise concatenation. We exclude the null combination $g^0_x(\mat{X}) \oplus g^0_z(\mat{Z})$, and define the set of valid pairings as $ \cP \coloneqq \cbrac{ \cbrac{0,\dots, k_{x}+1}\times \cbrac{0, \dots, k_{z}+1} }\setminus \cbrac{(0,0)}$. For each $(i,j) \in \cP$, the fused representation is used as the input to train a supervised model: 
\begin{align}\label{eq:student-model}
    f_{\Theta_{i,j}}(\mat{X}, \mat{Z}) \coloneqq f_{i,j}(g^i_x(\mat{X}) \oplus g^j_z(\mat{Z})),
\end{align}
where $f_{i,j}$ is the task-specific model, possibly varying in structures, and $\Theta_{i,j}$ denotes trainable parameters. This yields a cohort of $(k_x+2)(k_z+2)-1$ students: $k_x+1$ models use only $\mat{X}$ (i.e., $f_{\Theta_{i,0}}$ for $i \in [k_x+1]$), $k_z+1$ use only $\mat{Z}$ (i.e.,$f_{\Theta_{0,j}}$ for $j \in [k_z+1]$), and $(k_x+1)(k_z+1)$ use both modalities, with $f_{\Theta_{k_x+1,k_z+1}}$ retrieving the early fusion model that operates on concatenation of raw modalities. In practice, cohort size is flexible and can be adjusted for computational efficiency. This comprehensive pairing facilitates data-driven selection of optimal latent combinations.

\subsubsection{Training the student cohort}\label{sec:training}
Although model diversity can improve ensemble performance, recent studies have shown that optimal results depend on balancing diversity and predictive performance \citep{wood2023diversity, kumar2023fsp-ensemble}; maximizing diversity alone is not always ideal. In multimodal fusion, cross-modality pairing naturally increases diversity, but not all latent representations are equally beneficial for downstream tasks. To this end, we introduce a mutual learning framework that adaptively selects top-performing students and enables the cohort to learn from them, enhancing overall predictive performance through merit-guided information sharing across diverse ensemble representations.

Let $\mat{Y} \in \mathbb{R}^n$ denote the ground-truth labels and $\widehat{\mat{Y}}_{I}=f_{\Theta_{I}}(\mat{X}, \mat{Z})\in \mathbb{R}^{n\times d}$ denote the prediction from student model $f_{\Theta_{I}}$, where $d$ is the output dimension and $I \in \cP$ indexes a specific cross-modal pairing of latent features. Each student's loss consists of two components: a task-specific prediction loss, $\cL(\widehat{\mat{Y}}_I, \mat{Y})$, and a divergence loss that encourages agreement with other cohort members:
\begin{align}\label{eq:overall-loss}
    \cL_{\Theta_{I}} = \underbrace{\cL(\widehat{\mat{Y}}_I, \mat{Y})}_{\text{task loss}}+\rho \underbrace{\sum\nolimits_{J\in \cP, J\neq I}d_{I,J} \cD(\widehat{\mat{Y}}_I, \widehat{\mat{Y}}_J)}_{\text{divergence loss}},
\end{align}
where $\cD(\widehat{\mat{Y}}_I, \widehat{\mat{Y}}_J)$ measures the divergence between students $I$ and $J$. The objective in \eqref{eq:overall-loss} is task-agnostic: for regression, Mean Squared Error (MSE) can be used for both the task and divergence terms, while for classification, Cross-Entropy Loss is suitable for the task loss and Kullback-Leibler (KL) divergence for the second term.

The hyperparameter $\rho>0$, determined via cross-validation, controls the balance between the two loss terms. The weights $d_{I,J} \geq 0$ determine the influence of each peer in the divergence loss. In standard mutual learning, $\rho=1$ and $d_{I,J}=1$ for all $I,J\in \cP, J\neq I$, so each student learns from all others—an approach that is more suitable for homogeneous, single-modal data. However, in the multimodal case, students may be highly heterogeneous due to differing modality combinations or architectures, resulting in varying performance across the cohort. If  student $I$ performs significantly worse than others, it can be advantageous to set its associated $d_{J,I}$ for $J \in \cP, J\neq I$ 
to zero or a small value to avoid negative knowledge transfer. To address this, we propose a two-step adaptive mutual learning procedure: first, determine the divergence weights $d_{I,J}$ in a data-driven way, then train the cohort using these optimized weights.

\textbf{Step 1: Initial Screening}

To determine appropriate weights $\cbrac{d_{I,J}}_{I,J\in \cP, I\neq J}$ and prevent negative learning, models are first trained independently using only the task loss, namely, $\rho=0$ in \eqref{eq:overall-loss}. After this phase, each student’s initial loss, $\cbrac{L^{\mathrm{init}}_{\Theta_I}}_{I \in \cP}$ is evaluated on a holdout validation set $\cD^{\mathrm{val}} \subset \cbrac{\mat{X}, \mat{Z}, \mat{Y}}$. Next, we apply K-Means clustering to the set of initial losses $\cbrac{L^{\mathrm{init}}_{\Theta_I}}_{I \in \cP}$ to group students into $k_{\mathrm{cls}}>1$ performance-based clusters. The value of $k_{\mathrm{cls}}$ can be chosen automatically using approaches such as the Silhouette method~\citep{rousseeuw1987silhouettes} or the Elbow method~\citep{marutho2018elbow}. Let $\cS=\cbrac{S_{1}, \dots, S_{k_{\mathrm{cls}}}}$, where $\cup_{k\in[k_{\mathrm{cls}}] }S_{k} = \cP$, denote the resulting partition of the model indices $\cP$. 

We define $\cS_{\mathrm{top}}$ as the set of models belonging to the best-performing $k_{\mathrm{top}}\geq 1$ clusters with the lowest average initial loss. The divergence weights are then set according to cluster membership:
\begin{align}\label{eq:divergence-weights}
    d_{I,J} =
    \begin{cases} 
    1, & \text{if } J \in \cS_{\mathrm{top}} \\
    0, & \text{otherwise}.
    \end{cases}
\end{align}
That is, model $I$ learns from model $J$ only if $J$ is identified as a top performer in the initial screening. This strategy prioritizes students with stronger initial performance, potentially due to superior architectures or informative cross-modal features. By adaptively setting the divergence weights, we enable weaker models to benefit from the best performers, while protecting top students from negative influence during mutual learning.

\textbf{Step 2: Soft Information Sharing via Adaptive Mutual Learning}

With the divergence weights $\cbrac{d_{I,J}}_{I,J \in \cP, I\neq J}$ determined and a fixed hyperparameter $\rho >0$ (selected via cross-validation), the student cohort is jointly trained for $n_t >0$ epochs by minimizing the overall loss in \eqref{eq:overall-loss} using gradient descent. Training can be performed in parallel for efficiency. With $\rho >0$, student models are guided to align with top-performing peers during each update, promoting learning through soft information sharing. This process preserves diversity (since model parameters are not exchanged directly) while encouraging consensus by aligning model outputs, thus improving predictive performance and maintaining beneficial differences among students. Overall, this mutual learning step helps balance the trade-off between diversity and predictive performance \citep{wood2023diversity}. We further demonstrate theoretically in \Cref{sec:theory} that, under suitable conditions, this adaptive mutual learning strategy effectively reduces the generalization error of individual students.
The two step training procedure of the student cohort is summarized in \Cref{alg:training-cohort}.

\begin{algorithm}[!t]
\caption{Training with Adaptive Mutual Learning}\label{alg:training-cohort}
\begin{algorithmic}[1]
\STATE \textbf{Input:} Student models $f_{\Theta_I}$ for $I \in \cP$; multimodal dataset $(\mat{X}, \mat{Z})$; label set $\mat{Y}$; epochs $n_t$; and mutual learning parameter $\rho^* >0$.
\STATE Randomly initialize student parameters $\Theta_I$ for all $I \in \cP$.
\STATE Randomly split the multimodal data into two disjoint subsets $\cD^{\mathrm{train}}$ and $\cD^{\mathrm{val}}$.
\STATE \textbf{Step 1: Initial Screening}
\STATE \quad Train $\Theta_I$ for all $I \in \cP$ for $n_t$ epochs by minimizing the loss in \eqref{eq:overall-loss} with $\rho=0$.
\STATE \quad Evaluate the initial task loss $\cbrac{L^{\mathrm{init}}_{\Theta_I}}_{I \in \cP}$ on the holdout data $\cD^{\mathrm{val}}$.
\STATE \quad Compute the divergence weights $\cbrac{d_{I,J}}_{I,J \in \cP, I\neq J}$ with \Cref{eq:divergence-weights}.
\STATE \textbf{Step 2: Mutual Learning}
\STATE \quad Re-initialize student parameters $\Theta_I$ for all $I \in \cP$.
\STATE \quad Train $\Theta_I$ for all $I \in \cP$ for $n_t$ epochs by minimizing the loss in \eqref{eq:overall-loss} with $\rho=\rho^*$\\
\quad and $\cbrac{d_{I,J}}_{I,J \in \cP, I\neq J}$ obtained in \textbf{Step 1}.
\STATE \textbf{Output:} Updated student cohort $f_{\Theta_I}$ for $I \in \cP$.
\end{algorithmic}
\end{algorithm}

\subsubsection{Aggregating the student cohort}\label{sec:ensemble}
After training the student cohort, we aggregate their predictions using ensemble techniques to obtain a final, unified prediction. Any standard ensemble method appropriate for the task may be used, e.g., stacking, or averaging (see \Cref{app:ensemble-methods} for an overview). 
Given the heterogeneous nature of the cohort, we adopt \textit{ensemble selection}~\citep{caruana2004EnsembleSelection, caruana2006EnsembleSelection} to form a robust decision-making committee $\cC \subseteq \cP$. This method builds the ensemble by iteratively adding models that improve validation performance, optimizing for a chosen metric. To enhance efficiency and robustness, we first rank the models by task-specific loss on a holdout set $\cD^{\mathrm{val}}$, and prune the lowest-performing $p_{\mathrm{prune}} \in (0,1)$ fraction. We initialize $\cC$ with $n_{\mathrm{init}}>0$ top performers, then iteratively add models from the remaining candidates that yield the greatest improvement in ensemble performance, which may be evaluated via simple or weighted averaging. The detailed procedure is outlined in \Cref{alg:ensemble-selection}.

Ensemble selection maximizes performance by forming a robust committee using the most effective models, avoiding the inclusion of weaker models that could reduce overall quality. While both initial K-Means screening (\Cref{alg:training-cohort}) and ensemble selection (\Cref{alg:ensemble-selection}) identify strong performers, the former quickly filters students for mutual learning robustness, while the latter optimizes the final committee for ensemble performance. Notably, students who do not excel initially may improve and be selected for the final ensemble.

\begin{algorithm}[!t]
\caption{Forming Decision-making Committee with Ensemble Selection}\label{alg:ensemble-selection}
\begin{algorithmic}[1]
\STATE \textbf{Input:} Student models $f_{\Theta_I}$ for $I \in \cP$; validation set $\cD^{\mathrm{val}}$; pruning proportion $p_{\mathrm{prune}}$; number of initial models $n_{\mathrm{init}}>0$; maximum committee size $n_c > 0$; ensemble loss function $\cL^{\mathrm{ens}}(\cC, \cD)$ for committee $\cC$ on data $\cD$.
\STATE Evaluate and sort the task-specific loss of each $f_{\Theta_I}$ for $I \in \cP$ using $\cD^{\mathrm{val}}$.
\STATE Prune the lowest-performing $p_{\mathrm{prune}}$ fraction of students.
\STATE Initialize the committee $\cC$ with $n_{\mathrm{init}}$ best-performing students.
\STATE Set $\cP_{\mathrm{avail}}\subset \cP$ to all remaining students after pruning and initialization.
\WHILE{$\abs{\cC} < n_c$}
\STATE For each $I \in \cP_{\mathrm{avail}}$, evaluate $\cL^{\mathrm{ens}}(\cC \cup \{I\}, \cD^{\mathrm{val}})$.
\STATE Select $I^* = \arg\min_{I \in \cP_{\mathrm{avail}}} \cL^{\mathrm{ens}}(\cC \cup \{I\}, \cD^{\mathrm{val}})$.
\IF{$\cL^{\mathrm{ens}}(\cC \cup \{I^*\}, \cD^{\mathrm{val}}) < \cL^{\mathrm{ens}}(\cC, \cD^{\mathrm{val}})$}
    \STATE Add $I^*$ to committee: $\cC \gets \cC \cup \{I^*\}$.
    \STATE Remove $I^*$ from $\cP_{\mathrm{avail}}$: $\cP_{\mathrm{avail}} \gets \cP_{\mathrm{avail}} \setminus \{I^*\}$.
\ELSE \STATE \textbf{break}
\ENDIF
\ENDWHILE
\STATE \textbf{Output:} The decision-making committee $\cC$.
\end{algorithmic}
\end{algorithm}

\subsection{Relation to existing methods}\label{sec:relation}
Many widely used methods can be viewed as special cases of our proposed unified approach, offering insight into the superior performance observed in our experiments (\Crefrange{sec:synthetic-experiments}{sec:real-experiments}).
Classical paradigms of multimodal fusion, namely early, intermediate, and late fusion, are all encompassed within our unified framework. Specifically, our model reduces to early fusion when there is a single student, $f_{\Theta_{k_x+1, k_z+1}}(\mat{X}, \mat{Z})$, which operates on the concatenation of raw modalities. Intermediate fusion is similarly recovered when only one student $f_{\Theta_{i, j}}(\mat{X}, \mat{Z})$ with arbitrary feature extractors, $g^i_x(\mat{X})$ and $g^j_z(\mat{Z})$ for any $i \in [k_x], j \in [k_z]$, is considered. Late fusion fits within our framework as a cohort of two students $f_{\Theta_{k_x+1, 0}}(\mat{X}, \mat{Z})$, operating only on $\mat{X}$ and $f_{\Theta_{0, k_z+1}}(\mat{X}, \mat{Z})$ operating only on $\mat{Z}$, where both models are trained independently without mutual learning.

Cooperative learning can also be viewed as a special case, albeit in a more nuanced way. \Cref{app:connection-to-coop} analyzes a simplified form of the cooperative learning objective in \eqref{eq:coop-objective} that excludes explicit cross-modal interactions. The simplified objective is the primary focus in \citet{ding2022-coop} and \Cref{app:connection-to-coop} shows that it is equivalent to Meta Fusion using two single-modality students aggregated with simple averaging. 

Finally, several existing multimodal fusion methods employing deep mutual learning techniques \citep{zhang-2021-ml-image-segmentation, wang-2022-amnet, zhang2023-contrastive-ml-image, li-2024-ml-multimodal-recommender} can also be seen as special cases of our approach. In these cases, unimodal students are trained using non-adaptive mutual learning, with parameters $\rho=1$ and $d_{I,J}=1$ for any student pair $(I,J)$. This approach does not facilitate interactions between cross-modal features, which may lead to poor performance when modalities provide complementary information. Additionally, non-adaptive mutual learning can lead to degraded performance in the presence of noisy or adversarial modalities. 

\subsection{Extension to multiple modalities}\label{sec:extension-multiple-modalities}
Our framework readily extends to settings with more than two modalities. The main consideration is constructing the cross-modal student cohort; the processes for cohort training via adaptive mutual learning (\Cref{sec:training}) and final prediction through ensemble selection (\Cref{sec:ensemble}) remain unchanged. Appendix~\ref{app:extension-multiple-modalities} describes how to generalize the cross-modal pairing strategy in \Cref{sec:build-cohort} to build student cohorts for applications involving more than two modalities.

\section{Theoretical properties}\label{sec:theory}

\subsection{Notations}\label{sec:notations}
By convention, we use boldface to denote vectors or matrices (e.g. $\mat{0}$ for vector of zeros and 0 for scalar.) Additionally, we define the following technical notations:
For a matrix $\mat{M}$, $\mat{M}_{i:}$ denotes the $i$-th row and $\mat{M}_{:i}$ denotes the $i$-th column. For matries $\mat{M}_{I}$,$\mat{M}_{J}$ indxed by $I, J\in \cP$, we denote $\mat{M}^{\top}_I\mat{M}_J$ as $\mat{M}^{\mathstrut}_{IJ}$ for simplicity.  For $\mat{x} \in \mathbb{R}^d$, $\mat{x}^{\circ 2} \coloneqq (x_1^2, \dots, x_d^2)$ denotes the element-wise square, and $\Var^{\circ}(\mat{x}) \coloneqq (\Var(x_1), \dots, \Var(x_d))$ denotes the element-wise variance.

\subsection{Main theories}
In this section, we present the theoretical properties of the proposed Meta Fusion framework. Although Meta Fusion is both task-agnostic and model-agnostic by design, for clarity and tractability, we focus our theoretical analysis on regression tasks using mean squared error (MSE) as the loss function.

We assume the ground-truth labels $\mat{Y}=(Y_1, \dots, Y_n) \in \mathbb{R}^n$ are generated from a latent factor model, $\mat{Y} = \mat{V}\mat{\theta}$ for some coefficients $\mat{\theta} \in \mathbb{R}^p$ and latent factors $\mat{V} \in \mathbb{R}^{n \times p}$. Assume $\mat{V}_{i:}$ follows a multivariate Gaussian distribution, independently for each $i \in [n]$ with
\begin{align}\label{eq:latent-ground-truth}
    \mat{V}_{i:} \sim N(\mat{0}, \mat{\Sigma}),
\end{align}
where $\mat{\Sigma} \in \mathbb{R}^{p\times p}$ is the identity matrix.

Consider the general case with $M\geq 2$ modalities $\mat{X}_m$. For any student model $I = (i_1, \dots, i_M)\in \cP$, denote the fused presentation as $\mat{V}_{I} \coloneqq \bigoplus_{m=1}^M g^{i_m}_m(\mat{X}_{m})$, where each $g^{i_m}_m$ is a feature extractor for modality $\mat{X}_m$ (see \Cref{app:extension-multiple-modalities}). In our theoretical analysis, we assume a signal-plus-noise model for the fused representations where each $\mat{V}_{I}$ is modeled as a noisy transformation of the oracle latent factors $\mat{V}$.
\begin{assumption}\label{cond:related-latent-rep}
    For a student model $I \in \cP$, the fused representation satisfies
    \begin{align}\label{eq:latent-fused-representation}
        \mat{V}_{I} = \mat{V}\mat{T}_{I} + \mat{\epsilon}_{I},
    \end{align}
    where $ p_{I}$ is the feature dimension of $\mat{V}_{I}$ and $\mat{T}_{I}\in \mathbb{R}^{p\times p_{I}}$ is a linear transformation, $\mat{\epsilon}_{Ii:} \sim N(0,\sigma_{I}^2\mat{\Sigma}_{I})$ with $\sigma_{I} >0$, and $\mat{\Sigma}_{I} \in \mathbb{R}^{p_{I} \times p_{I}}$ is the identity matrix. 
\end{assumption}

We assume the features of the fused latent representations are independent. This assumption is standard in the literature, as it enables analytical simplifications~\citep{murphy-2012-ml-probabilistic}, and preprocessing steps such as PCA are often employed to promote feature independence.
\begin{assumption}\label{cond:orthogonal}
    For $I \in \cP$, we assume $\mat{T}_{I}\in \mathbb{R}^{p\times p_{I}}$ has orthogonal columns: $\mat{T}^{\top}_{I:i}\mat{T}_{I:j}=0.$
\end{assumption}
We analyze our method with a cohort of two students, i.e., $\cP=\cbrac{I,J}$ with $d_{I,J}=1$. Each student’s supervised model is a deep linear network, a neural network without activation functions. Although deep linear networks do not have greater capacity than simple linear regressors, they are frequently used as a foundation for understanding the behavior of deeper, nonlinear networks~\citep{Saxe2014ExactST, kawaguchi-poorlocalmin-NIPS2016, hardt2017identity, Phuong2019TowardsUK}.

Let $\mat{\theta}_I \in \mathbb{R}^{p_I}$ and $\mat{\theta}_J \in \mathbb{R}^{p_J}$ be the model parameters for student $I,J$, respectively. One can show that optimizing the individual objectives in~\eqref{eq:overall-loss}
is equivalent to minimizing the following combined loss, since both yield the same gradients:
\begin{equation}\label{eq:mse-overall-loss}
    \cL_{(\mat{\theta}_I,\mat{\theta}_J)} = \norm*{\mat{Y}-\mat{V}_{I}\mat{\theta}_I}^2 + \norm*{\mat{Y}-\mat{V}_{J}\mat{\theta}_J}^2 + \rho\norm*{\mat{V}_{I}\mat{\theta}_I - \mat{V}_{J}\mat{\theta}_J}^2.
\end{equation}
The parameters are estimated by minimizing the empirical objective above, namely,
\begin{equation}\label{eq:fitted-parameters}
\begin{aligned}
    (\hat{\mat{\theta}}_{I},\hat{\mat{\theta}}_{J}) = \argmin_{\mat{\theta}_I \in \mathbb{R}^{p_I},\mat{\theta}_J \in \mathbb{R}^{p_J}} \cL_{(\mat{\theta}_I,\mat{\theta}_J)}.
\end{aligned}
\end{equation}
Our first theoretical result shows that the global minimizer $\hat{\mat{\theta}}_{I}$ depends on the disagreement penalty $\rho$. Notably, $\hat{\mat{\theta}}_{I}$ is also a function of $\mat{V}_{J}$, indicating that student $J$ effectively shares information with $I$ through mutual learning, without directly disclosing its latent representations. Since students $I$ and $J$ play symmetric roles, all subsequent analysis for $I$ applies equally to $J$ by swapping indices.
\begin{proposition}\label{prop:convergence}
Assume both students in $\cP$ satisfy \Cref{cond:related-latent-rep}, and $n\geq \max \cbrac{p_I, p_J}$. The global minimizer $\hat{\mat{\theta}}_I$ of the objective function in \eqref{eq:mse-overall-loss} takes the form of:
\begin{align*}
    \hat{\mat{\theta}}_I = (1-\rho)\mat{V}^{-1}_{II}\mat{V}_I^{\top} \mat{Y} + \rho\mat{V}_{II}^{-1}\mat{V}^{\mathstrut}_{IJ}\mat{V}_{JJ}^{-1}\mat{V}_J^{\top} \mat{Y}+\mat{H}(\rho),
\end{align*}
where $\mat{H}(\rho)=\sum_{n=2}^{\infty}\rho^n \mat{G}_n$, for some $\mat{G}_n \in \mathbb{R}^{p_I}$ independent of $\rho$. 
\end{proposition}

Next, we investigate how the choice of $\rho$ influences the generalization error on an unseen \textit{test point}. Let $\mat{V}_* \in \mathbb{R}^p$ denote the oracle latent vector for the test point, independently drawn according to~\eqref{eq:latent-ground-truth}, and let $Y_*=\mat{V}_*^{\top}\mat{\theta}$ be the ground-truth label. The fused representations for this test point, $\mat{V}_{I*}, \mat{V}_{J*}$, are defined as in~\eqref{eq:latent-fused-representation}.  The generalization error for student $I$ is given by:
\begin{align}\label{eq:generalization-error}
    \text{MSE}(I;\rho) = \E\cbrac*{(Y_*-\mat{V}^{\top}_{I*} \hat{\mat{\theta}}_{I})^2 \mid \mat{V}_I}.
\end{align}
We introduce several key quantities that will later allow us to decompose the generalization error for a clearer understanding. Let $\Tilde{\mat{\Sigma}}_I \coloneqq \mat{T}^{\mathstrut}_{II}+ \sigma^2_I \mat{\Sigma}_I$ and $\mat{\theta}^*_I=\Tilde{\mat{\Sigma}}^{-1}_I\mat{T}_I^{\top} \mat{\theta}$ be some oracle quantities defined with more details in \Cref{lem:Y-given-I} in the Appendix. For $\mat{x}\in \mathbb{R}^d$, $\mat{x}^{\circ 2}$ is the element-wise square, and $\Var^{\circ}(\mat{x})$ is the element-wise variance as defined in \Cref{sec:notations}. With these, we define the following components: 
\begin{align*}
    & B^2(\mat{V}_I; \rho)=\mathrm{diag}\paren*{\Tilde{\mat{\Sigma}}_I}^{\top}\cbrac*{\E\paren{\mat{\theta}^*_I - \hat{\mat{\theta}}_{I} \mid \mat{V}_I}}^{\circ2}\\
    & V_a(\mat{V}_I; \rho)=\mathrm{diag}\paren*{\Tilde{\mat{\Sigma}}_I}^{\top} \E \cbrac*{\Var^{\circ}\paren{\hat{\mat{\theta}}_{I} \mid \mat{V}_I, \mat{V}_J} \mid \mat{V}_I}\\
    & V_e(\mat{V}_I; \rho)=\mathrm{diag}\paren*{\Tilde{\mat{\Sigma}}_I}^{\top} \Var^{\circ}\cbrac*{\E \paren{\hat{\mat{\theta}}_{I}\mid \mat{V}_I, \mat{V}_J} \mid \mat{V}_I},
\end{align*}
where $B^2(\mat{V}_I; \rho)$ is the bias related term; $V_a(\mat{V}_I; \rho)$ represents the \textbf{aleatoric} (intrinsic) variance, and $V_e(\mat{V}_I; \rho)$ corresponds to the \textbf{epistemic} (knowledge-based) variance.

The first main theoretical result demonstrates that increasing the disagreement penalty $\rho$ can reduce the generalization error, primarily by decreasing the intrinsic variance.

\begin{theorem}\label{thm:gen-error-general}
Suppose both students in $\cP$ satisfy \Cref{cond:related-latent-rep}. Under \Cref{cond:orthogonal}, the generalization error given $\mat{V}_I$ can be decomposed as follows:
\begin{align*}
    \text{MSE}_I(\mat{V}_I;\rho) 
    = \E\cbrac*{(Y_*-\mat{V}^{\top}_{I*} \hat{\mat{\theta}}_{I})^2 \mid \mat{V}_I}
    = B^2(\mat{V}_I; \rho) + V_a(\mat{V}_I; \rho) + V_e(\mat{V}_I; \rho) + \sigma_I^{*2},
\end{align*}
with oracle quantity $\sigma_I^{*2}$ defined in \Cref{lem:Y-given-I}. Furthermore, 
\begin{align*}
    \frac{d}{d\rho}B^2(\mat{V}_I; \rho) \mid_{\rho=0} = 0, \textcolor{white}{an} &\frac{d}{d\rho} V_e(\mat{V}_I; \rho)\mid_{\rho=0}= 0+\cO_p(n^{-1/2}), \textcolor{white}{an}
    \frac{d}{d\rho} V_a(\mat{V}_I; \rho)\mid_{\rho=0} = \Xi+\cO_p(n^{-3/2})
\end{align*}
where 
\begin{align*}
    \Xi = \frac{2\bar{\sigma}^{*2}}{n}\sum_{m=1}^{p_I}\paren*{\sum_{k=1}^{p_J}\frac{(\mat{T}_{J:k}^{\top}\mat{T}_{I:m}^{\mathstrut})^2}{(\mat{T}_{I:m}^{\top}\mat{T}_{I:m}^{\mathstrut}+\sigma_I^2)(\mat{T}_{J:k}^{\top}\mat{T}_{J:k}^{\mathstrut}+\sigma_J^2)}-1} <0,
\end{align*}
and $\bar{\sigma}^{*2}>0$ is a constant defined in \Cref{lem:Y-given-IJ}.
\end{theorem}

Equivalently, when both students receive informative latent representations satisfying \Cref{cond:related-latent-rep}, an increase in the disagreement penalty $\rho$ does not effect the bias and the epistemic variance, but effectively reduce the aleatoric variance which in turn decreases the generalization error to the unobserved data.

Note that the generalization error in \Cref{thm:gen-error-general} is averaged over the randomness of $\mat{V}_J$. The next theorem examines the generalization error under specific realizations of $\mat{V}_J$, which reveals additional insight on how it changes with respect to both $\mat{V}_I$ and $\mat{V}_J$.
\begin{theorem}\label{thm:gen-error-conditional}
Following the same conditions as in \Cref{thm:gen-error-general}, let $\mat{v}_I$ and $\mat{v}_J$ be the realizations of $\mat{V}_I$ and $\mat{V}_J$, respectively,
and define the event
\begin{align*}
\mathcal{E}=\cbrac*{\mat{v}_I, \mat{v}_J \,\middle\vert\, \paren*{\mat{v}^{-1}_{II}\mat{v}^{\mathstrut}_{IJ}\bar{\mat{\theta}}^*_J-\paren*{\mat{\theta}^*_{I} -\bar{\mat{\theta}}^*_I}} \succeq \mat{0} \text{ and }  \paren*{\mat{v}^{-1}_{II} \mat{v}^{\mathstrut}_{IJ} \mat{v}^{-1}_{JJ}\mat{v}^{\mathstrut}_{JI}\bar{\mat{\theta}}^*_I-\bar{\mat{\theta}}^*_I} \preceq \mat{0}},
\end{align*}
where the notation $\mat{x} \succeq\mat{y}$ (respectively, $\mat{x} \preceq \mat{y}$) means for $\mat{x}, \mat{y}\in \mathbb{R}^d, x_i \geq y_i$ (respectively, $ x_i \leq y_i$) for all $i \in [d]$. Under event $\cE$,
\begin{align*}
    \frac{d}{d\rho}B^2(\mat{V}_I; \rho) \mid_{\rho=0} \leq 0, \text{ and } \frac{d}{d\rho}V_e(\mat{V}_I; \rho) \mid_{\rho=0}\leq 0.
\end{align*}
\end{theorem}

Conceptually, event $\cE$ requires that the latent representations of both students provide mutually supportive knowledge. \Cref{thm:gen-error-conditional} asserts that in such cases, Meta Fusion enhances each student's performance by reducing both bias and aleatoric variance.

The event $\cE$ introduced here is abstract. To provide more concrete intuition, the following corollary analyzes a simplified, low-dimensional case.
\begin{corollary}\label{cor:simplified-event}
    Under the same condition as in \Cref{thm:gen-error-conditional}, consider the special case when $p=p_I=p_J=1$. Then, the event $\cE$ in \Cref{thm:gen-error-conditional} simplifies to
    \begin{align*}
        \mathcal{E}=\cbrac*{\mat{v}_I, \mat{v}_J \,\middle\vert\, \frac{\mat{v}_I^{\top}\mat{v}_J}{\mat{v}_I^{\top}\mat{v}_I}\geq  \frac{T_I T_J}{T_I^2+\sigma_I^2}}.
    \end{align*}
\end{corollary}
In this special case, $\mat{v}_I, \mat{v}_J \in \mathbb{R}^n$ reduce to vectors, and $T_I, T_J$ become scalars. Intuitively, this condition requires that the angle between $\mat{v}_I$ and $\mat{v}_J$ is not too large, ensuring that the latent representations from both students do not provide contradictory information.

\section{Synthetic experiments}\label{sec:synthetic-experiments}
Using synthetic experiments to illustrate the effectiveness of the proposed method and compare its performance with existing methods. We use a latent factor model where the ground-truth label depends on four components: information specific to modality $\mat{X}$, information specific to modality $\mat{Z}$, shared information, and interaction terms between modalities. More specifically,
\begin{align*}
    \mat{Y} = c_x \mat{\beta}_x f_x(\mat{X}^*) + c_z \mat{\beta}_z f_z(\mat{Z}^*) + c_s \mat{\beta}_s f_s(\mat{S}^*) + c_u \mat{\beta}_u \mat{U}^*,
\end{align*}
where $\mat{X}^*$ and $\mat{Z}^*$ are modality-specific latent covariates, $\mat{S}^*$ represents shared information, and $\mat{U}^*$ captures interaction. See \Cref{sec:synthetic-generation} for the full details on generating these variables. Here, $\mat{\beta}_t$ ($t \in \{x, z, s, u\}$) denotes the coefficient vector, $f_t(\cdot)$ is an element-wise transformation function, and $c_t$ is the component weight.

Let $p_x$ and $p_z$ denotes the dimensions of the observed feature, and let $\mat{T}_x$ and $\mat{T}_z$ be the linear mappings to the desired output dimension. Observed features are generated from a signal-plus-noise model:
\begin{align*}
        \mat{X} = (1-r_x)[\mat{X}^*, \mat{S}^*] \mat{T}_x + r_x \mat{\epsilon}_x; \quad 
        \mat{Z} = (1-r_z)[\mat{Z}^*, \mat{S}^*] \mat{T}_z + r_z \mat{\epsilon}_z,
    \end{align*}
where $[\cdot, \cdot]$ denotes column-wise concatenation, $r_x$ and $r_z \in [0,1]$ specify the noise ratios, and $\mat{\epsilon}_x$ and $\mat{\epsilon}_z$ are independent noise matrices.

\subsection{Complementary modalities}\label{sec:simu-comp-mods}
We consider scenarios where the modalities provide complementary information or contain interactions, such that the outcome variable depends jointly on both modalities. For such problems, early fusion strategies are expected to outperform late fusion approaches. Here, we examine three settings with increasing complexities:
\begin{itemize}
    \item \textbf{Setting 1.1: } Linear setting with $c_x = c_z = 1$, $c_s = c_u = 0$; $f_x$ and $f_z$ are identity functions; latent dimensions: $p^*_x = 20$, $p^*_z = 30$; observed dimensions: $p_x = 500$, $p_z = 400$; noise ratios: $r_x = r_z = 0.4$.
    \item \textbf{Setting 1.2: } Nonlinear setting with $c_x = c_z = c_u = 1$, $c_s = 0$; $f_x(\mat{X}) = \mat{X}^2 - \mat{X}$ and $f_z$ is the identity function. 
Latent dimensions: $p^*_x = p^*_z = 20$; observed dimensions: $p_x = 2000$, $p_z = 100$; noise ratios: $r_x = 0.1$, $r_z = 0.1$.
    \item \textbf{Setting 1.3: } Same as Setting 1.2, but with increased noise in the first modality: $r_x = 0.5$.
\end{itemize}
For each setting, we have generated 100 datasets. \Cref{tab:reg_early_fusion} presents the results averaged over these 100 repetitions. As expected, early fusion consistently outperforms late fusion across all settings due to the complementary nature of the modalities.
\begin{table}[!t]
    \centering
    \input{tables/regression_early_fusion}
    \caption{Performance of Meta Fusion and benchmarks under Setting 1.1-1.3. Numbers in bold highlight mean MSE values within 1 SE of the lowest MSE across all methods.}
    \label{tab:reg_early_fusion}
\end{table}
In Setting 1.1, Cooperative Learning (Coop) achieves comparable performance to early fusion by selecting an appropriate disagreement penalty. Meta Fusion, utilizing PCA-generated latent representations and the cohort construction from \Cref{sec:build-cohort}, surpasses early fusion via adaptive mutual learning and top-performer aggregation. Setting 1.2 introduces nonlinearity and dimensional disparity between $\mat{X}$ and $\mat{Z}$. While early fusion preserves all relevant information, it becomes less efficient due to modality heterogeneity. Meta Fusion, by exploring diverse student cohorts with various latent representation combinations, automatically identifies the most efficient representations, significantly outperforming benchmarks. Setting 1.3 increases the noise in $\mat{X}$, resulting in an even larger performance gap in favor of Meta Fusion. Notably, Meta Fusion also consistently exhibits smaller standard errors, demonstrating its robustness to noisy and heterogeneous modalities.

\subsection{Independent modalities}\label{sec:simu-ind-mods}
This section explores scenarios where each modality independently contains sufficient information to predict the outcome variable. For such cases, Late Fusion is expected to  outperform Early Fusion due to the independent nature of the modalities. We examine three settings with fixed 
 $c_x=c_z=c_u=0$, and $c_s=1$:
\begin{itemize}
    \item \textbf{Setting 2.1: }Linear setting with $f_s$ as the identity function; latent dimensions: $p^*_x=50, p^*_z=30, p^*_s=20$; observed dimensions: $p_x=500, p_z=400$; noise ratios: $r_x=r_z=0.4$.
    \item \textbf{Setting 2.2: }Nonlinear setting with $f_s(\mat{X})=\mat{X}^2-\mat{X}$; latent dimensions: $p^*_x=50, p^*_z=30, p^*_s=20$; observed dimensions: $p_x=2000, p_z=400$; noise ratios: $r_x=0.3, r_z=0.3$. 
    \item \textbf{Setting 2.3: }Same as Setting 1.2, but with increased noise in the first modality: $r_x = 0.5$.
\end{itemize}

\begin{table}[!t]
    \centering
    \input{tables/regression_late_fusion}
    \caption{Performance of Meta Fusion and benchmarks under Setting 2.1-2.3. Numbers in bold highlight mean MSE values within 1 SE of the lowest MSE across all methods.}
    \label{tab:reg_late_fusion}
\end{table}

Results in \Cref{tab:reg_late_fusion} are averaged over 100 repetitions. In Setting 2.1, a simple linear scenario, all benchmarks including Early Fusion perform well due to the absence of overfitting issues. Meta Fusion still outperforms other methods by effectively aggregating all students in the cohort. Setting 2.2 introduces nonlinearity and increased noise, causing Early Fusion's performance to decline. Coop successfully interpolates between Early and Late Fusion paradigms through its disagreement penalty, achieving comparable performance to Late Fusion. Notably, with moderate noise in the first modality, Late Fusion outperforms unimodal predictors. In Setting 2.3, as noise in the first modality increases further, Late Fusion's performance drops below that of the unimodal predictor trained on the cleaner second modality. However, Meta Fusion maintains its superior performance, significantly outperforming unimodal predictors. This highlights Meta Fusion's ability to automatically determine what to fuse in the presence of noisy modalities. 

In Appendix~\ref{app:ablation}, we use the above settings to conduct an ablation study that further investigate the two key components of Meta Fusion: adaptive mutual learning and ensemble learning.

\section{Real data applications}\label{sec:real-experiments}
In this section, we provide two real-data applications to illustrate the utility of our method in practice.
\subsection{Alzheimer's disease detection}\label{sec:nacc}
The rising prevalence of Alzheimer's disease (AD) poses a significant challenge to healthcare systems worldwide. 
There is an urgent need for developing efficient and accurate early detection methods. Timely diagnosis not only allows for better patient care and management but also potentially slows disease progression through early interventions. To this end, we apply Meta Fusion to analyze multimodal data from the National Alzheimer's Coordinating Center (NACC)~\citep{Besser-2018-nacc, Weintraub-2018-naccv3} dataset, aiming to enhance AD detection efficiency and accuracy. The NACC dataset includes four distinct modalities:
1.~\textit{Patient Profile} which encompasses demographic information, health history, and family medical background. 2.~\textit{Behavioral Assessment} which involves survey questions that gauge the patient's emotional state, daily functioning, and socioeconomic challenges. 3.~\textit{Clinical Evaluation} which consists of more in-depth examinations, including neurological tests (e.g., The Mini Mental State Examination \citep{folstein1975mini}) and physical assessments, conducted by healthcare professionals. 4.~\textit{MRI} which includes the summary statistics of the MRI scan, such as hippocampus and gray matter volume. The goal is to integrate these data modalities in order to classify patients into four cognitive status categories: normal cognition, mild cognitive impairment (MCI), impairment but not meeting the MCI criteria, and dementia.

\begin{figure}[!t]
    \centering
    \includegraphics[width=0.75\linewidth]{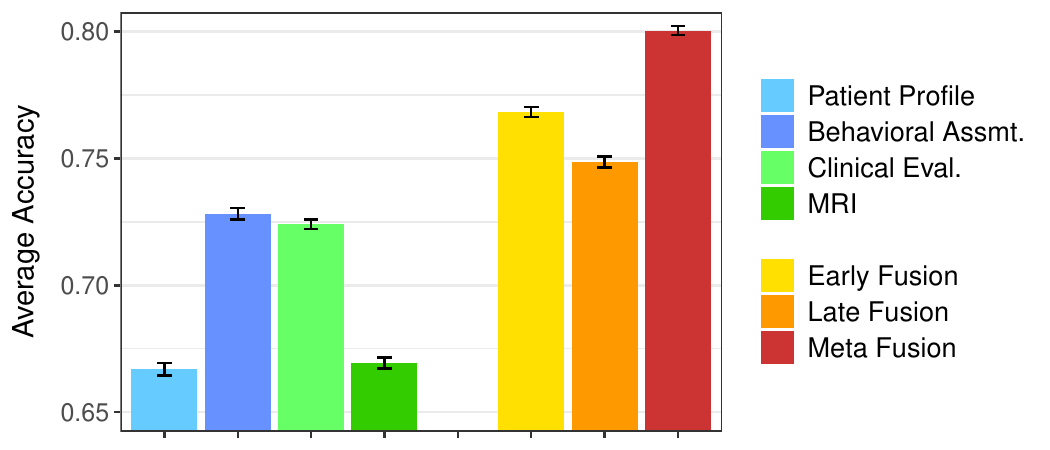}
    \caption{Average accuracy of Meta Fusion and benchmarks on the NACC dataset. Error bars indicate the standard errors. Results are summarized over 100 repetitions.}
    \label{fig:NACC}
\end{figure}

\Cref{fig:NACC} summarizes the classification performance for single-modal classifiers and fusion methods. We exclude cooperative learning from this comparison due to its undefined extension to multi-class classification and lack of clear formulation for more than two modalities, as discussed in \Cref{sec:related-work:coop}.  Our analysis, consistent with existing research, shows that behavioral assessments and clinical evaluations are most predictive for Alzheimer's disease detection~\citep{Weintraub2012-Neuropsychological-AD, Tsoi2015-cognitivetest-AD}. MRI data, while capable of detecting structural brain changes, may not reveal early-stage cognitive decline~\citep{Jack2010-biomarkers-AD}. Similarly, patient profiles indicate risk factors but may not directly reflect current cognitive status~\citep{Livingston2020-AD}.

While all fusion frameworks benefit from integrating multimodal data, the efficacy of conventional fusion benchmarks could be compromised by less-informative modalities. In contrast, Meta Fusion uses supervised encoders~\citep{Trinh2024-SE} to extract latent representations of each modality at different abstraction levels. By carefully selecting and combining powerful latent representations, Meta Fusion maximizes each modality's contribution, substantially outperforming other benchmarks.

\subsection{Neural decoding}\label{sec:neuron}

It is well established that temporal organization is critical to memory and underlies a wide range of perceptual, cognitive, and motor processes \citep{tulving72, merchant13}. While significant progress has been made in understanding how the brain encodes the spatial context of memories, our knowledge of their temporal organization remains comparatively limited. Recent electrophysiological studies have begun to address this gap (reviewed in \citep{eichenbaum14}), typically by analyzing spike trains, which are sequences of action potentials generated by neurons, and local field potentials (LFPs), which reflect the summed electrical activity in the vicinity of recording electrodes \cite{mitzdorf1985current}. Effectively integrating these dynamic, multimodal signals is essential for advancing our understanding of the neural mechanisms underlying temporal memory. To this end, we apply Meta Fusion to data collected from a novel experiment, where neural activity was recorded from the CA1 region of the hippocampus as rats performed a nonspatial sequential memory task \citep{allen-2016-neuron, shahbaba-2022-neuron}. 

The task involves the presentation of repeated sequences of odors at a single port and tests the rats' ability to identify each odor as ``in sequence'' (InSeq; by holding their nosepoke response until the signal at 1.2s) or ``out of sequence'' (OutSeq; by withdrawing their nose before the signal). Spiking and LFP activity was recorded using 24 tetrodes (bundles of 4 electrodes) in rats tested on sequences of 5 odors. For each odor presentation (trial), the data typically features spike counts from $\sim$40-70 neurons, LFP signals from 24 channels (one per tetrode), and trial identifiers (e.g., Odor presented, InSeq/OutSeq, response correct/incorrect). 

\begin{figure}[!t]
    \centering
    \includegraphics[width=0.9\linewidth]{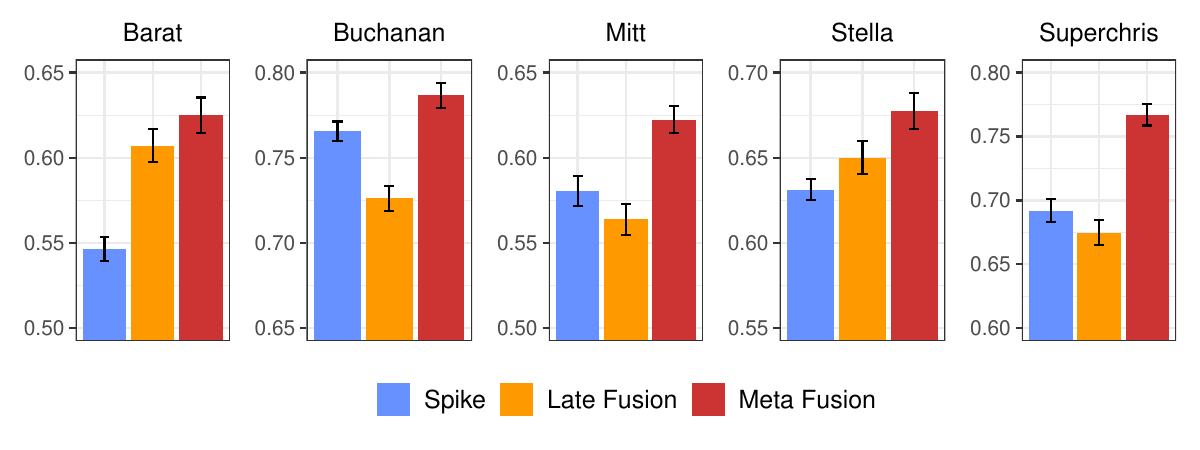}
    \caption{Classification accuracy of Meta Fusion and top benchmarks (Spike-only and Late Fusion) on the memory task for all five rats. Error bars indicate the standard errors. Results are summarized over 100 repetitions.}
    \label{fig:Neuron}
\end{figure}

Most electrophysiological studies primarily focus on spike data for decoding, as it is generally considered more informative. However, we hypothesize that incorporating both spike and LFP data could provide a more comprehensive view of hippocampal activity and potentially enhance decoding accuracy. We compare Meta Fusion's performance to that of unimodal predictors and fusion benchmarks in predicting the presented odor from the observed neural signals. \Cref{fig:Neuron} summarizes the classification accuracies among the five rats participated in the experiment. We focus on the three top performers: Meta Fusion, Late Fusion, and the unimodal classifier trained only on spike data. As expected, the LFP-only classifier performs poorly. Moreover, Early Fusion's performance is hindered by the less informative LFP data, which tends to overshadow the cleaner signals in spike data during raw modality fusion. Therefore, we focus our analysis on the top three methods. For a complete comparison of all benchmarks, see the supplementary tables and figures in \Cref{app:supp-tabs-figs}.

\Cref{fig:Neuron} shows varying performance patterns among the five rats. For rats Barat and Stella, Late Fusion outperforms the spike-only classifier, indicating benefits in incorporating the LFP modality. However, for the remaining three rats, particularly Buchanan, Late Fusion performs poorly, suggesting that complete removal of the LFP dataset might be in fact beneficial in some cases. This variability across subjects underscores the need for a fully data-driven, adaptive fusion framework, as no single approach can accommodate individual differences in neural patterns. Meta Fusion successfully adapts to these individual variations, consistently demonstrating strong performance and eliminating the need for human intervention in selecting the appropriate modality or optimal method for latent information extraction.

\section{Discussion}\label{sec:discussion}
This paper introduces Meta Fusion, a novel framework for effective integration of diverse data modalities. To address the challenges posed by heterogeneous modalities with varying informational value, Meta Fusion employs a fully data-driven three-step approach. 
Importantly, the proposed framework unifies existing data fusion paradigms as special cases. Meta Fusion's task-agnostic and model-agnostic nature ensures its broad applicability to diverse multimodal fusion tasks. 

Our work opens several new directions for future research. The current framework could be extended to handle missing modalities without requiring imputation, both during training and testing. Consider a bimodal dataset comprising one readily available modality and another that is sparsely observed. The model could be trained sequentially: initially, all students are trained on fully observed (complete) samples from both modalities. Once these are exhausted, we continue with unimodal students only, using samples with partially missing data. The soft information sharing mechanism could also be adapted to accommodate this sequential training. During the inference phase, for test data with missing modality, predictions could be aggregated from a subset of student models that utilize only the available modality. This way, our approach can offer a versatile solution for real-world scenarios where complete multimodal data are often difficult to obtain. 

The Meta Fusion framework can also be repurposed beyond the multimodal fusion context, presenting promising opportunities for extension to privacy-preserving distributed learning scenarios, particularly in federated learning (FL) contexts. Each student in the cohort could represent a local model with its private dataset, aligning with FL's decentralized nature~\citep{Li-2020-FL}. Meta Fusion's soft information sharing mechanism, which only exchanges predictions, could be adapted to guide local models towards mutual alignment while preserving data privacy. The resulting ensemble could serve as a robust global model, effectively capturing diverse local distributions without compromising individual data security.


%% file: tables/regression_early_fusion.tex
\begin{tabular}{lccc}
\toprule
 & Setting 1.1 & Setting 1.2 & Setting 1.3\\
\midrule
Modality 1 & 94.18 (1.69) & 180.71 (2.71) & 206.44 (3.00)\\
Modality 2 & 64.56 (1.30) & 256.50 (4.51) & 256.50 (4.51)\\
\addlinespace
Early Fusion & 5.38 (0.09) & 64.37 (1.07) & 122.78 (1.97)\\
Late Fusion & 41.98 (0.68) & 161.04 (2.66) & 185.83 (2.98)\\
Coop & 5.45 (0.09) & 109.53 (1.72) & 139.14 (2.30)\\
Meta Fusion & \textbf{5.07 (0.08)} & \textbf{38.51 (0.64)} & \textbf{49.59 (0.84)}\\
\bottomrule
\end{tabular}

%% file: tables/regression_late_fusion.tex
\begin{tabular}{lccc}
\toprule
 & Setting 2.1 & Setting 2.2 & Setting 2.3\\
\midrule
Modality 1 & 5.30 (0.14) & 75.23 (2.17) & 113.93 (2.86)\\
Modality 2 & 4.33 (0.10) & 64.68 (1.88) & 64.69 (1.88)\\
\addlinespace
Early Fusion & 2.67 (0.06) & 71.57 (2.07) & 100.05 (2.59)\\
Late Fusion & 2.70 (0.07) & 59.35 (1.75) & 74.75 (2.02)\\
Coop & 2.71 (0.07) & 58.17 (1.63) & 73.94 (1.93)\\
Meta Fusion & \textbf{2.47 (0.06)} & \textbf{52.27 (1.35)} & \textbf{53.38 (1.40)}\\
\bottomrule
\end{tabular}

%% file: appendix.tex
\appendix

\renewcommand{\thesection}{A\arabic{section}}
\renewcommand{\theequation}{A\arabic{equation}}
\renewcommand{\thetheorem}{A\arabic{theorem}}
\renewcommand{\thecorollary}{A\arabic{corollary}}
\renewcommand{\theproposition}{A\arabic{proposition}}
\renewcommand{\thelemma}{A\arabic{lemma}}
\renewcommand{\thetable}{A\arabic{table}}
\renewcommand{\thefigure}{A\arabic{figure}}
\renewcommand{\thealgorithm}{A\arabic{algorithm}}

\section{Implementation Details}\label{app:implementation}

\subsection{Extension to multiple modalities}\label{app:extension-multiple-modalities}
In this section, we describe how to build student cohorts for applications involving more than two modalities. Suppose we have $M>2$ modalities $\mat{X}_m \in \mathbb{R}^{n \times p_m}$, for $m \in [M]$, where $p_m \in \mathbb{Z}$ is the number of features for modality $\mat{X}_m$. Consider $k_m \geq 1$ feature extractors for $\mat{X}_m$, and denote the latent representations, including the null and identity representation, as $g^{i_m}_m(\mat{X}_m)$ for $i_m \in \cbrac{0, \dots, k_{m}+1}$. We generalize the cross-modal pairing strategy from \Cref{sec:build-cohort} into a cross-modal matching strategy, forming fused representations from different combinations of unimodal latent representations. Let $\bigtimes_{m=1}^{M}\cS_m$ denote the cartesian product of sets $\cbrac{\cS_m}_{m\in[M]}$. There are $\prod_{m=1}^M (k_{m}+2)-1$ possible nontrivial combinations, which we denote by
\begin{align*}
    \cP \coloneqq \cbrac{\bigtimes_{m=1}^M \cbrac{0, \dots, k_{m}+1}} \setminus \cbrac{(0, \dots, 0)}.
\end{align*} 
For any cross-modal combination $I = (i_1, \dots, i_M) \in \cP$, the fused representation is $\bigoplus_{m=1}^M g^{i_m}_m(\mat{X}_{m})$. Given a supervised model $f_{I}$, the corresponding student is defined as
\begin{align}\label{eq:generalized-student-model}
    f_{\Theta_{I}}(\mat{X}_{1}, \dots, \mat{X}_{m}) \coloneqq f_{I} \paren*{\bigoplus_{m=1}^M g^{i_m}_m(\mat{X}_{m})}
\end{align}
This cross-modal combination strategy explores all possible valid combinations of unimodal latent representations, resulting in a complexity of $O(\prod_{m=1}^M k_{m})$. Although the number of modalities $M$ is usually fixed and moderate in most applications, the cohort size can still grows substantially if each modality has many feature extractors. It is worth noting that the construction of the student cohort is flexible and does not need to reach its maximum complexity. If computational costs become prohibitive, it is not necessary to include all possible combinations. Instead, a smaller, more manageable subset can be selected based on domain knowledge. 

\subsection{Details of Synthetic Data Generation}\label{sec:synthetic-generation}
For all simulation studies, we use a latent factor model where the ground-truth label depends on four components: information specific to $\mat{X}$, information specific to $\mat{Z}$, shared information, and interaction terms between $\mat{X}$ and $\mat{Z}$. The data are generated as follows:
\begin{enumerate}
    \item Represent the latent dimensions for $\mat{X}$-specific, $\mat{Z}$-specific, and shared information as $p^*_x$, $p^*_z$, and $p^*_s$, respectively. Generate latent variables:
    \begin{align*}
        &\mat{X}^* \in \mathbb{R}^{n\times p^*_x},\quad \mat{X}^*_{ij} \sim N(0,1);\\
        &\mat{Z}^* \in \mathbb{R}^{n\times p^*_z},\quad \mat{Z}^*_{ij} \sim N(0,1);\\
        &\mat{S}^* \in \mathbb{R}^{n\times p^*_s},\quad \mat{S}^*_{ij} \sim N(0,1).
    \end{align*}
    \item Define the interaction terms using the Kronecker product:
    \begin{align*}
        \mat{U}^* \in \mathbb{R}^{n\times p^*_x p^*_z},\quad \mat{U}^*_i = \mat{X}^*_i \otimes \mat{Z}^*_i,
    \end{align*}
    where $\mat{X}^*_i$ and $\mat{Z}^*_i$ are the $i$th rows of $\mat{X}^*$ and $\mat{Z}^*$.
    \item Generate the ground-truth label $\mat{Y}$ as:
    \begin{align*}
        \mat{Y} = c_x \mat{\beta}_x f_x(\mat{X}^*) + c_z \mat{\beta}_z f_z(\mat{Z}^*) + c_s \mat{\beta}_s f_s(\mat{S}^*) + c_u \mat{\beta}_u \mat{U}^*,
    \end{align*}
    where $\mat{\beta}_t$ ($t \in \{x,z,s,u\}$) is the vector of coefficients, $f_t(\cdot)$ applies an element-wise transformation (e.g., quadratic) to each latent component, and $c_t$ is the weight that controls the contribution of a specific latent component. For instance, setting $c_t=0$ removes that component's influence on $\mat{Y}$.
    \item Let $p_x$ and $p_z$ be the observed feature dimensions. The observed features are generated using a signal-plus-noise model based on the latent components: 
\begin{align*}
        &\mat{X} = (1-r_x)[\mat{X}^*, \mat{S}^*] \mat{T}_x + r_x \mat{\epsilon}_x, \quad \mat{T}_x \in \mathbb{R}^{(p^*_x+p^*_s)\times p_x};\\
        &\mat{Z} = (1-r_z)[\mat{Z}^*, \mat{S}^*] \mat{T}_z + r_z \mat{\epsilon}_z ,\quad \mat{T}_z \in \mathbb{R}^{(p^*_z+p^*_s)\times p_z},
    \end{align*}
    where $[\cdot, \cdot]$ denotes column-wise concatenation, $r_x, r_z \in [0,1]$ specify the noise ratios, and $\mat{\epsilon}_x$ and $\mat{\epsilon}_z$ are noise matrices with entries independently sampled from $N(0,1)$.
\end{enumerate}

\subsection{Different Ensemble techniques}\label{app:ensemble-methods}
In addition to the ensemble selection method presented in \Cref{alg:ensemble-selection}, the final prediction can be aggregated using various standard ensemble techniques. Recall that for $I \in \cP$, $\widehat{\mat{Y}}_{I}\in \mathbb{R}^{n\times d}$ denotes the prediction from student model $f_{\Theta_{I}}$, where $d$ is the output dimension. For instance, $d=1$ for regression tasks and $d=C$ for classification tasks with $C$ possible classes, where $\widehat{\mat{Y}}_{I}\in \mathbb{R}^{n\times C}$ denotes the corresponding logits. Below, we present the implementation details of several ensemble alternatives used in our numerical analysis in \Cref{app:ablation} and \Cref{app:supp-tabs-figs}. Let $\mat{Y}^{\mathrm{ens}} \in \mathbb{R}^{n\times d}$ denote the final ensemble prediction formulated as follows:
\begin{itemize}
    \item \textbf{Best Single}: This method selects the best performer $I^*$ among the student cohort based on task-specific loss evaluated on the holdout sample $\cD^{\mathrm{val}}$. The final prediction is:
    \begin{align*}
        \mat{Y}^{\mathrm{ens}} = \widehat{\mat{Y}}_{I^*}
    \end{align*}
    
    \item \textbf{Stacking}: This method fits an additional supervised model based on stacked outputs of all students. Specifically, recall that $\bigoplus_{I\in \cP}\mat{Y}_{I}$ denotes column-wise concatenation. Let $f^{\mathrm{ens}}$ denote the supervised model (e.g., a simple DNN model). The final prediction is:
    \begin{align*}
        \mat{Y}^{\mathrm{ens}} = f^{\mathrm{ens}} \paren*{\bigoplus\nolimits_{I\in \cP}\mat{Y}_{I}}
    \end{align*}
    \item \textbf{Simple Average}: This method averages the outputs from all students:
    \begin{align*}
        \mat{Y}^{\mathrm{ens}} = \frac{1}{|\cP|} \sum_{I \in \cP}\widehat{\mat{Y}}_{I}
    \end{align*}
    \item \textbf{Weighted Average}: Given the diversity of the student cohort, a simple average may not be optimal, as the presence of low-performing students could degrade the overall ensemble prediction. Alternatively, one can use a weighted average to dynamically scale the contribution of individual students. Let $\cbrac{w_I}_{I \in \cP}$ be a set of normalized weights that sum to 1. The ensemble prediction is then given by:
    \begin{align*}
        \mat{Y}^{\mathrm{ens}} = \sum_{I \in \cP} w_{I}\widehat{\mat{Y}}_{I},
    \end{align*}
    In practice, the weights $\cbrac{w_I}_{I \in \cP}$ are chosen to be proportional to the student's task performance evaluated on the holdout set $\cD^{\mathrm{val}}$.
\end{itemize}
For classification tasks, we consider two additional ensemble methods based on voting. Note that $\widehat{\mat{Y}}_{I}\in \mathbb{R}^{n\times C}$ denotes the logits. Let $\tilde{\mat{Y}}_{I} \coloneqq \argmax_{c\in [C]} \widehat{\mat{Y}}_{I} \in \mathbb{R}^{n}$ denote the labels predicted by student $I$, which finds the largest logit for each sample, and let $\tilde{\mat{Y}}^{\mathrm{ens}}$ denote the label predicted by the ensemble. The two additional ensemble methods are:
\begin{itemize}
    \item \textbf{Majority Vote}: This method determines the most frequent prediction among the student cohort:
    \begin{align*}
        \tilde{\mat{Y}}^{\mathrm{ens}} = \argmax_{c\in [C]} \paren*{ \sum_{I\in \cP} \I \cbrac*{\tilde{\mat{Y}}_{I} = c}}
    \end{align*}
    \item \textbf{Weighted Vote}: Similar to weighted averaging, we can adjust the vote based on weights proportional to each student's performance:
    \begin{align*}
        \tilde{\mat{Y}}^{\mathrm{ens}} = \argmax_{c\in [C]} \paren*{ w_{I} \sum_{I\in \cP} \I \cbrac*{\tilde{\mat{Y}}_{I} = c}}.
    \end{align*}
\end{itemize}

\section{Connection to Cooperative Learning}\label{app:connection-to-coop}
In this section, we show an alternative form of cooperative learning could be viewed as a special case of our proposed framework. Specifically, consider the following simplified objective function of \eqref{eq:coop-objective} by excluding explicit cross-modal interactions. This simplified objective function is the primary focus in \citet{ding2022-coop}.
\begin{align}\label{eq:simple-coop-objective}
    \min_{f_x, f_z} \mathbb{E}\cbrac*{\frac{1}{2}\paren*{Y-f_{x}(X)-f_{z}(Z)}^2 + \frac{\rho}{2}\norm*{f_{x}(X)-f_{z}(Z)}^2},
\end{align}
which has solutions
\begin{equation}\label{eq:simple-coop-solution}
\begin{aligned}    f_x(X) = \mathbb{E}\cbrac*{\frac{Y}{1+\rho} - \frac{(1-\rho)f_{z}(Z)}{(1+\rho)} \bigmid X}, \quad 
    f_{z}(Z) = \mathbb{E}\cbrac*{\frac{Y}{1+\rho} - \frac{(1-\rho)f_{x}(X)}{(1+\rho)} \bigmid Z}.
\end{aligned}
\end{equation}
To clarify this connection, we consider the population setting used in cooperative learning, as specified in Equation \eqref{eq:simple-coop-objective}, where the modalities $X\in \mathbb{R}^{p_x},Z\in \mathbb{R}^{p_z}$ are treated as random variables. We show that \eqref{eq:simple-coop-objective} is a special case of our framework involving two students: $\widehat{Y}_{\Theta_{k_x+1, 0}} \coloneqq f_{\Theta_{k_x+1, 0}}(X, Z)$ that uses $X$ only, and $\widehat{Y}_{\Theta_{0, k_z+1}} \coloneqq f_{\Theta_{0, k_z+1}}(X, Z)$, that uses $Z$ only. Denote $I_x \coloneqq (k_x+1, 0)$, and $I_z \coloneqq (0, k_z+1)$. We use simple averaging to aggregate predictions. For regression analysis, suppose that both the task loss $\cL$ and the divergence measure $\cD$ are defined based on MSE, and that $d_{I_x, I_z}=d_{I_z, I_x}=1$. In this case, the objective in Equation \eqref{eq:overall-loss} for the ensemble predictor becomes:
\begin{align}\label{eq:simplified-overall-loss}
    \cL_{\Theta_{I_x}}= \cL_{\Theta_{I_z}}= \mathbb{E}\cbrac*{ \paren*{Y-\frac{1}{2}\paren*{\widehat{Y}_{I_x}+\widehat{Y}_{I_z}}}^2+\rho \paren*{\widehat{Y}_{I_z}-\widehat{Y}_{I_x}}^2},
\end{align}
One can show that the solutions are
\begin{equation}\label{eq:simplified-solution}
\begin{aligned}
    \widehat{Y}_{I_x}  = \mathbb{E}\cbrac*{\frac{2Y}{1+\Tilde{\rho}} - \frac{(1-\Tilde{\rho})\widehat{Y}_{I_z}}{(1+\Tilde{\rho})} \Bigmid X},\quad
    \widehat{Y}_{I_z} = \mathbb{E}\cbrac*{\frac{2Y}{1+\Tilde{\rho}} - \frac{(1-\Tilde{\rho}) \widehat{Y}_{I_x}}{(1+\Tilde{\rho})} \Bigmid Z},
\end{aligned}
\end{equation}
where $\Tilde{\rho} = 4\rho$ is an adjustable hyperparameter. Consequently, the solutions obtained through cooperative learning in \eqref{eq:simple-coop-solution}, and the solutions from the alternative implementation of our framework in \eqref{eq:simplified-solution} result in the same final prediction.

\section{Additional numerical experiments}\label{app:additional-num-exp}
\subsection{Ablation study}\label{app:ablation}
This section examines the two key components of Meta Fusion: adaptive mutual learning and ensemble of the student cohort, using the numerical settings in \Cref{sec:simu-ind-mods}.

\begin{figure}[!htb]
    \centering
    \includegraphics[width=\linewidth]{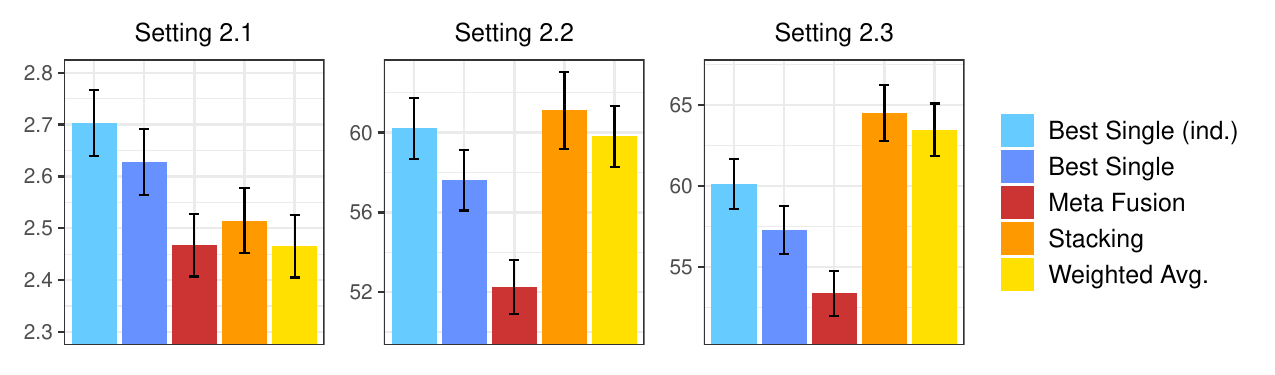}
    \caption{Average MSE of Meta Fusion and ensemble baselines across Settings 2.1-2.3 (\Cref{sec:simu-ind-mods}). Error bars represent standard errors.}
    \label{fig:late_ensemble}
\end{figure}

To demonstrate the impact of adaptive mutual learning, we compare two key baselines detailed in \Cref{app:ensemble-methods}: ``Best Single," which selects the top-performing model from a cohort trained with adaptive mutual learning, and ``Best Single (ind.)," which chooses the best model from a cohort trained independently. In \Cref{fig:late_ensemble}, Best Single consistently outperforms its independent counterpart, highlighting the efficacy of adaptive mutual learning in enhancing individual model performance within the cohort.

Further, we evaluate several ensemble techniques within our Meta Fusion framework (see Appendix~\ref{app:ensemble-methods} for implementation details). While all methods perform similarly in the simple Setting 2.1, the challenging Settings 2.2 and 2.3 reveal key differences. Classical techniques like Stacking and Weighted Average struggle with increased student diversity, often performing worse than the best single model. In contrast, Meta Fusion's ensemble selection strategy demonstrates robust performance across all settings, effectively handling student heterogeneity in challenging scenarios.

\subsection{Comparison of Different Divergence Weights}\label{app:comp-div-weights} 
As discussed in \Cref{sec:training}, incorrectly specified divergence weights can potentially introduce negative knowledge transfer and hinder the effectiveness of mutual learning. In this section, we examine the impact of different divergence weights on model performance. 

\begin{figure}[ht]
    \centering
    \begin{minipage}[b]{0.32\textwidth}  
        \centering
        \includegraphics[width=\textwidth]{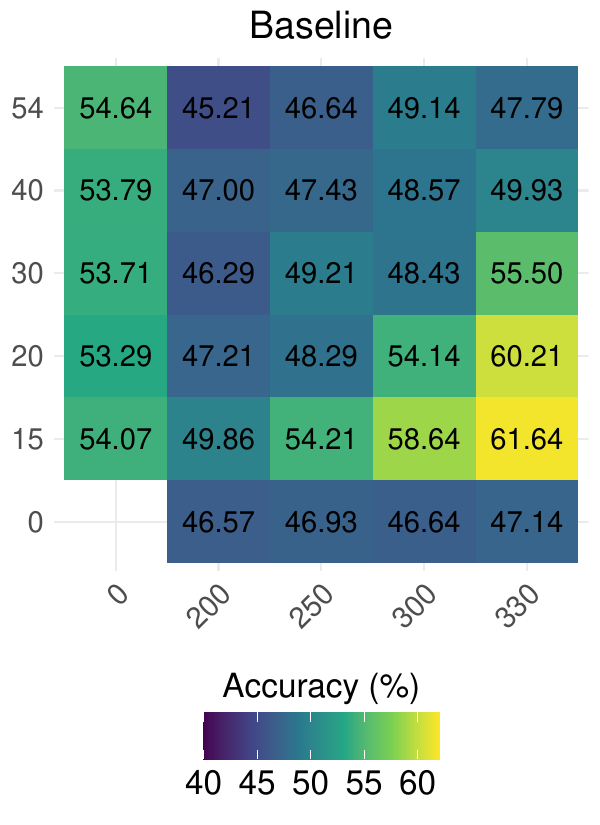} 
        \subcaption{Baseline Accuracies}
        \label{fig:baseline}
    \end{minipage}
    \hfill
    \begin{minipage}[b]{0.64\textwidth} 
        \centering
        \includegraphics[width=\textwidth]{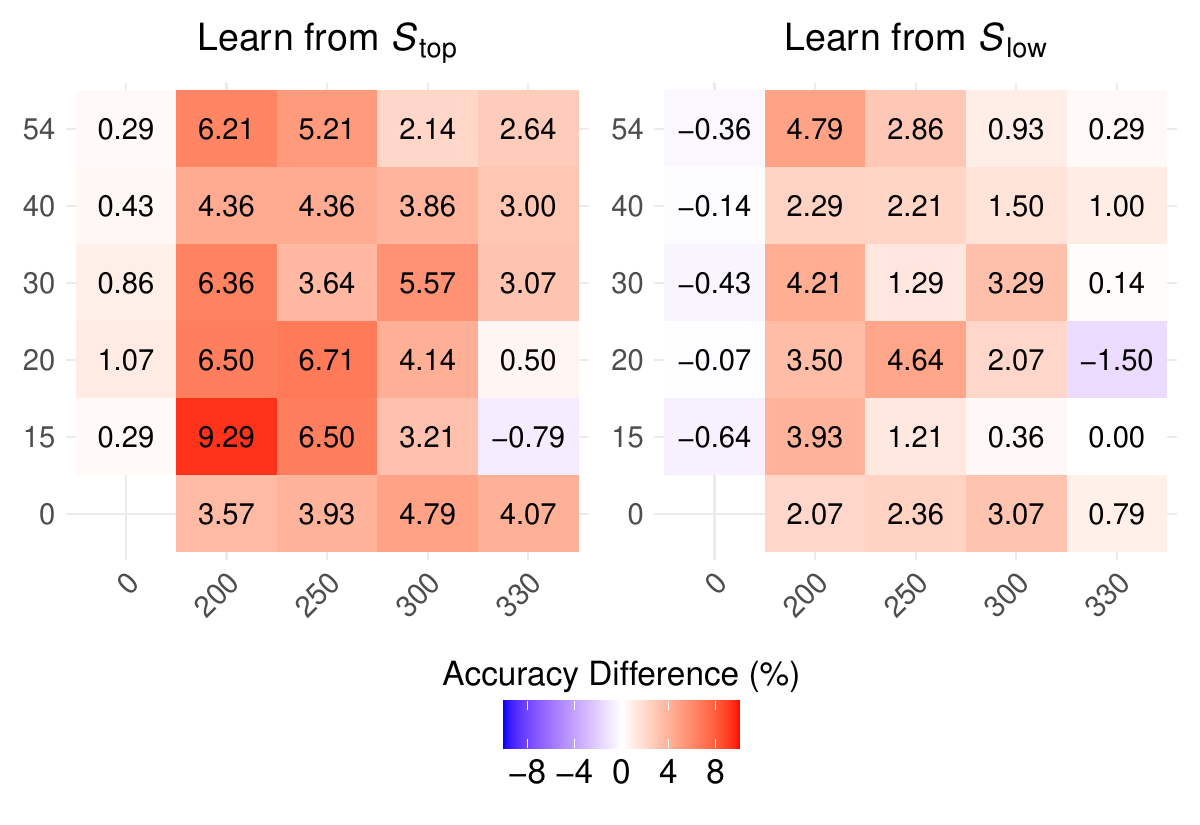}  
        \subcaption{Accuracy Differences Compared to the Baseline} 
        \label{fig:diff}
    \end{minipage}
    \caption{Comparison of student performance trained with different divergence weights for the neural decoding task using Barat's data (see details in \Cref{sec:neuron}). The results are averaged over 100 experimental repetitions. In all panels, x-axis and y-axis represent latent dimensions of LFP and spike data, respectively. Each grid cell shows a student model with a specific combination of latent representations. (a) shows baseline accuracies of independently trained models without mutual learning. (b) displays accuracy differences relative to the baseline when models learn from the best-performing cluster (left) and the worst-performing cluster (right).}
    \label{fig:div-comp}
\end{figure}

Recall that the divergence weights $\cbrac{d_{I,J}}_{I,J \in \cP, J\neq I}$ in \eqref{eq:overall-loss} are determined by cluster membership:
\begin{align*}
    d_{I,J} =
    \begin{cases} 
    1, & \text{if } J \in \cS_{\mathrm{top}} \\
    0, & \text{otherwise},
    \end{cases}
\end{align*}
where $\cS_{\mathrm{top}}$ is the set of models belonging to the best-performing $k_{\mathrm{top}}\geq 1$ clusters with the lowest average initial loss evaluated on the holdout validation set.
We investigate the effect of different divergence weights using the neural decoding experiment for one of the rats (Barat, see \Cref{sec:neuron}). \Cref{fig:div-comp} illustrates our findings. In all panels, the x-axis represents the latent dimension of the LFP data, while the y-axis represents the latent dimension of spike data. Each grid cell corresponds to a student model with a specific combination of latent representations.

\Cref{fig:baseline} shows the baseline average accuracy of each student in the cohort trained independently (i.e., $\rho=0$ in \eqref{eq:overall-loss}). \Cref{fig:diff} displays the accuracy differences (with respect to the baseline) for cohorts trained with different divergence weights, both using $\rho=5$. In the left panel of \Cref{fig:diff}, we set $k_{\mathrm{top}}=1$, meaning students only learn from $\cS_{\mathrm{top}}$ in the best-performing cluster. Conversely, in the right panel, we define $\cS_{\mathrm{low}}$ as the set of models in the worst-performing cluster and set $d_{I,J}=1$ if $J \in \cS_{\mathrm{low}}$ and 0 otherwise, so students only learn from the low performers.

As we can see, both strategies lead to improved individual performances compared to the baseline cohort. However, when students learn from $\cS_{\mathrm{top}}$, the performance gain is substantially larger than learning from $\cS_{\mathrm{low}}$. The performance gap between the two sets of divergence weights indicates the presence of negative knowledge transfer when weights are inappropriately specified, which in turn reduces the effectiveness of mutual learning. This underscores the importance of careful weight selection in mutual learning scenarios with heterogeneous student models.

\section{Proofs}\label{app:proofs}
This section presents the proofs of our theoretical results. \Cref{app:proofs-lemmas} contains auxiliary technical lemmas, \Cref{app:proofs-main} provides detailed proofs for the main theoretical results presented in the manuscript.

\subsection{Notations}
We use the following technical notations:
\begin{itemize}
    \item For a matrix $\mat{M}$, $\mat{M}_{i:}$ denotes the $i$-th row and $\mat{M}_{:i}$ denotes the $i$-th column.
    \item For matrices $\mat{M}_{I}$,$\mat{M}_{J}$ indexed by $I, J\in \cP$, we denote $\mat{M}^{\top}_I\mat{M}_J$ as $\mat{M}^{\mathstrut}_{IJ}$ for simplicity.
    \item For $\mat{x} \in \mathbb{R}^d$, $\mat{x}^{\circ 2} \coloneqq (x_1^2, \dots, x_d^2)$ denotes the element-wise square, and $\Var^{\circ}(\mat{x}) \coloneqq (\Var(x_1), \dots, \Var(x_d))$ denotes the element-wise variance.
\end{itemize}

\subsection{Additional Technical Results}\label{app:proofs-lemmas}

\begin{lemma}\label{lem:Y-given-IJ}
Assume the ground truth label $\mat{Y}=(Y_1, \dots, Y_n) \in \mathbb{R}^n$ follows a linear latent factor model $\mat{Y} = \mat{V}\mat{\theta}$ for some coefficients $\mat{\theta} \in \mathbb{R}^p$ and latent factors $\mat{V} \in \mathbb{R}^{n\times p}$ according to model \eqref{eq:latent-ground-truth}. Assume the fused representation $\mat{V}_I$ and $\mat{V}_J$ satisfy \Cref{cond:related-latent-rep}. 
Let 
\begin{itemize}
   \item $\mat{V}_{Ii:}$ and $\mat{V}_{Ji:}$ denote the $i$-th row of $\mat{V}_I$ and $\mat{V}_J$, respectively;
   \item $\Tilde{\mat{\Sigma}}_I \coloneqq \mat{T}^{\mathstrut}_{II}+ \sigma^2_I \mat{\Sigma}_I$ and $\Tilde{\mat{\Sigma}}_J \coloneqq \mat{T}^{\mathstrut}_{JJ}+ \sigma^2_J \mat{\Sigma}_J$.
\end{itemize} 
Then $Y_i$, for $i \in [n]$, can be expressed as:
\begin{align*}
    Y_i =  \mat{V}_{Ii:}^{\top}\bar{\mat{\theta}}^*_I + \mat{V}_{Ji:}^{\top}\bar{\mat{\theta}}^*_J + \bar{\epsilon}_i^*, \quad \bar{\epsilon}_i^* \sim \cN(0, \bar{\sigma}^{*2}),
\end{align*}
where $\bar{\epsilon}_i^*$ is an independent error term and 
\begin{align*}
    \bar{\mat{\theta}}^*_I
    &=(\mat{A}^{\top}\mat{T}_I^{\top}+\mat{C}^{\top}\mat{T}_J^{\top})\mat{\theta}, \\
    \bar{\mat{\theta}}^*_J 
    &= (\mat{B}^{\top}\mat{T}_I^{\top}+\mat{D}^{\top}\mat{T}_J^{\top})\mat{\theta}, \\
    \bar{\sigma}^{*2}
    &=\mat{\theta}^{\top}\paren*{\mat{\Sigma}-(\mat{T}_I\mat{A}+\mat{T}_J\mat{C})\mat{T}_I^{\top}-(\mat{T}_I\mat{B}+\mat{T}_J\mat{D})\mat{T}_J^{\top}}\mat{\theta},
\end{align*}
where $\mat{A}\in \mathbb{R}^{p_I\times p_I}$, $\mat{B}\in \mathbb{R}^{p_I\times p_J}$, $\mat{C}\in \mathbb{R}^{p_J\times p_I}$ and $\mat{D}\in \mathbb{R}^{p_J\times p_J}$ satisfies
\begin{align}\label{eq:block-inverse-covar}
     \begin{pmatrix}
        \mat{A} & \mat{B}\\
        \mat{C} & \mat{D}
    \end{pmatrix} 
    =\begin{pmatrix}
        \Tilde{\mat{\Sigma}}_I & \mat{T}^{\mathstrut}_{IJ}\\
        \mat{T}^{\mathstrut}_{JI} & \Tilde{\mat{\Sigma}}_J
    \end{pmatrix}^{-1}.
\end{align}
\end{lemma}
\begin{remark}
    The inverse in \eqref{eq:block-inverse-covar} exists since the block matrix on the right hand side is positive definite.
\end{remark}
\begin{proof}
    By model \eqref{eq:latent-ground-truth} and \Cref{cond:related-latent-rep}, the latent presentations for each $i\in [n]$ follows a multivariate Gaussian distribution, namely 
\begin{equation}\label{eq:joint-distribution-all}
\begin{aligned}
    \begin{pmatrix}
        \mat{V}_{i:}\\
        \mat{V}_{Ii:}\\
        \mat{V}_{Ji:}
    \end{pmatrix}
    \sim
    \cN \paren*{
    \begin{pmatrix}
        \mat{0}\\
        \mat{0}\\
        \mat{0}
    \end{pmatrix},
    \begin{pmatrix}
        \mat{\Sigma} & \mat{T}_I & \mat{T}_J\\
        \mat{T}_I^{\top} & \Tilde{\mat{\Sigma}}_I & \mat{T}^{\mathstrut}_{IJ}\\
        \mat{T}_J^{\top} & \mat{T}^{\mathstrut}_{JI} & \Tilde{\mat{\Sigma}}_J
    \end{pmatrix}
    }.
\end{aligned}
\end{equation}

The conditional distribution of $\mat{V}_{i:}$ given $\mat{V}_{Ii:}$ and $\mat{V}_{Ji:}$ is still Gaussian with
\begin{align}\label{eq:gaussian-given-IJ}
    \mat{V}_{i:}\mid \mat{V}_{Ii:}, \mat{V}_{Ji:} \sim \cN \paren*{\E \brac*{\mat{V}_{i:}\mid \mat{V}_{Ii:}, \mat{V}_{Ji:}}, \Var \brac*{\mat{V}_{i:}\mid \mat{V}_{Ii:}, \mat{V}_{Ji:}}},
\end{align}
where
\begin{align*}
    \E \brac*{\mat{V}_{i:}\mid \mat{V}_{Ii:}, \mat{V}_{Ji:}} 
    &=
    \begin{pmatrix}
        \mat{T}_I & \mat{T}_J
    \end{pmatrix}
    \begin{pmatrix}
        \Tilde{\mat{\Sigma}}_I & \mat{T}^{\mathstrut}_{IJ}\\
        \mat{T}^{\mathstrut}_{JI} & \Tilde{\mat{\Sigma}}_J
    \end{pmatrix}^{-1}
    \begin{pmatrix}
        \mat{V}_{Ii:} \\
        \mat{V}_{Ji:}
    \end{pmatrix}
\end{align*}
The inverse in the right hand side always exists since the target covariance matrix is positive definite. Let $\mat{A}\in \mathbb{R}^{p_I\times p_I}$, $\mat{B}\in \mathbb{R}^{p_I\times p_J}$, $\mat{C}\in \mathbb{R}^{p_J\times p_I}$ and $\mat{D}\in \mathbb{R}^{p_J\times p_J}$ be the matrices satisfying \eqref{eq:block-inverse-covar}, then we have
\begin{align*}
    \E \brac*{\mat{V}_{i:}\mid \mat{V}_{Ii:}, \mat{V}_{Ji:}} 
    &=\begin{pmatrix}
        \mat{T}_I & \mat{T}_J
    \end{pmatrix}
    \begin{pmatrix}
        \mat{A} & \mat{B}\\
        \mat{C} & \mat{D}\
    \end{pmatrix}
    \begin{pmatrix}
        \mat{V}_{Ii:} \\
        \mat{V}_{Ji:}
    \end{pmatrix}\\
    &=(\mat{T}_I\mat{A}+\mat{T}_J\mat{C})\mat{V}_{Ii:}+(\mat{T}_I\mat{B}+\mat{T}_J\mat{D})\mat{V}_{Ji:}
\end{align*}
and 
\begin{align*}
    \Var \brac*{\mat{V}_{i:}\mid \mat{V}_{Ii:}, \mat{V}_{Ji:}}
    &= \mat{\Sigma}-
   \begin{pmatrix}
        \mat{T}_I & \mat{T}_J
    \end{pmatrix}
    \begin{pmatrix}
        \Tilde{\mat{\Sigma}}_I & \mat{T}^{\mathstrut}_{IJ}\\
        \mat{T}^{\mathstrut}_{JI} & \Tilde{\mat{\Sigma}}_J
    \end{pmatrix}^{-1}
     \begin{pmatrix}
        \mat{T}_I \\
        \mat{T}_J
    \end{pmatrix}\\
    &= \mat{\Sigma}-
   \begin{pmatrix}
        \mat{T}_I & \mat{T}_J
    \end{pmatrix}
    \begin{pmatrix}
        \mat{A} & \mat{B}\\
        \mat{C} & \mat{D}\
    \end{pmatrix}
     \begin{pmatrix}
        \mat{T}_I \\
        \mat{T}_J
    \end{pmatrix}\\
    &=\mat{\Sigma}-(\mat{T}_I\mat{A}+\mat{T}_J\mat{C})\mat{T}_I^{\top}-(\mat{T}_I\mat{B}+\mat{T}_J\mat{D})\mat{T}_J^{\top}.
\end{align*}
Recall that $Y_i=\mat{V}_{i:}^{\top}\mat{\theta}$, hence 
\begin{equation}\label{eq:exp-Y-given-IJ}
\begin{aligned}
    \E \brac*{Y_i \mid \mat{V}_{Ii:}, \mat{V}_{Ji:}} 
    &= \E \brac*{\mat{V}_{i:}^{\top}\mat{\theta} \mid \mat{V}_{Ii:}, \mat{V}_{Ji:}} \\
    &= \mat{V}_{Ii:}^{\top}(\mat{A}^{\top}\mat{T}_I^{\top}+\mat{C}^{\top}\mat{T}_J^{\top})\mat{\theta} + \mat{V}_{Ji:}^{\top}(\mat{B}^{\top}\mat{T}_I^{\top}+\mat{D}^{\top}\mat{T}_J^{\top})\mat{\theta},
\end{aligned}
\end{equation}

and 
\begin{equation}\label{eq:var-Y-given-IJ}
\begin{aligned}
    \Var \brac*{Y_i \mid \mat{V}_{Ii:}, \mat{V}_{Ji:}} 
    &= \Var \brac*{\mat{V}_{i:}^{\top}\mat{\theta} \mid \mat{V}_{Ii:}, \mat{V}_{Ji:}}\\
    &=  \mat{\theta}^{\top} \Var\brac*{\mat{V}_{i:} \mid \mat{V}_{Ii:}, \mat{V}_{Ji:}}\mat{\theta}\\
    &= \mat{\theta}^{\top}\paren*{\mat{\Sigma}-(\mat{T}_I\mat{A}+\mat{T}_J\mat{C})\mat{T}_I^{\top}-(\mat{T}_I\mat{B}+\mat{T}_J\mat{D})\mat{T}_J^{\top}}\mat{\theta}
\end{aligned}
\end{equation}
Let
\begin{align*}
    \bar{\mat{\theta}}^*_I &=(\mat{A}^{\top}\mat{T}_I^{\top}+\mat{C}^{\top}\mat{T}_J^{\top})\mat{\theta},\\ 
    \bar{\mat{\theta}}^*_J &= (\mat{B}^{\top}\mat{T}_I^{\top}+\mat{D}^{\top}\mat{T}_J^{\top})\mat{\theta}, \\ 
    \bar{\sigma}^{*2} &=\mat{\theta}^{\top}\paren*{\mat{\Sigma}-(\mat{T}_I\mat{A}+\mat{T}_J\mat{C})\mat{T}_I^{\top}-(\mat{T}_I\mat{B}+\mat{T}_J\mat{D})\mat{T}_J^{\top}}\mat{\theta},
\end{align*}
then combining \eqref{eq:gaussian-given-IJ}, \eqref{eq:exp-Y-given-IJ} and \eqref{eq:var-Y-given-IJ}, $Y_i$ can be expressed as 
\begin{align*}
     Y_i =  \mat{V}_{Ii:}^{\top}\bar{\mat{\theta}}^*_I + \mat{V}_{Ji:}^{\top}\bar{\mat{\theta}}^*_J + \bar{\epsilon}_i^*, \quad \bar{\epsilon}_i^* \sim \cN(0, \bar{\sigma}^{*2})
\end{align*}
where $\bar{\epsilon}_i^*$ is an independent error term.
\end{proof}

\Cref{lem:Y-given-IJ} characterizes the distribution of $\mat{Y}$ given fused representations $\mat{V}_I, \mat{V}_J$ from both students. The following Lemma delineates the conditional distribution of $\mat{Y}$ given only $\mat{V}_I$.

\begin{lemma}\label{lem:Y-given-I}
   Under the same conditions as in \Cref{lem:Y-given-IJ}, $Y_i$, for $i \in [n]$, can be alternatively expressed as:
\begin{align*}
    Y_i = \mat{V}_{Ii:}^{\top} \mat{\theta}^*_I + \epsilon_{Ii}^*, \quad \epsilon_{Ii}^* \sim \cN(0, \sigma_I^{*2}),
\end{align*}
where $\epsilon_{Ii}^*$ is an error term independent of $\mat{V}_{Ii:}$ and 
\begin{align*}
    \mat{\theta}^*_I&=(\Tilde{\mat{\Sigma}}_I)^{-1}\mat{T}_I^{\top} \mat{\theta}, \\
    \sigma_I^{*2}&=\mat{\theta}^{\top}\paren*{\mat{\Sigma}-\mat{T}_I\Tilde{\mat{\Sigma}}^{-1}_I\mat{T}_I^{\top}}\mat{\theta},
\end{align*}
\end{lemma}

\begin{proof}
    By \Cref{cond:related-latent-rep}, the joint distribution of $(\mat{V}_{i:}, \mat{V}_{Ii:})$, for $i\in[n]$ is
\begin{align*}
    \begin{pmatrix}
        \mat{V}_{i:}\\
        \mat{V}_{Ii:}
    \end{pmatrix}
    \sim
    \cN \paren*{
    \begin{pmatrix}
        \mat{0}\\
        \mat{0}
    \end{pmatrix},
    \begin{pmatrix}
        \mat{\Sigma} & \mat{T}_I\\
        \mat{T}_I^{\top} & \Tilde{\mat{\Sigma}}_I
    \end{pmatrix}
    }.
\end{align*}
The conditional distribution of $\mat{V}_{i:}$ given $\mat{V}_{Ii:}$ is still Gaussian with
\begin{align}\label{eq:gaussian-given-I}
    \mat{V}_{i:}\mid \mat{V}_{Ii:} \sim \cN \paren*{\E \paren*{\mat{V}_{i:}\mid \mat{V}_{Ii:}}, \Var \paren*{\mat{V}_{i:}\mid \mat{V}_{Ii:}}},
\end{align}
where
\begin{align*}
    \E \paren*{\mat{V}_{i:}\mid \mat{V}_{Ii:}} 
    &= \mat{T}_I\Tilde{\mat{\Sigma}}^{-1}_I\mat{V}_{Ii:}
\end{align*}
and 
\begin{align*}
    \Var \paren*{\mat{V}_{i:}\mid \mat{V}_{Ii:}}
    &= \mat{\Sigma}-\mat{T}_I\Tilde{\mat{\Sigma}}^{-1}_I\mat{T}_I^{\top}.
\end{align*}
Note that the inverse on the right hand side always exist since the target covariance matrix is positive definite. Recall that $Y_i=\mat{V}_{i:}^{\top}\mat{\theta}$, it follows that
\begin{align}\label{eq:exp-Y-given-I}
    \E \paren*{Y_i \mid \mat{V}_{Ii:}} = \E \paren*{\mat{V}_{i:}^{\top}\mat{\theta} \mid \mat{V}_{Ii:}} 
    =  \mat{V}_{Ii:}^{\top}\Tilde{\mat{\Sigma}}^{-1}_I\mat{T}_I^{\top} \mat{\theta},
\end{align}
and 
\begin{align}\label{eq:var-Y-given-I}
    \Var \paren*{Y_i \mid \mat{V}_{Ii:}} = \Var \paren*{\mat{V}_{i:}^{\top}\mat{\theta} \mid \mat{V}_{Ii:}}
    = \mat{\theta}^{\top} \paren*{\mat{\Sigma}-\mat{T}_I\Tilde{\mat{\Sigma}}^{-1}_I\mat{T}_I^{\top}}\mat{\theta}.
\end{align}
Let
\begin{align*}
    \mat{\theta}^*_I&=\Tilde{\mat{\Sigma}}^{-1}_I\mat{T}_I^{\top} \mat{\theta},\\
    \sigma_I^{*2}&=\mat{\theta}^{\top} \paren*{\mat{\Sigma}-\mat{T}_I\Tilde{\mat{\Sigma}}^{-1}_I\mat{T}_I^{\top}}\mat{\theta},
\end{align*}
by \eqref{eq:gaussian-given-I}, \eqref{eq:exp-Y-given-I} and \eqref{eq:var-Y-given-I}, $Y_i$ can be expressed as 
\begin{align*}
    Y_i =  \mat{V}_{Ii:}^{\top}\mat{\theta}^*_I + \epsilon_{Ii}^*, \quad \epsilon_{Ii}^* \sim \cN(0, \sigma_I^{*2}),
\end{align*}
where $\epsilon_{Ii}^*$ is an error term independent of $\mat{V}_{Ii:}$.
\end{proof}

\begin{lemma}\label{lem:estimator-exp&deriv-given-IJ}
Under the same condition as in \Cref{lem:Y-given-IJ}, let $\hat{\mat{\theta}}_I$ be the fitted estimator given by \Cref{prop:convergence}, the following equalities characterizes the conditional expectation of  $\hat{\mat{\theta}}_I$ given $\mat{V}_I, \mat{V}_J$ and its derivative with respect to $\rho$ evaluated at $\rho =0$:
\begin{align*}
    \E\paren*{\hat{\mat{\theta}}_I \mid \mat{V}_I, \mat{V}_J} \mid_{\rho=0} = \bar{\mat{\theta}}^*_I+\mat{V}^{-1}_{II}\mat{V}^{\mathstrut}_{IJ}\bar{\mat{\theta}}^*_J, 
\end{align*}
where $\bar{\mat{\theta}}^*_I$ and $\bar{\mat{\theta}}^*_J$ are the constants given by \Cref{lem:Y-given-IJ}. Furthermore
\begin{align*}
     \frac{d}{d\rho} \cbrac*{\E\paren*{\hat{\mat{\theta}}_{I}\mid \mat{V}_I, \mat{V}_J}}  \mid_{\rho=0}
     = \mat{V}^{-1}_{II} \mat{V}^{\mathstrut}_{IJ} \mat{V}^{-1}_{JJ}\mat{V}^{\mathstrut}_{JI}\bar{\mat{\theta}}^*_I-\bar{\mat{\theta}}^*_I.
\end{align*}
\end{lemma}

\begin{proof}
Recall from \Cref{prop:convergence}, the expression of estimator $\hat{\mat{\theta}}_{I}$ can be expanded as
\begin{align*}
    \hat{\mat{\theta}}_I = (1-\rho)\mat{V}^{-1}_{II}\mat{V}_I^{\top} \mat{Y} + \rho\mat{V}^{-1}_{II}\mat{V}^{\mathstrut}_{IJ}\mat{V}^{-1}_{JJ}\mat{V}_J^{\top} \mat{Y}+\mat{H}(\rho),
\end{align*}
where $\mat{H}(\rho)=\sum_{n=2}^{\infty}\rho^n \mat{G}_n$, for some $\mat{G}_n \in \mathbb{R}^{p_I}$ independent of $\rho$.
Together with \Cref{lem:Y-given-IJ}, we have 
\begin{align*}
    \E\paren*{\hat{\mat{\theta}}_I \mid \mat{V}_I, \mat{V}_J} \mid_{\rho=0}
    &=\E\paren*{\mat{V}^{-1}_{II}\mat{V}_I^{\top} \mat{Y}\mid \mat{V}_I, \mat{V}_J}\\
    &= \mat{V}^{-1}_{II}\mat{V}_I^{\top} \E\paren*{\mat{Y}\mid \mat{V}_I, \mat{V}_J}\\
    &= \mat{V}^{-1}_{II}\mat{V}_I^{\top} (\mat{V}_{I}\bar{\mat{\theta}}^*_I + \mat{V}_{J}\bar{\mat{\theta}}^*_J)\\
    &= \mat{V}^{-1}_{II}\mat{V}^{\mathstrut}_{II}\bar{\mat{\theta}}^*_I + \mat{V}^{-1}_{II}\mat{V}^{\mathstrut}_{IJ}\bar{\mat{\theta}}^*_J\\
    &= \bar{\mat{\theta}}^*_I+\mat{V}^{-1}_{II}\mat{V}^{\mathstrut}_{IJ}\bar{\mat{\theta}}^*_J
\end{align*}
where $\bar{\mat{\theta}}^*_I$ and $\bar{\mat{\theta}}^*_J$ are constant vectors given by \Cref{lem:Y-given-IJ}.

Furthermore, by Leibniz integral rule, we have
\begin{align*}
    &\frac{d}{d\rho} \cbrac*{\E\paren*{\hat{\mat{\theta}}_{I}\mid \mat{V}_I, \mat{V}_J}}  \mid_{\rho=0}\\
    &=\E\brac*{ \frac{d}{d\rho} \cbrac*{(1-\rho)\mat{V}^{-1}_{II}\mat{V}_I^{\top} \mat{Y} + \rho\mat{V}^{-1}_{II}\mat{V}^{\mathstrut}_{IJ}\mat{V}^{-1}_{JJ}\mat{V}_J^{\top} \mat{Y}+\mat{H}(\rho) } \mid_{\rho=0} \mid \mat{V}_I, \mat{V}_J}\\
    &= \E\paren*{ \mat{V}^{-1}_{II}\mat{V}^{\mathstrut}_{IJ}\mat{V}^{-1}_{JJ}\mat{V}_J^{\top} \mat{Y}-\mat{V}^{-1}_{II}\mat{V}_I^{\top} \mat{Y} \mid \mat{V}_I, \mat{V}_J}\\
    &= \mat{V}^{-1}_{II}\mat{V}^{\mathstrut}_{IJ}\mat{V}^{-1}_{JJ}\mat{V}_J^{\top}(\mat{V}_{I}\bar{\mat{\theta}}^*_I + \mat{V}_{J}\bar{\mat{\theta}}^*_J)-\mat{V}^{-1}_{II}\mat{V}_I^{\top}(\mat{V}_{I}\bar{\mat{\theta}}^*_I + \mat{V}_{J}\bar{\mat{\theta}}^*_J)\\
    &= \mat{V}^{-1}_{II} \mat{V}^{\mathstrut}_{IJ} \mat{V}^{-1}_{JJ}\mat{V}^{\mathstrut}_{JI}\bar{\mat{\theta}}^*_I-\bar{\mat{\theta}}^*_I
\end{align*}
\end{proof}

\begin{lemma}\label{lem:xi-negative}
Under \Cref{cond:related-latent-rep,,cond:orthogonal}, the following inequality holds.
\begin{align*}
    \Xi \coloneqq \frac{2\bar{\sigma}^{*2}}{n}\sum_{m=1}^{p_I}\paren*{\sum_{k=1}^{p_J}\frac{(\mat{T}_{J:k}^{\top}\mat{T}_{I:m}^{\mathstrut})^2}{(\mat{T}_{I:m}^{\top}\mat{T}_{I:m}^{\mathstrut}+\sigma_I^2)(\mat{T}_{J:k}^{\top}\mat{T}_{J:k}^{\mathstrut}+\sigma_J^2)}-1} <0,
\end{align*}
where $\bar{\sigma}^{*2}>0$ is a constant as defined in \Cref{lem:Y-given-IJ}.
\end{lemma}
\begin{proof}
Since $\bar{\sigma}^{*2}>0$, it suffices to show that for $\forall m \in [p_I]$,
\begin{align*}
    \sum_{k=1}^{p_J} \frac{(\mat{T}_{J:k}^{\top}\mat{T}_{I:m}^{\mathstrut})^2}{(\mat{T}_{I:m}^{\top}\mat{T}_{I:m}^{\mathstrut}+\sigma_I^2)(\mat{T}_{J:k}^{\top}\mat{T}_{J:k}^{\mathstrut}+\sigma_J^2)} < 1.
\end{align*}
For $k \in [p_J]$, let $\Tilde{\mat{T}}_{J:k} \coloneqq \mat{T}_{J:k}/\norm{\mat{T}_{J:k}}$ be the normalized columns of $\mat{T}_{J}$,  and let $\Tilde{\mat{T}}_{J}$ denote the corresponding normalized matrix. Similarly, For $m \in [p_I]$, let $\Tilde{\mat{T}}_{I:m} \coloneqq \mat{T}_{I:m}/\norm{\mat{T}_{I:m}}$ be the normalized columns of $\mat{T}_{I}$, then
\begin{align*}
    \sum_{k=1}^{p_J} \frac{(\mat{T}_{J:k}^{\top}\mat{T}_{I:m}^{\mathstrut})^2}{(\mat{T}_{I:m}^{\top}\mat{T}_{I:m}^{\mathstrut}+\sigma_I^2)(\mat{T}_{J:k}^{\top}\mat{T}_{J:k}^{\mathstrut}+\sigma_J^2)} 
    &< \sum_{k=1}^{p_J} \frac{(\mat{T}_{J:k}^{\top}\mat{T}_{I:m}^{\mathstrut})^2}{(\mat{T}_{I:m}^{\top}\mat{T}_{I:m}^{\mathstrut})(\mat{T}_{J:k}^{\top}\mat{T}_{J:k}^{\mathstrut})}\\
    &= \sum_{k=1}^{p_J} \Tilde{\mat{T}}_{J:k}^{\top}\Tilde{\mat{T}}_{I:m}^{\mathstrut}\\
    &= \norm{\Tilde{\mat{T}}_{J}^{\top}\Tilde{\mat{T}}_{I:m}^{\mathstrut}}^2\\
    &= \Tilde{\mat{T}}_{I:m}^{\top}\Tilde{\mat{T}}_{J}^{\mathstrut}\Tilde{\mat{T}}_{J}^{\top}\Tilde{\mat{T}}_{I:m}^{\mathstrut}.
\end{align*}
By \Cref{cond:orthogonal}, $p_I, p_J \leq p$ and $\Tilde{\mat{T}}_{J}$ has orthonormal columns. It follows that, $\Tilde{\mat{T}}_{J}$ has a singular value decomposition with $\Tilde{\mat{T}}_{J}=\Tilde{\mat{U}}\Tilde{\mat{D}}\Tilde{\mat{V}}^{\top}$, where $\Tilde{\mat{U}} \in \mathbb{R}^{p\times p}$, $\Tilde{\mat{V}} \in \mathbb{R}^{p_J\times p_J}$ are orthogonal matrices and $\Tilde{\mat{D}} \in \mathbb{R}^{p\times p_J}$ is a rectangular diagonal matrix whose diagonal entries are all equal to $1$. Therefore,
\begin{align*}
     \sum_{k=1}^{p_J} \frac{(\mat{T}_{J:k}^{\top}\mat{T}_{I:m}^{\mathstrut})^2}{(\mat{T}_{I:m}^{\top}\mat{T}_{I:m}^{\mathstrut}+\sigma_I^2)(\mat{T}_{J:k}^{\top}\mat{T}_{J:k}^{\mathstrut}+\sigma_J^2)}
     &< \Tilde{\mat{T}}_{I:m}^{\top}\Tilde{\mat{T}}_{J}^{\mathstrut}\Tilde{\mat{T}}_{J}^{\top}\Tilde{\mat{T}}_{I:m}^{\mathstrut}\\
     &= \Tilde{\mat{T}}_{I:m}^{\top} \Tilde{\mat{U}}\Tilde{\mat{D}}\Tilde{\mat{V}} ^{\top}\Tilde{\mat{V}}\Tilde{\mat{D}}^{\top}\Tilde{\mat{U}}^{\top}\Tilde{\mat{T}}_{I:m}^{\mathstrut}\\
     &= \Tilde{\mat{T}}_{I:m}^{\top} \Tilde{\mat{U}}\Tilde{\mat{D}}\Tilde{\mat{D}}^{\top}\Tilde{\mat{U}}^{\top}\Tilde{\mat{T}}_{I:m}^{\mathstrut}\\
     &\leq \Tilde{\mat{T}}_{I:m}^{\top} \Tilde{\mat{U}}\Tilde{\mat{U}}^{\top}\Tilde{\mat{T}}_{I:m}^{\mathstrut}\\
     &= \norm{\Tilde{\mat{T}}_{I:m}}^2=1,
\end{align*}
which concludes the proof.
\end{proof}

\begin{lemma}\label{lem:oracle-coeff-relation}
Under the same condition as in \Cref{lem:Y-given-IJ}, the following equality holds.
\begin{align*}
\Tilde{\mat{\Sigma}}_I^{-1}\mat{T}^{\mathstrut}_{IJ}\bar{\mat{\theta}}^*_J-\paren*{\mat{\theta}^*_{I} -\bar{\mat{\theta}}^*_I} =\mat{0},
\end{align*}
where $\mat{\theta}^*_{I}$ is a constant defined in \Cref{lem:Y-given-I}, and $\bar{\mat{\theta}}^*_I$, $\bar{\mat{\theta}}^*_J$ are constants given by \Cref{lem:Y-given-IJ}.
\end{lemma}
\begin{proof}
Recall from \Cref{lem:Y-given-IJ} and \Cref{lem:Y-given-I}, 
\begin{align*}
    \mat{\theta}^*_I&=\Tilde{\mat{\Sigma}}^{-1}_I\mat{T}_I^{\top} \mat{\theta},\\
    \bar{\mat{\theta}}^*_I
    &=(\mat{A}^{\top}\mat{T}_I^{\top}+\mat{C}^{\top}\mat{T}_J^{\top})\mat{\theta}, \\
    \bar{\mat{\theta}}^*_J 
    &= (\mat{B}^{\top}\mat{T}_I^{\top}+\mat{D}^{\top}\mat{T}_J^{\top})\mat{\theta},
\end{align*}
where $\mat{A}\in \mathbb{R}^{p_I\times p_I}$, $\mat{B}\in \mathbb{R}^{p_I\times p_J}$, $\mat{C}\in \mathbb{R}^{p_J\times p_I}$ and $\mat{D}\in \mathbb{R}^{p_J\times p_J}$ satisfies
\begin{align*}
     \begin{pmatrix}
        \mat{A} & \mat{B}\\
        \mat{C} & \mat{D}
    \end{pmatrix} 
    =\begin{pmatrix}
        \Tilde{\mat{\Sigma}}_I & \mat{T}^{\mathstrut}_{IJ}\\
        \mat{T}^{\mathstrut}_{JI} & \Tilde{\mat{\Sigma}}_J
    \end{pmatrix}^{-1}.
\end{align*}
Then we must have that
\begin{align*}
    \begin{pmatrix}
        \mat{A} & \mat{B}\\
        \mat{C} & \mat{D}
    \end{pmatrix}
    \begin{pmatrix}
        \Tilde{\mat{\Sigma}}_I & \mat{T}^{\mathstrut}_{IJ}\\
        \mat{T}^{\mathstrut}_{JI} & \Tilde{\mat{\Sigma}}_J
    \end{pmatrix}
    =\begin{pmatrix}
        \mat{A}\Tilde{\mat{\Sigma}}_I+ \mat{B} \mat{T}^{\mathstrut}_{JI}& \mat{A}\mat{T}^{\mathstrut}_{IJ} +\mat{B}\Tilde{\mat{\Sigma}}_J\\
        \mat{C}\Tilde{\mat{\Sigma}}_I+\mat{D}\mat{T}^{\mathstrut}_{JI}& \mat{C}\mat{T}^{\mathstrut}_{IJ}+\mat{D}\Tilde{\mat{\Sigma}}_J
    \end{pmatrix}
    =\begin{pmatrix}
        \mat{\Sigma}_{I} & \mat{0}\\
        \mat{0} & \mat{\Sigma}_{J}
    \end{pmatrix},
\end{align*}
where $\mat{\Sigma}_{I} \in \mathbb{R}^{p_I\times p_I}$, $\mat{\Sigma}_{J}\in \mathbb{R}^{p_J\times p_J}$ are identity matrices, and $\mat{0}$ represents rectangular block matrix of $0$s. It implies that,
\begin{alignat*}{2}
&& (\mat{\Sigma}_{I}-\Tilde{\mat{\Sigma}}_I\mat{A}^{\top}- \mat{T}^{\mathstrut}_{IJ}\mat{B}^{\top})\mat{T}^{\top}_{I} 
&=(\Tilde{\mat{\Sigma}}_I\mat{C}^{\top}+\mat{T}^{\mathstrut}_{IJ}\mat{D}^{\top})\mat{T}^{\top}_{J},\\
&\implies \qquad
&\mat{T}^{\top}_{I}-\Tilde{\mat{\Sigma}}_I\mat{A}^{\top}\mat{T}^{\top}_{I}-\mat{T}^{\mathstrut}_{IJ}\mat{B}^{\top}\mat{T}^{\top}_{I}
&=\Tilde{\mat{\Sigma}}_I\mat{C}^{\top}\mat{T}^{\top}_{J}+\mat{T}^{\mathstrut}_{IJ}\mat{D}^{\top}\mat{T}^{\top}_{J}\\
&\implies \qquad
&\mat{T}^{\top}_{I}-\mat{T}^{\mathstrut}_{IJ}(\mat{B}^{\top}\mat{T}^{\top}_{I}+\mat{D}^{\top}\mat{T}^{\top}_{J}) 
&= \Tilde{\mat{\Sigma}}_I(\mat{A}^{\top}\mat{T}^{\top}_{I}+\mat{C}^{\top}\mat{T}^{\top}_{J})\\
&\implies \qquad
&\Tilde{\mat{\Sigma}}_I^{-1}\mat{T}^{\top}_{I}-\Tilde{\mat{\Sigma}}_I^{-1}\mat{T}^{\mathstrut}_{IJ}(\mat{B}^{\top}\mat{T}^{\top}_{I}+\mat{D}^{\top}\mat{T}^{\top}_{J}) 
&=\mat{A}^{\top}\mat{T}^{\top}_{I}+\mat{C}^{\top}\mat{T}^{\top}_{J}\\
&\implies \qquad
&\Tilde{\mat{\Sigma}}_I^{-1}\mat{T}^{\mathstrut}_{IJ}(\mat{B}^{\top}\mat{T}^{\top}_{I}+\mat{D}^{\top}\mat{T}^{\top}_{J})+(\mat{A}^{\top}\mat{T}^{\top}_{I}+\mat{C}^{\top}\mat{T}^{\top}_{J})
&= \Tilde{\mat{\Sigma}}_I^{-1}\mat{T}^{\top}_{I}\\
&\implies \qquad
&\Tilde{\mat{\Sigma}}_I^{-1}\mat{T}^{\mathstrut}_{IJ}(\mat{B}^{\top}\mat{T}^{\top}_{I}+\mat{D}^{\top}\mat{T}^{\top}_{J})\mat{\theta}+(\mat{A}^{\top}\mat{T}^{\top}_{I}+\mat{C}^{\top}\mat{T}^{\top}_{J})\mat{\theta}
&= \Tilde{\mat{\Sigma}}_I^{-1}\mat{T}^{\top}_{I}\mat{\theta}\\
&\implies \qquad
&\Tilde{\mat{\Sigma}}_I^{-1}\mat{T}^{\mathstrut}_{IJ} \bar{\mat{\theta}}^*_J+\bar{\mat{\theta}}^*_I
&=\mat{\theta}^*_I,
\end{alignat*}
and rearranging the last equality concludes the proof.
\end{proof}

\subsection{Proofs of Main Results}\label{app:proofs-main}

\begin{proof}[Proof of \Cref{prop:convergence}]
We begin to write down the explicit expression for the fitted parameters $\hat{\mat{\theta}}_{I}, \hat{\mat{\theta}}_{J}$, let
\begin{align*}
\Tilde{\mat{\theta}}=
    \begin{pmatrix}
        \mat{\theta}_I\\
        \mat{\theta}_J
    \end{pmatrix},
\Tilde{\mat{V}}=
    \begin{pmatrix}
        \mat{V}_I & \mat{0} \\
        \mat{0} & \mat{V}_J \\
        \sqrt{\rho}\mat{V}_I & -\sqrt{\rho}\mat{V}_J
    \end{pmatrix}, 
    \Tilde{\mat{Y}} =
    \begin{pmatrix}
        \mat{Y}\\
        \mat{Y}\\
        \mat{0}
    \end{pmatrix},
\end{align*}
where $\mat{0}$ is boldfaced to indicate that it is a zero matrix. Its dimension is left unspecified unless needed for clarity. Note that solving \eqref{eq:mse-overall-loss} is equivalent to solving
\begin{align*}
   \hat{\mat{\theta}} = \argmin_{\Tilde{\mat{\theta}}}\norm{\Tilde{\mat{Y}}-\Tilde{\mat{V}}\Tilde{\mat{\theta}}}^2,
\end{align*}
when $n>\max\cbrac{p_I, p_J}$, the equation above has solution in the form of:
\begin{align}\label{eq:ols-formula}
\hat{\mat{\theta}}=
    \begin{pmatrix}
        \hat{\mat{\theta}}_I\\
        \hat{\mat{\theta}}_J
    \end{pmatrix}=
    \paren*{\Tilde{\mat{V}}^{\top}\Tilde{\mat{V}}}^{-1} \paren*{\Tilde{\mat{V}}^{\top}\Tilde{\mat{Y}}},
\end{align}
where 
\begin{equation}\label{eq:VV}
\begin{aligned}
\Tilde{\mat{V}}^{\top}\Tilde{\mat{V}} 
    &=
    \begin{pmatrix}
        \mat{V}_I^{\top} & \mat{0} & \sqrt{\rho}\mat{V}_I^{\top}\\
        \mat{0} & \mat{V}_J^{\top} & -\sqrt{\rho}\mat{V}_J^{\top}
    \end{pmatrix}
    \begin{pmatrix}
        \mat{V}_I & \mat{0} \\
        \mat{0} & \mat{V}_J \\
        \sqrt{\rho}\mat{V}_I & -\sqrt{\rho}\mat{V}_J
    \end{pmatrix}\\
    &=
    \begin{pmatrix}
        (1+\rho)\mat{V}^{\mathstrut}_{II} & -\rho \mat{V}^{\mathstrut}_{IJ} \\
        -\rho \mat{V}^{\mathstrut}_{JI} & (1+\rho) \mat{V}^{\mathstrut}_{JJ}
    \end{pmatrix},
\end{aligned}
\end{equation}

and 
\begin{align}\label{eq:VY}
    \Tilde{\mat{V}}^{\top}\Tilde{\mat{Y}} = \begin{pmatrix}
        \mat{V}_I^{\top} & \mat{0} & \sqrt{\rho}\mat{V}_I^{\top}\\
        \mat{0} & \mat{V}_J^{\top} & -\sqrt{\rho}\mat{V}_J^{\top}
    \end{pmatrix} 
    \begin{pmatrix}
        \mat{Y}\\
        \mat{Y}\\
        \mat{0}
    \end{pmatrix} = 
    \begin{pmatrix}
        \mat{V}_I^{\top} \mat{Y}\\
        \mat{V}_J^{\top} \mat{Y}
    \end{pmatrix}.
\end{align}

We first establish the invertibility of  $\Tilde{\mat{V}}^{\top}\Tilde{\mat{V}} \in \mathbb{R}^{(p_I+p_J)\times(p_I+p_J)}$, then derive a power series expansion of the inverse matrix. 

First note that $\Tilde{\mat{V}}^{\top}\Tilde{\mat{V}}$ in \eqref{eq:VV} can be rearranged into 
\begin{equation}\label{eq:VV-rearrange}  
\begin{aligned}
    \Tilde{\mat{V}}^{\top}\Tilde{\mat{V}} 
    &=
    \begin{pmatrix}
        (1+\rho)\mat{V}^{\mathstrut}_{II} & -\rho \mat{V}^{\mathstrut}_{IJ} \\
        -\rho \mat{V}^{\mathstrut}_{JI} & (1+\rho) \mat{V}^{\mathstrut}_{JJ}
    \end{pmatrix}\\
    &= \underbrace{\begin{pmatrix}
        \mat{V}^{\mathstrut}_{II} & 0 \\
        0 & \mat{V}^{\mathstrut}_{JJ}
    \end{pmatrix}}_{\coloneqq \mat{E}}
    -\rho \underbrace{\begin{pmatrix}
        -\mat{V}^{\mathstrut}_{II} & \mat{V}^{\mathstrut}_{IJ} \\
        \mat{V}^{\mathstrut}_{JI} & -\mat{V}^{\mathstrut}_{JJ}
    \end{pmatrix}}_{\coloneqq \mat{F}},
\end{aligned}
\end{equation}

which isolates the effect of the $\rho$. Equivalently, we have $\Tilde{\mat{V}}^{\top}\Tilde{\mat{V}} =\mat{E} + \rho(-\mat{F})$. One can easily show that $-\mat{F}$ is positive semidefinite and $\mat{E}$ is positive definite when $\mat{V}^{\mathstrut}_{II}$ and $\mat{V}^{\mathstrut}_{JJ}$ are both of full rank. Under \Cref{cond:related-latent-rep}, $\mat{V}^{\mathstrut}_{II}$ and $\mat{V}^{\mathstrut}_{JJ}$ follow Wishart distribution and are invertible if $n\geq p_I$ and $n \geq p_J$. By definition, $\rho \geq 0$, hence, $\Tilde{\mat{V}}^{\top}\Tilde{\mat{V}} =\mat{E} + \rho(-\mat{F})$ is positive definite when $n\geq \max\cbrac{p_I, p_J}$.

The inverse of $\Tilde{\mat{V}}^{\top}\Tilde{\mat{V}} \in \mathbb{R}^{(p_I+p_J)\times(p_I+p_J)}$ has a nontrivial analytical solution in high dimensional case.
While theoretically, it is possible to derive a closed form of $\Tilde{\mat{V}}^{\top}\Tilde{\mat{V}}^{-1}$ by applying the general formula of inverting $2\times2$ block matrices~\citep{Lu-2002-inverse-block-matrix}, the resulting inverse matrix will be a complex expression in terms of $\rho$ which makes it nearly impossible to conduct the subsequent analysis. Therefore, we derive a power series expansion of $\Tilde{\mat{V}}^{\top}\Tilde{\mat{V}}$ below.

By Neumann series, inverse of $\Tilde{\mat{V}}^{\top}\Tilde{\mat{V}}$ can be expanded as a power series
\begin{equation}\label{eq:neumann-expansion}
\begin{aligned}
    (\Tilde{\mat{V}}^{\top}\Tilde{\mat{V}})^{-1}
    &=(\mat{E} - \rho\mat{F})^{-1}\\
    &=(\mat{I}-\rho \mat{E}^{-1}\mat{F})^{-1}\mat{E}^{-1}\\
    &=\cbrac*{\sum_{k=0}^{\infty} \paren*{\rho\mat{E}^{-1}\mat{F}}^k}\mat{E}^{-1}\\
    &= \mat{E}^{-1} +\rho \mat{E}^{-1}\mat{F}\mat{E}^{-1} + \rho^2 \mat{E}^{-1}\mat{F}\mat{E}^{-1}\mat{F}\mat{E}^{-1} +\mat{H}(\rho)
\end{aligned}
\end{equation}
where $\mat{H}(\rho)$ is the higher order terms of $\rho$. Since we will be analyzing how the generalization error changes around $\rho =0$, we focus on the quadratic approximation which dominates the power series. The benefits of the expansion above is that it bypasses the needs of computing the explicit inverse of a general block matrix and we only need to compute the inverse of a sparse matrix $\mat{E}$, which is simply
\begin{equation}\label{eq:inv-A-diagonal}
\begin{aligned}
    \mat{E}^{-1}
    =\begin{pmatrix}
        \mat{V}^{\mathstrut}_{II} & 0 \\
        0 & \mat{V}^{\mathstrut}_{JJ}
    \end{pmatrix}^{-1} 
    = \begin{pmatrix}
        \mat{V}^{-1}_{II} & 0 \\
        0 & \mat{V}^{-1}_{JJ}
    \end{pmatrix}
\end{aligned}
\end{equation}

Plugging in \eqref{eq:inv-A-diagonal} into \eqref{eq:neumann-expansion}, we can show with direct computation that the estimator is 
\begin{align*}
    \hat{\mat{\theta}}_I = (1-\rho)\mat{V}^{-1}_{II}\mat{V}_I^{\top} \mat{Y} + \rho\mat{V}_{II}^{-1}\mat{V}^{\mathstrut}_{IJ}\mat{V}_{JJ}^{-1}\mat{V}_J^{\top} \mat{Y}+\mat{H}(\rho),
\end{align*}
where $\mat{H}(\rho)=\sum_{n=2}^{\infty}\rho^n \mat{G}_n$, for some $\mat{G}_n \in \mathbb{R}^{p_I}$ independent of $\rho$.
\end{proof}

\begin{proof}[Proof of \Cref{thm:gen-error-general}]
Recall that $Y_*$ is the ground-truth label of a new test point and $\mat{V}_{I*}$ is the corresponding fused representation for student $I$. We begin to examine the MSE of the individual student model. By \Cref{lem:Y-given-I}, we have 
\begin{align}\label{eq:tmp-mse}
    \text{MSE}(I;\rho) 
    = \E\cbrac*{(Y_*-\mat{V}^{\top}_{I*} \hat{\mat{\theta}}_{I})^2 \mid \mat{V}_I} = \E\brac*{\cbrac*{\mat{V}^{\top}_{I*}(\mat{\theta}^*_I - \hat{\mat{\theta}}_{I})}^2 \mid \mat{V}_I} + \sigma_I^{*2}.
\end{align}
The first term in the last equality is
\begin{align*}
    &\E\brac*{\cbrac*{\mat{V}^{\top}_{I*}(\mat{\theta}^*_I - \hat{\mat{\theta}}_{I})}^2 \mid \mat{V}_I} \\
    &=\E\brac*{ \cbrac*{\sum_{i=1}^{p_I} V_{I*i} (\theta^*_{Ii} - \hat{\theta}_{Ii})}^2 \,\middle\vert\, \mat{V}_I}\\
    &= \sum_{i=1}^{p_I} \E\cbrac*{ V_{I*i}^2 (\theta^*_{Ii} - \hat{\theta}_{Ii})^2 \mid \mat{V}_I} + 2\sum_{1\leq i <j \leq p_I} \E\cbrac*{V_{I*i}V_{I*j} (\theta^*_{Ii} - \hat{\theta}_{Ii})(\theta^*_{Ij} - \hat{\theta}_{Ij}) \mid \mat{V}_I}\\
    &= \sum_{i=1}^{p_I} \E\paren*{ V_{I*i}^2}\E\cbrac*{ (\theta^*_{Ii} - \hat{\theta}_{Ii})^2 \mid \mat{V}_I} + 2\sum_{1\leq i <j \leq p_I} \E\paren*{V_{I*i}V_{I*j} }\E\cbrac*{(\theta^*_{Ii} - \hat{\theta}_{Ii})(\theta^*_{Ij} - \hat{\theta}_{Ij}) \mid \mat{V}_I},
\end{align*}
by \Cref{cond:orthogonal} which implies features are mutually independent, the second term in the last equality vanishes, together with \eqref{eq:joint-distribution-all} in \Cref{lem:Y-given-IJ}, we have
\begin{equation}\label{eq:tmp-decomp}
\begin{aligned}
    &\E\brac*{\cbrac*{\mat{V}^{\top}_{I*}(\mat{\theta}^*_I - \hat{\mat{\theta}}_{I})}^2 \mid \mat{V}_I} \\
    &= \sum_{i=1}^{p_I} \E\paren*{ V_{I*i}^2}\E\cbrac*{ (\theta^*_{Ii} - \hat{\theta}_{Ii})^2 \mid \mat{V}_I}\\
    &= \sum_{i=1}^{p_I} \E\paren*{ V_{I*i}^2}\Var\cbrac*{ (\theta^*_{Ii} - \hat{\theta}_{Ii})^2 \mid \mat{V}_I} +\sum_{i=1}^{p_I} \E\paren*{ V_{I*i}^2}\cbrac*{\E \paren*{\theta^*_{Ii} - \hat{\theta}_{Ii} \mid \mat{V}_I}}^2\\
    &= \underbrace{\mathrm{diag}\paren*{\Tilde{\mat{\Sigma}}_I}^{\top} \Var^{\circ}\paren*{\hat{\mat{\theta}}_{I}\mid \mat{V}_I}}_{\coloneqq V(\mat{V}_I; \rho)}+\underbrace{\mathrm{diag}\paren*{\Tilde{\mat{\Sigma}}_I}^{\top} \cbrac*{\E\paren*{\mat{\theta}^*_I - \hat{\mat{\theta}}_{I} \mid \mat{V}_I}}^{\circ2}}_{\coloneqq B^2(\mat{V}_I; \rho)},
\end{aligned}
\end{equation}
where for vector $\mat{x}=(x_1, \dots, x_n)$, $\Var^{\circ}(\mat{x})=(\Var(x_1), \dots, \Var(x_n))$ denotes the element-wise variance and $\mat{x}^{\circ2}=(x^2_1, \dots, x^2_n)$ denotes the element-wise square. 

Combining \eqref{eq:tmp-mse} and \eqref{eq:tmp-decomp}, we have that
\begin{align}\label{eq:mse-initial-decomp}
    \text{MSE}(I;\rho) 
    = V(\mat{V}_I; \rho) + B^2(\mat{V}_I; \rho) + \sigma_I^{*2},
\end{align}
where $V(\mat{V}_I; \rho)$ is the variance related term and $ B^2(\mat{V}_I; \rho)$ is the bias related term.

We first inspect how $ B^2(\mat{V}_I; \rho)$ change with respect to $\rho$,
\begin{align*}
    \frac{d}{d\rho}B^2(\mat{V}_I; \rho)
    &= \frac{d}{d\rho}\brac*{\mathrm{diag}\paren*{\Tilde{\mat{\Sigma}}_I}^{\top}\cbrac*{\E\paren*{\mat{\theta}^*_I - \hat{\mat{\theta}}_{I} \mid \mat{V}_I}}^{\circ2}}\\
    &= 2\mathrm{diag}\paren*{\Tilde{\mat{\Sigma}}_I}^{\top} \brac*{ \E\cbrac*{ (\mat{\theta}^*_I - \hat{\mat{\theta}}_{I}) \mid \mat{V}_I} \circ \frac{d}{d\rho}\E\cbrac*{ (\mat{\theta}^*_I - \hat{\mat{\theta}}_{I}) \mid \mat{V}_I}},
\end{align*}
where $\mat{x}\circ \mat{y}$ denotes the Hadamard product of two vectors $\mat{x}$ and $\mat{y}$. Recall from \Cref{prop:convergence}, the fitted estimator $\hat{\mat{\theta}}_{I}$ is
\begin{align*}
    \hat{\mat{\theta}}_I = (1-\rho)\mat{V}^{-1}_{II}\mat{V}_I^{\top} \mat{Y} + \rho\mat{V}_{II}^{-1}\mat{V}^{\mathstrut}_{IJ}\mat{V}_{JJ}^{-1}\mat{V}_J^{\top} \mat{Y}+\mat{H}(\rho),
\end{align*}
where $\mat{H}(\rho)=\sum_{n=2}^{\infty}\rho^n \mat{G}_n$, for some $\mat{G}_n \in \mathbb{R}^{p_I}$ independent of $\rho$. From \Cref{lem:Y-given-I}, we have
\begin{equation}\label{eq:exp-estimator-I}
\begin{aligned}
    \E\paren*{ \mat{\theta}^*_I-\hat{\mat{\theta}}_{I} \mid \mat{V}_I} \mid_{\rho=0} 
    &= \mat{\theta}^*_I-\E\paren*{\mat{V}^{-1}_{II}\mat{V}_I^{\top} \mat{Y} \mid \mat{V}_I}\\
    &= \mat{\theta}^*_I-\mat{V}^{-1}_{II}\mat{V}_I^{\top} \E\paren*{\mat{Y} \mid \mat{V}_I}\\
    &= \mat{\theta}^*_I-\mat{V}^{-1}_{II}\mat{V}^{\mathstrut}_{II}\mat{\theta}^*_I = \mat{0}.
\end{aligned}
\end{equation}
It follows that
\begin{equation}\label{eq:deriv-bias}  
\begin{aligned}
    &\frac{d}{d\rho}B^2(\mat{V}_I; \rho) \mid_{\rho=0}\\
    &=2\mathrm{diag}\paren*{\Tilde{\mat{\Sigma}}_I}^{\top} \brac*{ \E\cbrac*{ (\mat{\theta}^*_I - \hat{\mat{\theta}}_{I}) \mid \mat{V}_I}\mid_{\rho=0}  \circ \frac{d}{d\rho}\E\cbrac*{ (\mat{\theta}^*_I - \hat{\mat{\theta}}_{I}) \mid \mat{V}_I}\mid_{\rho=0} } = 0,
\end{aligned}
\end{equation}
which implies that an increase of the penalty $\rho$ around $0$ does not effect the bias term. Next we examine the variance term, by law of total variance, we have 
\begin{equation}\label{eq:cond-varI-decomp}
\begin{aligned}
    V(\mat{V}_I; \rho)
    &=\mathrm{diag}\paren*{\Tilde{\mat{\Sigma}}_I}^{\top} \Var^{\circ}\paren*{\hat{\mat{\theta}}_{I}\mid \mat{V}_I} \\
    &= \underbrace{\mathrm{diag}\paren*{\Tilde{\mat{\Sigma}}_I}^{\top} \E \cbrac*{\Var^{\circ}\paren{\hat{\mat{\theta}}_{I} \mid \mat{V}_I, \mat{V}_J} \mid \mat{V}_I}}_{V_a(\mat{V}_I; \rho)} + \underbrace{\mathrm{diag}\paren*{\Tilde{\mat{\Sigma}}_I}^{\top} \Var^{\circ}\cbrac*{\E \paren{\hat{\mat{\theta}}_{I}\mid \mat{V}_I, \mat{V}_J} \mid \mat{V}_I}}_{V_e(\mat{V}_I; \rho)},
\end{aligned}
\end{equation}
where the first term is the aleatoric variance and the second term is the epistemic variance. 

We first examine the aleatoric variance in \eqref{eq:cond-varI-decomp}. Let
\begin{align*}
    \mat{M} \coloneqq (1-\rho)\mat{V}^{-1}_{II}\mat{V}_I^{\top} + \rho\mat{V}_{II}^{-1}\mat{V}^{\mathstrut}_{IJ}\mat{V}_{JJ}^{-1}\mat{V}_J^{\top} + \mat{H}_1(\rho),
\end{align*}
where $\mat{H}_1(\rho) \in \mathbb{R}^{p_I\times n}$ is the higher order term of $\rho$. From the from of $\hat{\mat{\theta}}_{I}$ and the conditional distribution of $\mat{Y}$ given $\mat{V}_I, \mat{V}_J$, as characterized in \Cref{lem:Y-given-IJ}, we have 
\begin{align*}
    \Var^{\circ}\paren*{\hat{\mat{\theta}}_{I} \mid \mat{V}_I, \mat{V}_J} 
    &=  \Var^{\circ}\paren*{\mat{M}\mat{Y} \mid \mat{V}_I, \mat{V}_J} \\
    &= \bar{\sigma}^{*2} \mathrm{diag}(\mat{M}\mat{M}^{\top})\\
    &= \bar{\sigma}^{*2} \mathrm{diag}\paren*{\mat{V}^{-1}_{II}+2\rho(\mat{V}_{II}^{-1}\mat{V}^{\mathstrut}_{IJ}\mat{V}_{JJ}^{-1}\mat{V}^{\mathstrut}_{JI}\mat{V}_{II}^{-1}-\mat{V}^{-1}_{II})+\mat{H}_2(\rho)},
\end{align*}
where $\mat{H}_2(\rho) \in \mathbb{R}^{p_I\times p_I}$ contains all higher order terms of $\rho$. 
Therefore,
\begin{equation} \label{eq:deriv-var-given-IJ-raw}
\begin{aligned}
    \frac{d}{d\rho} \Var^{\circ}\paren*{\hat{\mat{\theta}}_{I} \mid \mat{V}_I, \mat{V}_J}\mid_{\rho=0} 
    &= 2\bar{\sigma}^{*2} \mathrm{diag}\paren*{\mat{V}_{II}^{-1}\mat{V}^{\mathstrut}_{IJ}\mat{V}_{JJ}^{-1}\mat{V}^{\mathstrut}_{JI}\mat{V}_{II}^{-1}-\mat{V}^{-1}_{II}}\\
    & 2\bar{\sigma}^{*2} \mathrm{diag}\paren*{(\mat{V}_{II}^{-1}\mat{V}^{\mathstrut}_{IJ}\mat{V}_{JJ}^{-1}\mat{V}^{\mathstrut}_{JI}-\mat{\Sigma}_I)\mat{V}_{II}^{-1}},
\end{aligned}
\end{equation}
where recall that $\mat{\Sigma}_I\in \mathbb{R}^{p_I\times p_I}$ is an identity matrix. 

By central limit theorem and \Cref{lem:Y-given-IJ}, components in \eqref{eq:deriv-var-given-IJ-raw} can be approximated as 
\begin{equation}\label{eq:clt}
\begin{aligned}
    \mat{V}^{\mathstrut}_{II} &= n\Tilde{\mat{\Sigma}}_I+\cO_p(\sqrt{n})\mat{1},\\
    \mat{V}^{\mathstrut}_{IJ} &= n\mat{T}^{\mathstrut}_{IJ}+\cO_p(\sqrt{n})\mat{1},\\
    \mat{V}^{\mathstrut}_{JI} &= n\mat{T}^{\mathstrut}_{JI}+\cO_p(\sqrt{n})\mat{1},\\
    \mat{V}^{\mathstrut}_{JJ} &= n\Tilde{\mat{\Sigma}}_J+\cO_p(\sqrt{n})\mat{1},
\end{aligned}
\end{equation}
where $\mat{1}$ denotes a matrix with all elements equal to 1, and the notation $X_n=\cO_p(a_n)$ for a sequence of positive constants $\cbrac{a_n}$ means: $\forall \epsilon >0$, $\exists M>0$ and an integer $N$ such that $\P\paren*{\abs{X_n/a_n}>M}<\epsilon$, $\forall n>N$. 

Plugging in \eqref{eq:clt} into \eqref{eq:deriv-var-given-IJ-raw}, one can show with some straightforward computation that
\begin{align*}  &\mat{V}_{II}^{-1}\mat{V}^{\mathstrut}_{IJ}\mat{V}_{JJ}^{-1}\mat{V}^{\mathstrut}_{JI} =\\
&\begin{pmatrix}
\sum_{k=1}^{p_J}\frac{(\mat{T}_{J:k_{}}^{\top}\mat{T}_{I:1_{}}^{\mathstrut})^2}{(\mat{T}_{I:1_{}}^{\top}\mat{T}_{I:1_{}}^{\mathstrut}+\sigma_I^2)(\mat{T}_{J:k_{}}^{\top}\mat{T}_{J:k_{}}^{\mathstrut}+\sigma_J^2)} & \cdots & \sum_{k=1}^{p_J}\frac{(\mat{T}_{J:k_{}}^{\top}\mat{T}_{I:1_{}}^{\mathstrut})(\mat{T}_{J:k}^{\top}\mat{T}_{I:p_I}^{\mathstrut})}{(\mat{T}_{I:1_{}}^{\top}\mat{T}_{I:1_{}}^{\mathstrut}+\sigma_I^2)(\mat{T}_{J:k_{}}^{\top}\mat{T}_{J:k_{}}^{\mathstrut}+\sigma_J^2)} \\
\sum_{k=1}^{p_J}\frac{(\mat{T}_{J:k_{}}^{\top}\mat{T}_{I:2_{}}^{\mathstrut})(\mat{T}_{J:k_{}}^{\top}\mat{T}_{I:1_{}}^{\mathstrut})}{(\mat{T}_{I:2_{}}^{\top}\mat{T}_{I:2_{}}^{\mathstrut}+\sigma_I^2)(\mat{T}_{J:k}^{\top}\mat{T}_{J:k}^{\mathstrut}+\sigma_J^2)} & \cdots & \sum_{k=1}^{p_J}\frac{(\mat{T}_{J:k_{}}^{\top}\mat{T}_{I:2_{}}^{\mathstrut})(\mat{T}_{J:k}^{\top}\mat{T}_{I:p_I}^{\mathstrut})}{(\mat{T}_{I:2_{}}^{\top}\mat{T}_{I:2_{}}^{\mathstrut}+\sigma_I^2)(\mat{T}_{J:k_{}}^{\top}\mat{T}_{J:k_{}}^{\mathstrut}+\sigma_J^2)}\\
\vdots  & \ddots & \vdots \\
\sum_{k=1}^{p_J}\frac{(\mat{T}_{J:k_{}}^{\top}\mat{T}_{I:p_I}^{\mathstrut})(\mat{T}_{J:k_{}}^{\top}\mat{T}_{I:1_{}}^{\mathstrut})}{(\mat{T}_{I:p_I}^{\top}\mat{T}_{I:p_I}^{\mathstrut}+\sigma_I^2)(\mat{T}_{J:k}^{\top}\mat{T}_{J:k}^{\mathstrut}+\sigma_J^2)}  & \cdots & \sum_{k=1}^{p_J}\frac{(\mat{T}_{J:k_{}}^{\top}\mat{T}_{I:p_I}^{\mathstrut})^2}{(\mat{T}_{I:p_I}^{\top}\mat{T}_{I:p_I}^{\mathstrut}+\sigma_I^2)(\mat{T}_{J:k_{}}^{\top}\mat{T}_{J:k_{}}^{\mathstrut}+\sigma_J^2)}
\end{pmatrix}+\cO_p(n^{-1/2})\mat{1}.
\end{align*}
Hence, 
\begin{equation}\label{eq:diag-clt}
\begin{aligned}
&\mathrm{diag}\paren*{(\mat{V}_{II}^{-1}\mat{V}^{\mathstrut}_{IJ}\mat{V}_{JJ}^{-1}\mat{V}^{\mathstrut}_{JI}-\mat{\Sigma}_I)\mat{V}_{II}^{-1}}=\\
&\begin{pmatrix}
\frac{1}{n(\mat{T}_{I:1_{}}^{\top}\mat{T}_{I:1_{}}^{\mathstrut}+\sigma_I^2)}\paren*{\sum_{k=1}^{p_J}\frac{(\mat{T}_{J:k_{}}^{\top}\mat{T}_{I:1_{}}^{\mathstrut})^2}{(\mat{T}_{I:1_{}}^{\top}\mat{T}_{I:1_{}}^{\mathstrut}+\sigma_I^2)(\mat{T}_{J:k_{}}^{\top}\mat{T}_{J:k_{}}^{\mathstrut}+\sigma_J^2)}-1}\\
\vdots\\
\frac{1}{n(\mat{T}_{I:p_I}^{\top}\mat{T}_{I:p_I}^{\mathstrut}+\sigma_I^2)}\paren*{\sum_{k=1}^{p_J}\frac{(\mat{T}_{J:k_{}}^{\top}\mat{T}_{I:p_I}^{\mathstrut})(\mat{T}_{J:k_{}}^{\top}\mat{T}_{I:2_{}}^{\mathstrut})}{(\mat{T}_{I:p_I}^{\top}\mat{T}_{I:p_I}^{\mathstrut}+\sigma_I^2)(\mat{T}_{J:k}^{\top}\mat{T}_{J:k}^{\mathstrut}+\sigma_J^2)}-1}
\end{pmatrix}+\cO_p(n^{-3/2})\mat{1}.
\end{aligned}
\end{equation}
Putting together \eqref{eq:cond-varI-decomp}, \eqref{eq:deriv-var-given-IJ-raw} and \eqref{eq:diag-clt}, derivative of the aleatoric variance is 
\begin{equation}\label{eq:deriv-aleatoric}
\begin{aligned}
    \frac{d}{d\rho} V_a(\mat{V}_I; \rho)\mid_{\rho=0}
    &= \frac{d}{d\rho} \brac*{\mathrm{diag}\paren*{\Tilde{\mat{\Sigma}}_I}^{\top} \E \cbrac*{\Var^{\circ}\paren*{\hat{\mat{\theta}}_{I} \mid \mat{V}_I, \mat{V}_J} \mid \mat{V}_I}}\mid_{\rho=0}\\
    &= \mathrm{diag}\paren*{\Tilde{\mat{\Sigma}}_I}^{\top} \E \cbrac*{\frac{d}{d\rho}\Var^{\circ}\paren*{\hat{\mat{\theta}}_{I} \mid \mat{V}_I, \mat{V}_J}\mid_{\rho=0} \mid \mat{V}_I}\\
    &= \underbrace{\frac{2\bar{\sigma}^{*2}}{n}\sum_{m=1}^{p_I}\paren*{\sum_{k=1}^{p_J}\frac{(\mat{T}_{J:k}^{\top}\mat{T}_{I:m}^{\mathstrut})^2}{(\mat{T}_{I:m}^{\top}\mat{T}_{I:m}^{\mathstrut}+\sigma_I^2)(\mat{T}_{J:k}^{\top}\mat{T}_{J:k}^{\mathstrut}+\sigma_J^2)}-1}}_{\coloneqq \Xi}+\cO_p(n^{-3/2})
\end{aligned}
\end{equation}

Let $\Xi$ denote the asymptotic quantity, from \Cref{lem:xi-negative}, $\Xi <0$, which implies an increase of penalty $\rho$ around 0 is beneficial in the sense that it reduces the intrinsic variance of $\hat{\mat{\theta}}_{I}$ given $\mat{V}_I, \mat{V}_J$.

Next, we examine the epistemic variance in \eqref{eq:cond-varI-decomp}, namely, $\mathrm{diag}\paren*{\Tilde{\mat{\Sigma}}_I}^{\top} \Var^{\circ}\cbrac*{\E \paren*{\hat{\mat{\theta}}_{I}\mid \mat{V}_I, \mat{V}_J} \mid \mat{V}_I}$. Let $\bar{\mat{\theta}}^*_I$ be the constant defined in \Cref{lem:Y-given-IJ}, by definition of conditional variance,
\begin{align*}
     \Var^{\circ}\cbrac*{\E \paren{\hat{\mat{\theta}}_{I}\mid \mat{V}_I, \mat{V}_J} \mid \mat{V}_I} 
    &=\Var^{\circ}\cbrac*{\E \paren{\hat{\mat{\theta}}_{I} -\bar{\mat{\theta}}^*_I \mid \mat{V}_I, \mat{V}_J} \mid \mat{V}_I}\\
    &=\E\brac*{\cbrac*{\E \paren{\hat{\mat{\theta}}_{I} -\bar{\mat{\theta}}^*_I \mid \mat{V}_I, \mat{V}_J}}^{\circ2} \mid \mat{V}_I}-\brac*{\E\cbrac*{\E \paren{\hat{\mat{\theta}}_{I} -\bar{\mat{\theta}}^*_I \mid \mat{V}_I, \mat{V}_J}\mid \mat{V}_I}}^{\circ2}\\
    &=\E\brac*{\cbrac*{\E \paren{\hat{\mat{\theta}}_{I} -\bar{\mat{\theta}}^*_I \mid \mat{V}_I, \mat{V}_J}}^{\circ2} \mid \mat{V}_I}-\cbrac*{\E\paren{\hat{\mat{\theta}}_{I} -\bar{\mat{\theta}}^*_I \mid \mat{V}_I}}^{\circ2}.
\end{align*}
Therefore,
\begin{equation}\label{eq:deriv-cond-var} 
\begin{aligned}
     &\frac{d}{d\rho} \Var^{\circ}\cbrac*{\E \paren{\hat{\mat{\theta}}_{I}\mid \mat{V}_I, \mat{V}_J} \mid \mat{V}_I} \mid_{\rho=0}\\
     &=\frac{d}{d\rho} \E\brac*{\cbrac*{\E \paren{\hat{\mat{\theta}}_{I} -\bar{\mat{\theta}}^*_I \mid \mat{V}_I, \mat{V}_J}}^{\circ2} \mid \mat{V}_I}\mid_{\rho=0} -\frac{d}{d\rho}\cbrac*{\E\paren{\hat{\mat{\theta}}_{I} -\bar{\mat{\theta}}^*_I \mid \mat{V}_I}}^{\circ2}\mid_{\rho=0}.
\end{aligned}
\end{equation}
We begin with the first term in \eqref{eq:deriv-cond-var}, by Leibniz Integral rule, 
\begin{equation}\label{eq:deriv-exp-squared-estimator}
\begin{aligned}
    &\frac{d}{d\rho} \E\brac*{\cbrac*{\E \paren{\hat{\mat{\theta}}_{I}-\bar{\mat{\theta}}^*_I \mid \mat{V}_I, \mat{V}_J}}^{\circ2} \mid \mat{V}_I}\mid_{\rho=0} \\
    &=\E\brac*{ \frac{d}{d\rho} \cbrac*{\E \paren{\hat{\mat{\theta}}_{I}-\bar{\mat{\theta}}^*_I \mid \mat{V}_I, \mat{V}_J}}^{\circ2} \mid_{\rho=0}  \mid \mat{V}_I}\\
    &= 2\E\cbrac*{\E \paren{\hat{\mat{\theta}}_{I}-\bar{\mat{\theta}}^*_I  \mid \mat{V}_I, \mat{V}_J}\mid_{\rho=0} \circ \frac{d}{d\rho} \E \paren{\hat{\mat{\theta}}_{I}-\bar{\mat{\theta}}^*_I \mid \mat{V}_I, \mat{V}_J}\mid_{\rho=0} \mid \mat{V}_I}\\
    &= 2\E\cbrac*{\mat{V}^{-1}_{II}\mat{V}^{\mathstrut}_{IJ}\bar{\mat{\theta}}^*_J \circ  \paren*{\mat{V}^{-1}_{II} \mat{V}^{\mathstrut}_{IJ} \mat{V}^{-1}_{JJ}\mat{V}^{\mathstrut}_{JI}\bar{\mat{\theta}}^*_I-\bar{\mat{\theta}}^*_I} \mid \mat{V}_I},
\end{aligned}
\end{equation}
where the last equality follows directly from \Cref{lem:estimator-exp&deriv-given-IJ}.

Next we examine the second term in \eqref{eq:deriv-cond-var}, by \eqref{eq:exp-estimator-I}, we have
\begin{equation}\label{eq:deriv-squared-exp-estimator}
\begin{aligned}
    &\frac{d}{d\rho}\cbrac*{\E\paren{\hat{\mat{\theta}}_{I} -\bar{\mat{\theta}}^*_I \mid \mat{V}_I}}^{\circ2}\mid_{\rho=0} \\
    &= 2\cbrac*{\E\paren{\hat{\mat{\theta}}_{I} -\bar{\mat{\theta}}^*_I \mid \mat{V}_I} }\mid_{\rho=0} \circ \frac{d}{d\rho} \E\paren*{\hat{\mat{\theta}}_{I} -\bar{\mat{\theta}}^*_I  \mid \mat{V}_I} \mid_{\rho=0}\\
    &= 2\paren*{\mat{\theta}^*_{I} -\bar{\mat{\theta}}^*_I } \circ \frac{d}{d\rho} \E\paren{\hat{\mat{\theta}}_{I} -\bar{\mat{\theta}}^*_I \mid \mat{V}_I} \mid_{\rho=0}\\
    &= 2\paren*{\mat{\theta}^*_{I} -\bar{\mat{\theta}}^*_I } \circ \E\cbrac*{\frac{d}{d\rho} \E\paren{\hat{\mat{\theta}}_{I} -\bar{\mat{\theta}}^*_I \mid \mat{V}_I, \mat{V}_J} \mid_{\rho=0} \mid \mat{V}_I}\\
    &= 2\paren*{\mat{\theta}^*_{I} -\bar{\mat{\theta}}^*_I } \circ \E\paren*{\mat{V}^{-1}_{II} \mat{V}^{\mathstrut}_{IJ} \mat{V}^{-1}_{JJ}\mat{V}^{\mathstrut}_{JI}\bar{\mat{\theta}}^*_I-\bar{\mat{\theta}}^*_I \mid \mat{V}_I},
\end{aligned}
\end{equation}

Combining \eqref{eq:clt}, \eqref{eq:deriv-cond-var}, \eqref{eq:deriv-exp-squared-estimator} and \eqref{eq:deriv-squared-exp-estimator}, we have
\begin{equation}\label{eq:deriv-epstemic}
\begin{aligned}
    &\frac{d}{d\rho} V_e(\mat{V}_I; \rho) \mid_{\rho=0}\\
    &= \frac{d}{d\rho} \mathrm{diag}\paren*{\Tilde{\mat{\Sigma}}_I}^{\top} \Var^{\circ}\cbrac*{\E \paren{\hat{\mat{\theta}}_{I}\mid \mat{V}_I, \mat{V}_J}} \mid_{\rho=0}\\
    &= 2\mathrm{diag}\paren*{\Tilde{\mat{\Sigma}}_I}^{\top}\E\brac*{\cbrac*{\mat{V}^{-1}_{II}\mat{V}^{\mathstrut}_{IJ}\bar{\mat{\theta}}^*_J-\paren*{\mat{\theta}^*_{I} -\bar{\mat{\theta}}^*_I}} \circ  \paren*{\mat{V}^{-1}_{II} \mat{V}^{\mathstrut}_{IJ} \mat{V}^{-1}_{JJ}\mat{V}^{\mathstrut}_{JI}\bar{\mat{\theta}}^*_I-\bar{\mat{\theta}}^*_I} \mid \mat{V}_I}\\
    &= 2\mathrm{diag}\paren*{\Tilde{\mat{\Sigma}}_I}^{\top}\E\brac*{\cbrac*{\Tilde{\mat{\Sigma}}_I^{-1}\mat{T}^{\mathstrut}_{IJ}\bar{\mat{\theta}}^*_J-\paren*{\mat{\theta}^*_{I} -\bar{\mat{\theta}}^*_I}} \circ  \paren*{\Tilde{\mat{\Sigma}}_I^{-1} \mat{T}^{\mathstrut}_{IJ} \Tilde{\mat{\Sigma}}_J^{-1}\mat{T}^{\mathstrut}_{JI}\bar{\mat{\theta}}^*_I-\bar{\mat{\theta}}^*_I} \mid \mat{V}_I} + \cO_p(n^{-1/2})\\
    &= 0+\cO_p(n^{-1/2}),
\end{aligned}
\end{equation}
where the last equality follow directly from \Cref{lem:oracle-coeff-relation}.

In conclusion, from \eqref{eq:mse-initial-decomp} and \eqref{eq:cond-varI-decomp}, the MSE can be decomposed into 
\begin{align*}
    \text{MSE}(I;\rho) 
    = B^2(\mat{V}_I; \rho) + V_a(\mat{V}_I; \rho) + V_e(\mat{V}_I; \rho) + \sigma_I^{*2},
\end{align*}
where by \eqref{eq:deriv-bias}, \eqref{eq:deriv-aleatoric} and \eqref{eq:deriv-epstemic},
\begin{align*}
    &\frac{d}{d\rho}B^2(\mat{V}_I; \rho) \mid_{\rho=0} = 0,\\
    &\frac{d}{d\rho} V_a(\mat{V}_I; \rho)\mid_{\rho=0} = \Xi+\cO_p(n^{-3/2}),\\
    &\frac{d}{d\rho} V_e(\mat{V}_I; \rho) \mid_{\rho=0}= 0+\cO_p(n^{-1/2}).
\end{align*}
\end{proof}

\begin{proof}[Proof of \Cref{thm:gen-error-conditional}]
Let $\mat{v}_I \in \mathbb{R}^{n\times p_I}, \mat{v}_J\in \mathbb{R}^{n\times p_J}$ denote the realizations of $\mat{V}_I, \mat{V}_J$. Recall that the definition of $\cE$ is
\begin{align*}
    \mathcal{E}=\cbrac*{\mat{v}_I, \mat{v}_J \,\middle\vert\, \paren*{\mat{v}^{-1}_{II}\mat{v}^{\mathstrut}_{IJ}\bar{\mat{\theta}}^*_J-\paren*{\mat{\theta}^*_{I} -\bar{\mat{\theta}}^*_I}} \succeq \mat{0} \text{ and }  \paren*{\mat{v}^{-1}_{II} \mat{v}^{\mathstrut}_{IJ} \mat{v}^{-1}_{JJ}\mat{v}^{\mathstrut}_{JI}\bar{\mat{\theta}}^*_I-\bar{\mat{\theta}}^*_I} \preceq \mat{0}},
\end{align*}
where the notation $\mat{x} \succeq\mat{y}$ (respectively, $\mat{x} \preceq \mat{y}$) means for $\mat{x}, \mat{y}\in \mathbb{R}^d, x_i \geq y_i$ (respectively, $ x_i \leq y_i$) for all $i \in [d]$.

Next we begin to analyze the bias term under event $\cE$. From the proof of \Cref{thm:gen-error-general}, we have that
\begin{align*}
    &\frac{d}{d\rho}B^2(\mat{V}_I; \rho) \mid_{\rho=0}\\
    &=2\mathrm{diag}\paren*{\Tilde{\mat{\Sigma}}_I}^{\top} \brac*{ \E\cbrac*{ (\hat{\mat{\theta}}_{I}-\mat{\theta}^*_I) \mid \mat{V}_I}\mid_{\rho=0}  \circ \frac{d}{d\rho}\E\cbrac*{ (\hat{\mat{\theta}}_{I}-\mat{\theta}^*_I) \mid \mat{V}_I}\mid_{\rho=0} }\\
    &=2\mathrm{diag}\paren*{\Tilde{\mat{\Sigma}}_I}^{\top} \brac*{ \E\cbrac*{\E\cbrac{(\hat{\mat{\theta}}_{I}-\bar{\mat{\theta}}^*_I)-(\mat{\theta}^*_I-\bar{\mat{\theta}}^*_I)\mid \mat{V}_I, \mat{V}_J} \mid \mat{V}_I}\mid_{\rho=0} \circ \frac{d}{d\rho}\E\cbrac*{ (\hat{\mat{\theta}}_{I}-\mat{\theta}^*_I) \mid \mat{V}_I}\mid_{\rho=0} }\\
    &=2\mathrm{diag}\paren*{\Tilde{\mat{\Sigma}}_I}^{\top} \brac*{ \E\cbrac*{\mat{V}^{-1}_{II}\mat{V}^{\mathstrut}_{IJ}\bar{\mat{\theta}}^*_J-(\mat{\theta}^*_I-\bar{\mat{\theta}}^*_I)  \mid \mat{V}_I}  \circ \frac{d}{d\rho}\E\cbrac*{\E \paren{\hat{\mat{\theta}}_{I} -\mat{\theta}^*_I \mid \mat{V}_I, \mat{V}_J}\mid \mat{V}_I}\mid_{\rho=0} }\\
    &= 2\mathrm{diag}\paren*{\Tilde{\mat{\Sigma}}_I}^{\top} \brac*{ \E\cbrac*{\mat{V}^{-1}_{II}\mat{V}^{\mathstrut}_{IJ}\bar{\mat{\theta}}^*_J-(\mat{\theta}^*_I-\bar{\mat{\theta}}^*_I)  \mid \mat{V}_I}  \circ \E\paren*{\mat{V}^{-1}_{II} \mat{V}^{\mathstrut}_{IJ} \mat{V}^{-1}_{JJ}\mat{V}^{\mathstrut}_{JI}\bar{\mat{\theta}}^*_I-\bar{\mat{\theta}}^*_I\mid \mat{V}_I}}\\
    &\leq 0 \text{ under } \cE,
\end{align*}
where the last equality follows from \Cref{lem:estimator-exp&deriv-given-IJ} and the last inequality follows directly from the definition of $\cE$.
Furthermore, by \eqref{eq:deriv-epstemic} in the proof of \Cref{thm:gen-error-general}, derivative of the epistemic variance is
\begin{align*}
    \frac{d}{d\rho} V_e(\mat{V}_I; \rho) \mid_{\rho=0}&= 2\mathrm{diag}\paren*{\Tilde{\mat{\Sigma}}_I}^{\top}\E\brac*{\cbrac*{\mat{V}^{-1}_{II}\mat{V}^{\mathstrut}_{IJ}\bar{\mat{\theta}}^*_J-\paren*{\mat{\theta}^*_{I} -\bar{\mat{\theta}}^*_I}}\circ  \paren*{\mat{V}^{-1}_{II} \mat{V}^{\mathstrut}_{IJ} \mat{V}^{-1}_{JJ}\mat{V}^{\mathstrut}_{JI}\bar{\mat{\theta}}^*_I-\bar{\mat{\theta}}^*_I} \mid \mat{V}_I}\\
    &\leq 0 \text{ under } \cE,
\end{align*}
by definition of $\cE$.
\end{proof}

\begin{proof}[Proof of \Cref{cor:simplified-event}]
In the special case where $p=p_I=p_J=1$, latent representations $\mat{v}_I, \mat{v}_J \in \mathbb{R}^n$ reduces to vectors, the oracle quantities $T_I, T_J, \bar{\theta}^*_I, \bar{\theta}^*_J$ and $\theta^*_I$ becomes scalars, and the event $\cE$ in \Cref{thm:gen-error-conditional} reduces to
\begin{align*}
\mathcal{E}=\cbrac*{\mat{v}_I, \mat{v}_J \,\middle\vert\, 
 \frac{\bar{\theta}^*_J\mat{v}_I^{\top}\mat{v}_J}{\mat{v}_I^{\top}\mat{v}_I}-\paren*{\theta^*_I - \bar{\theta}^*_I} \geq 0 \text{ and }  \frac{\bar{\theta}^*_I \cbrac*{(\mat{v}_I^{\top}\mat{v}_J)^2-(\mat{v}_J^{\top}\mat{v}_J)(\mat{v}_I^{\top}\mat{v}_I)}}{(\mat{v}_J^{\top}\mat{v}_J)(\mat{v}_I^{\top}\mat{v}_I)} \leq 0},
\end{align*}

Assume without loss of generality that $T_I, T_J \geq 0$, from the definition of $\bar{\theta}^*_I, \bar{\theta}^*_J$ in \Cref{lem:Y-given-IJ} and the definition of $\theta^*_I$ in \Cref{lem:Y-given-I}, we have $\bar{\theta}^*_I, \bar{\theta}^*_J , \theta^*_I \geq 0$. 

Therefore, the second condition in $\cE$ trivially holds since $(\mat{v}_I^{\top}\mat{v}_J)^2 \leq (\mat{v}_J^{\top}\mat{v}_J)(\mat{v}_I^{\top}\mat{v}_I)$ for any $\mat{v}_I, \mat{v}_J \in \mathbb{R}^n$. Furthermore
\begin{align*}
    \theta^*_I - \bar{\theta}^*_I 
    &= \frac{\theta T_I}{T_I^2+\sigma_I^2} - \frac{\theta T_I}{\sigma_I^2 + T_I^2 + T_J^2 \sigma_I^2 /\sigma_J^2 }\\
    &= \frac{\theta T_I( T_J^2 \sigma_I^2 /\sigma_J^2)}{(T_I^2+\sigma_I^2)(\sigma_I^2 + T_I^2 + T_J^2 \sigma_I^2 /\sigma_J^2)}\\
    &=\frac{T_I T_J}{T_I^2+\sigma_I^2} \cdot \frac{\theta T_J}{\sigma_J^2 + T_J^2 + T_I^2 \sigma_J^2 /\sigma_I^2 } = \frac{T_I T_J}{T_I^2+\sigma_I^2} \bar{\theta}^*_J,
\end{align*}
therefore the first condition in $\cE$ is equivalent as
\begin{align*}
\frac{\mat{v}_I^{\top}\mat{v}_J}{\mat{v}_I^{\top}\mat{v}_I}\geq  \frac{T_I T_J}{T_I^2+\sigma_I^2}.
\end{align*}
Hence $\cE$ is equivalent as $\cbrac*{\mat{v}_I, \mat{v}_J \,\middle\vert\, \mat{v}_I^{\top}\mat{v}_J/\mat{v}_I^{\top}\mat{v}_I \geq  T_I T_J/(T_I^2+\sigma_I^2)}$.
\end{proof}

\section{Supplementary Tables and Figures}\label{app:supp-tabs-figs}

\begin{figure}[!htb]
    \centering
    \includegraphics[width=\linewidth]{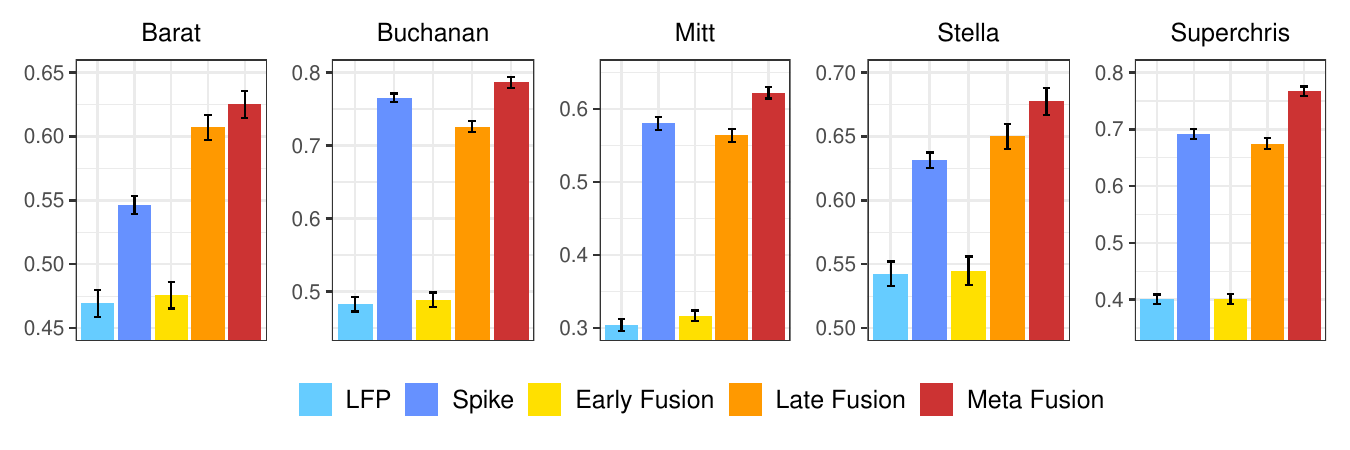}
    \caption{Classification accuracies for neural decoding across five rats using various data fusion strategies (see \Cref{sec:neuron}; these are extended results for \Cref{fig:Neuron}). Error bars indicate standard errors.}
    \label{fig:neuron_full}
\end{figure}

Here, we present additional results that extend those shown in the main manuscript. \Crefrange{tab:reg_early_fusion_app}{tab:Neuron_combined} show that our proposed Meta Fusion strategy outperforms alternative ensemble techniques. \Cref{fig:neuron_full} provides detailed results for the neural decoding task.  

\begin{table}[!htb]
    \centering
    \input{tables/regression_early_ensemble}
    \caption{Comparison of various ensemble techniques under the settings described in \Cref{sec:simu-comp-mods}. The details are the same as in \Cref{tab:reg_early_fusion}.} \label{tab:reg_early_fusion_app}
    
    \vspace{6pt}
    
    \centering
    \input{tables/regression_late_ensemble}
    \caption{Comparison of various ensemble techniques under the settings described in \Cref{sec:simu-ind-mods}. The details are the same as in \Cref{tab:reg_late_fusion}.}\label{tab:reg_late_fusion_app}

    \vspace{6pt}
    
        \centering
        \input{tables/NACC_ensemble}
        \label{tab:nacc_ensemble}
    \caption{Classification accuracies and the corresponding standard errors for Alzheimer's disease detection (\Cref{sec:nacc}) using various ensemble techniques.}
\end{table}

\begin{table}[!htb]
    
    
        \centering
        \fontsize{10}{12}\selectfont
        \input{tables/Neuron_odor_ensemble}
        \caption{Ensemble Technique Comparison}
        \label{tab:neuron_odor_ensemble}
    \caption{Classification accuracies and the corresponding standard errors for neural decoding (\Cref{sec:neuron}) across five rats using various ensemble techniques. Bold numbers indicate mean accuracy within 1 SE of the highest value across all methods. }
    \label{tab:Neuron_combined}
\end{table}

%% file: tables/regression_early_ensemble.tex
\begin{tabular}{lccc}
\toprule
 & Setting 1.1 & Setting 1.2 & Setting 1.3\\
\midrule
Best Single (ind.) & 5.51 (0.09) & 43.86 (0.69) & 53.94 (0.88)\\
Best Single & 5.33 (0.09) & 42.13 (0.68) & 53.61 (0.90)\\
Meta Fusion & \textbf{5.07 (0.08)} & \textbf{38.51 (0.64)} & \textbf{49.59 (0.84)}\\
\addlinespace
Stacking & 5.20 (0.08) & \textbf{38.73 (0.67)} & 52.99 (0.90)\\
Simple Avg. & 10.87 (0.17) & 59.57 (0.94) & 73.04 (1.18)\\
Weighted Avg. & 5.29 (0.09) & 42.72 (0.73) & 58.16 (0.98)\\
\bottomrule
\end{tabular}

%% file: tables/regression_late_ensemble.tex
\begin{tabular}{lccc}
\toprule
 & Setting 2.1 & Setting 2.2 & Setting 2.3\\
\midrule
Best Single (ind.) & 2.70 (0.06) & 60.20 (1.53) & 60.14 (1.53)\\
Best Single & 2.63 (0.06) & 57.61 (1.51) & 57.30 (1.49)\\
Meta Fusion & \textbf{2.47 (0.06)} & \textbf{52.27 (1.35)} & \textbf{53.38 (1.40)}\\
\addlinespace
Stacking & \textbf{2.51 (0.06)} & 61.11 (1.95) & 64.51 (1.72)\\
Simple Avg. & \textbf{2.48 (0.06)} & 60.72 (1.57) & 64.99 (1.66)\\
Weighted Avg. & \textbf{2.47 (0.06)} & 59.80 (1.55) & 63.47 (1.62)\\
\bottomrule
\end{tabular}

%% file: tables/NACC_ensemble.tex
\begin{tabular}[t]{lc}
\toprule
 & Mean Acc (SE)\\
\midrule
Best Single (ind.) & 0.7944 (0.0020)\\
Best Single & 0.7980 (0.0020)\\
Meta Fusion & \textbf{0.8004 (0.0019)}\\
\addlinespace
Stacking & 0.7272 (0.0050)\\
Simple Avg. & 0.7966 (0.0021)\\
Weighted Avg. & 0.7968 (0.0021)\\
\addlinespace
Majority Vote & 0.7961 (0.0020)\\
Weighted Vote & 0.7961 (0.0020)\\
\bottomrule
\end{tabular}

%% file: tables/Neuron_odor_ensemble.tex
\begin{tabular}[t]{lccccc}
\toprule
  & Barat & Buchanan & Mitt & Stella & Superchris\\
\midrule
Best Single (ind.) & 0.556 (0.010) & 0.768 (0.007) & 0.619 (0.008) & 0.632 (0.007) & \textbf{0.768 (0.008)}\\
Best Single & 0.608 (0.010) & 0.780 (0.009) & \textbf{0.630 (0.008)} & 0.649 (0.010) & \textbf{0.766 (0.009)}\\
Meta Fusion & \textbf{0.625 (0.010)} & \textbf{0.787 (0.007)} & \textbf{0.622 (0.008)} & \textbf{0.677 (0.011)} & \textbf{0.767 (0.009)}\\
\addlinespace
Stacking & 0.562 (0.012) & 0.693 (0.010) & 0.495 (0.012) & 0.591 (0.011) & 0.660 (0.013)\\
Simple Avg. & \textbf{0.621 (0.010)} & 0.774 (0.008) & 0.560 (0.009) & 0.644 (0.009) & 0.704 (0.009)\\
Weighted Avg. & \textbf{0.622 (0.010)} & \textbf{0.789 (0.007)} & 0.579 (0.008) & 0.648 (0.009) & 0.738 (0.008)\\
\addlinespace
Majority Vote & 0.601 (0.010) & 0.751 (0.008) & 0.414 (0.008) & 0.618 (0.010) & 0.510 (0.009)\\
Weighted Vote & \textbf{0.618 (0.010)} & \textbf{0.786 (0.007)} & 0.462 (0.008) & 0.628 (0.010) & 0.689 (0.010)\\
\bottomrule
\end{tabular}

%% file: ref.bib
@incollection{tulving72,
	Address = {New York},
	Author = {Tulving, E.},
	Booktitle = {Organization of Memory},
	Editor = {Tulving, Endel and Donaldson, W.},
	Pages = {381--403},
	Publisher = {Academic Press},
	Title = {Episodic and Semantic Memory},
	Year = 1972}

@article{merchant13,
	Author = {Merchant, H. and Harrington, D. L and Meck, W. H},
	Journal = {Annual review of neuroscience},
	Pages = {313--336},
	Publisher = {Annual Reviews},
	Title = {Neural basis of the perception and estimation of time},
	Volume = {36},
	Year = {2013}}

@article{eichenbaum14,
	Author = {Eichenbaum, H.B. },
	Journal = {Nature Reviews Neuroscience},
	Pages = {732- 744},
	Title = {Time cells in the hippocampus: a new dimension for mapping memories.},
	Volume = {15},
	Year = {2014}}

@book{mitzdorf1985current,
	title={Current source-density method and application in cat cerebral cortex: investigation of evoked potentials and EEG phenomena},
	author={Mitzdorf, U. and others},
	year={1985},
	publisher={American Physiological Society}
}

@article{folstein1975mini,
  title={{“Mini-mental state”}: A practical method for grading the cognitive state of patients for the clinician},
  author={Folstein, Marshal F and Folstein, Susan E and McHugh, Paul R},
  journal={Journal of Psychiatric Research},
  volume={12},
  number={3},
  pages={189--198},
  year={1975},
  publisher={Elsevier},
  doi={10.1016/0022-3956(75)90026-6}
}

@ARTICLE{Li-2020-FL,
  author={Li, Tian and Sahu, Anit Kumar and Talwalkar, Ameet and Smith, Virginia},
  journal={IEEE Signal Processing Magazine}, 
  title={Federated Learning: Challenges, Methods, and Future Directions}, 
  year={2020},
  volume={37},
  number={3},
  pages={50-60},
  doi={10.1109/MSP.2020.2975749}}

@article{Trinh2024-SE,
  title = {Alzheimer’s disease detection using data fusion with a deep supervised encoder},
  volume = {3},
  ISSN = {2813-3919},
  url = {http://dx.doi.org/10.3389/frdem.2024.1332928},
  DOI = {10.3389/frdem.2024.1332928},
  journal = {Frontiers in Dementia},
  publisher = {Frontiers Media SA},
  author = {Trinh,  Minh and Shahbaba,  Ryan and Stark,  Craig and Ren,  Yueqi},
  year = {2024},
  month = {2}
}

@article{Livingston2020-AD,
  title = {Dementia prevention, intervention, and care: 2020 report of the Lancet Commission},
  volume = {396},
  ISSN = {0140-6736},
  url = {http://dx.doi.org/10.1016/S0140-6736(20)30367-6},
  DOI = {10.1016/s0140-6736(20)30367-6},
  number = {10248},
  journal = {The Lancet},
  publisher = {Elsevier BV},
  author = {Livingston,  Gill and Huntley,  Jonathan and Sommerlad,  Andrew and Ames,  David and Ballard,  Clive and Banerjee,  Sube and Brayne,  Carol and Burns,  Alistair and Cohen-Mansfield,  Jiska and Cooper,  Claudia and Costafreda,  Sergi G and Dias,  Amit and Fox,  Nick and Gitlin,  Laura N and Howard,  Robert and Kales,  Helen C and Kivim\"{a}ki,  Mika and Larson,  Eric B and Ogunniyi,  Adesola and Orgeta,  Vasiliki and Ritchie,  Karen and Rockwood,  Kenneth and Sampson,  Elizabeth L and Samus,  Quincy and Schneider,  Lon S and Selbæk,  Geir and Teri,  Linda and Mukadam,  Naaheed},
  year = {2020},
  month = {8},
  pages = {413–446}
}

@article{Jack2010-biomarkers-AD,
  title = {Hypothetical model of dynamic biomarkers of the Alzheimer’s pathological cascade},
  volume = {9},
  ISSN = {1474-4422},
  url = {http://dx.doi.org/10.1016/S1474-4422(09)70299-6},
  DOI = {10.1016/s1474-4422(09)70299-6},
  number = {1},
  journal = {The Lancet Neurology},
  publisher = {Elsevier BV},
  author = {Jack,  Clifford R and Knopman,  David S and Jagust,  William J and Shaw,  Leslie M and Aisen,  Paul S and Weiner,  Michael W and Petersen,  Ronald C and Trojanowski,  John Q},
  year = {2010},
  month = {1},
  pages = {119–128}
}

@article{Tsoi2015-cognitivetest-AD,
  title = {Cognitive Tests to Detect Dementia: A Systematic Review and Meta-analysis},
  volume = {175},
  ISSN = {2168-6106},
  url = {http://dx.doi.org/10.1001/jamainternmed.2015.2152},
  DOI = {10.1001/jamainternmed.2015.2152},
  number = {9},
  journal = {JAMA Internal Medicine},
  publisher = {American Medical Association (AMA)},
  author = {Tsoi,  Kelvin K. F. and Chan,  Joyce Y. C. and Hirai,  Hoyee W. and Wong,  Samuel Y. S. and Kwok,  Timothy C. Y.},
  year = {2015},
  month = {9},
  pages = {1450}
}

@article{Weintraub2012-Neuropsychological-AD,
  title = {The Neuropsychological Profile of Alzheimer Disease},
  volume = {2},
  ISSN = {2157-1422},
  url = {http://dx.doi.org/10.1101/cshperspect.a006171},
  DOI = {10.1101/cshperspect.a006171},
  number = {4},
  journal = {Cold Spring Harbor Perspectives in Medicine},
  publisher = {Cold Spring Harbor Laboratory},
  author = {Weintraub,  S. and Wicklund,  A. H. and Salmon,  D. P.},
  year = {2012},
  month = {1},
  pages = {a006171–a006171}
}

@article{shahbaba-2022-neuron,
  title={Hippocampal ensembles represent sequential relationships among an extended sequence of nonspatial events},
  author={Shahbaba, Babak and Li, Lingge and Agostinelli, Forest and Saraf, Mansi and Cooper, Keiland W and Haghverdian, Derenik and Elias, Gabriel A and Baldi, Pierre and Fortin, Norbert J},
  journal={Nature communications},
  volume={13},
  number={1},
  pages={787},
  year={2022},
  publisher={Nature Publishing Group UK London}
}

@article{allen-2016-neuron,
  title={Nonspatial sequence coding in CA1 neurons},
  author={Allen, Timothy A and Salz, Daniel M and McKenzie, Sam and Fortin, Norbert J},
  journal={Journal of Neuroscience},
  volume={36},
  number={5},
  pages={1547--1563},
  year={2016},
  publisher={Soc Neuroscience}
}

@article{Besser-2018-nacc,
    author = {Besser, Lilah M and Kukull, Walter A and Teylan, Merilee A and Bigio, Eileen H and Cairns, Nigel J and Kofler, Julia K and Montine, Thomas J and Schneider, Julie A and Nelson, Peter T},
    title = {The Revised National Alzheimer’s Coordinating Center’s Neuropathology Form—Available Data and New Analyses},
    journal = {Journal of Neuropathology \& Experimental Neurology},
    volume = {77},
    number = {8},
    pages = {717-726},
    year = {2018},
    month = {06},
    issn = {0022-3069},
    doi = {10.1093/jnen/nly049},
    url = {https://doi.org/10.1093/jnen/nly049},
    eprint = {https://academic.oup.com/jnen/article-pdf/77/8/717/25154410/nly049.pdf},
}

@article{Weintraub-2018-naccv3,
  title = {Version 3 of the Alzheimer Disease Centers’ Neuropsychological Test Battery in the Uniform Data Set (UDS)},
  volume = {32},
  ISSN = {0893-0341},
  url = {http://dx.doi.org/10.1097/WAD.0000000000000223},
  DOI = {10.1097/wad.0000000000000223},
  number = {1},
  journal = {Alzheimer Disease \& Associated Disorders},
  publisher = {Ovid Technologies (Wolters Kluwer Health)},
  author = {Weintraub,  Sandra and Besser,  Lilah and Dodge,  Hiroko H. and Teylan,  Merilee and Ferris,  Steven and Goldstein,  Felicia C. and Giordani,  Bruno and Kramer,  Joel and Loewenstein,  David and Marson,  Dan and Mungas,  Dan and Salmon,  David and Welsh-Bohmer,  Kathleen and Zhou,  Xiao-Hua and Shirk,  Steven D. and Atri,  Alireza and Kukull,  Walter A. and Phelps,  Creighton and Morris,  John C.},
  year = {2018},
  month = {1},
  pages = {10–17}
}

@book{murphy-2012-ml-probabilistic,
author = {Murphy, Kevin P.},
title = {Machine Learning: A Probabilistic Perspective},
year = {2012},
isbn = {0262018020},
publisher = {The MIT Press}
}

@inproceedings{Phuong2019TowardsUK,
  title={Towards Understanding Knowledge Distillation},
  author={Mary Phuong and Christoph H. Lampert},
  booktitle={International Conference on Machine Learning},
  year={2019},
  url={https://api.semanticscholar.org/CorpusID:174800711}
}

@inproceedings{
hardt2017identity,
title={Identity Matters in Deep Learning},
author={Moritz Hardt and Tengyu Ma},
booktitle={International Conference on Learning Representations},
year={2017},
url={https://openreview.net/forum?id=ryxB0Rtxx}
}

@inproceedings{kawaguchi-poorlocalmin-NIPS2016,
 author = {Kawaguchi, Kenji},
 booktitle = {Advances in Neural Information Processing Systems},
 editor = {D. Lee and M. Sugiyama and U. Luxburg and I. Guyon and R. Garnett},
 pages = {},
 publisher = {Curran Associates, Inc.},
 title = {Deep Learning without Poor Local Minima},
 url = {https://proceedings.neurips.cc/paper_files/paper/2016/file/f2fc990265c712c49d51a18a32b39f0c-Paper.pdf},
 volume = {29},
 year = {2016}
}

@article{Saxe2014ExactST,
  title={Exact solutions to the nonlinear dynamics of learning in deep linear neural networks},
  author={Andrew M. Saxe and James L. McClelland and Surya Ganguli},
  journal={International Conference on
Learning Representations (ICLR)},
  year={2014},
}

@article{Zhang2017DeepML,
  title={Deep Mutual Learning},
  author={Ying Zhang and Tao Xiang and Timothy M. Hospedales and Huchuan Lu},
  journal={2018 IEEE/CVF Conference on Computer Vision and Pattern Recognition},
  year={2017},
  pages={4320-4328},
  url={https://api.semanticscholar.org/CorpusID:26071966}
}

@INPROCEEDINGS{caruana2006EnsembleSelection,
  author={Caruana, Rich and Munson, Art and Niculescu-Mizil, Alexandru},
  booktitle={Sixth International Conference on Data Mining (ICDM'06)}, 
  title={Getting the Most Out of Ensemble Selection}, 
  year={2006},
  volume={},
  number={},
  pages={828-833},
  doi={10.1109/ICDM.2006.76}}

@inproceedings{caruana2004EnsembleSelection,
author = {Caruana, Rich and Niculescu-Mizil, Alexandru and Crew, Geoff and Ksikes, Alex},
title = {Ensemble selection from libraries of models},
year = {2004},
isbn = {1581138385},
publisher = {Association for Computing Machinery},
address = {New York, NY, USA},
url = {https://doi.org/10.1145/1015330.1015432},
doi = {10.1145/1015330.1015432},
booktitle = {Proceedings of the Twenty-First International Conference on Machine Learning},
pages = {18},
location = {Banff, Alberta, Canada},
series = {ICML '04}
}

@InProceedings{Rashed2019lidar&camera,
author = {Rashed, Hazem and Ramzy, Mohamed and Vaquero, Victor and El Sallab, Ahmad and Sistu, Ganesh and Yogamani, Senthil},
title = {FuseMODNet: Real-Time Camera and LiDAR Based Moving Object Detection for Robust Low-Light Autonomous Driving},
booktitle = {Proceedings of the IEEE/CVF International Conference on Computer Vision (ICCV) Workshops},
month = {10},
year = {2019}
}

@INPROCEEDINGS{Jelena_2018_sensor_fusion,
  author={Kocić, Jelena and Jovičić, Nenad and Drndarević, Vujo},
  booktitle={2018 26th Telecommunications Forum (TELFOR)}, 
  title={Sensors and Sensor Fusion in Autonomous Vehicles}, 
  year={2018},
  volume={},
  number={},
  pages={420-425},
  doi={10.1109/TELFOR.2018.8612054}}

@article{Das_2023_sentiment_survery,
author = {Das, Ringki and Singh, Thoudam Doren},
title = {Multimodal Sentiment Analysis: A Survey of Methods, Trends, and Challenges},
year = {2023},
publisher = {Association for Computing Machinery},
address = {New York, NY, USA},
volume = {55},
number = {13s},
issn = {0360-0300},
url = {https://doi.org/10.1145/3586075},
doi = {10.1145/3586075},
month = {7},
articleno = {270},
numpages = {38}
}

@inproceedings{bagher-zadeh-etal-2018-multimodal,
    title = "Multimodal Language Analysis in the Wild: {CMU}-{MOSEI} Dataset and Interpretable Dynamic Fusion Graph",
    author = "Bagher Zadeh, AmirAli  and
      Liang, Paul Pu  and
      Poria, Soujanya  and
      Cambria, Erik  and
      Morency, Louis-Philippe",
    editor = "Gurevych, Iryna  and
      Miyao, Yusuke",
    booktitle = "Proceedings of the 56th Annual Meeting of the Association for Computational Linguistics (Volume 1: Long Papers)",
    month = {7},
    year = "2018",
    address = "Melbourne, Australia",
    publisher = "Association for Computational Linguistics",
    url = "https://aclanthology.org/P18-1208",
    doi = "10.18653/v1/P18-1208",
    pages = "2236--2246"
}

@article{gandhi-2023-multimodal-sentiment,
title = {Multimodal sentiment analysis: A systematic review of history, datasets, multimodal fusion methods, applications, challenges and future directions},
journal = {Information Fusion},
volume = {91},
pages = {424-444},
year = {2023},
issn = {1566-2535},
doi = {https://doi.org/10.1016/j.inffus.2022.09.025},
url = {https://www.sciencedirect.com/science/article/pii/S1566253522001634},
author = {Ankita Gandhi and Kinjal Adhvaryu and Soujanya Poria and Erik Cambria and Amir Hussain}
}

@article{Qiu-2022-multimodal-alzheimer,
  author = {Qiu,  Shangran and Miller,  Matthew I. and Joshi,  Prajakta S. and Lee,  Joyce C. and Xue,  Chonghua and Ni,  Yunruo and Wang,  Yuwei and De Anda-Duran,  Ileana and Hwang,  Phillip H. and Cramer,  Justin A. and Dwyer,  Brigid C. and Hao,  Honglin and Kaku,  Michelle C. and Kedar,  Sachin and Lee,  Peter H. and Mian,  Asim Z. and Murman,  Daniel L. and O’Shea,  Sarah and Paul,  Aaron B. and Saint-Hilaire,  Marie-Helene and Alton Sartor,  E. and Saxena,  Aneeta R. and Shih,  Ludy C. and Small,  Juan E. and Smith,  Maximilian J. and Swaminathan,  Arun and Takahashi,  Courtney E. and Taraschenko,  Olga and You,  Hui and Yuan,  Jing and Zhou,  Yan and Zhu,  Shuhan and Alosco,  Michael L. and Mez,  Jesse and Stein,  Thor D. and Poston,  Kathleen L. and Au,  Rhoda and Kolachalama,  Vijaya B.},
  title = {Multimodal deep learning for Alzheimer’s disease dementia assessment},
  volume = {13},
  ISSN = {2041-1723},
  url = {http://dx.doi.org/10.1038/s41467-022-31037-5},
  DOI = {10.1038/s41467-022-31037-5},
  number = {1},
  journal = {Nature Communications},
  publisher = {Springer Science and Business Media LLC},
  year = {2022},
  month = {6}
}

@article{zhang-2011-ad-multimodal,
title = {Multimodal classification of Alzheimer's disease and mild cognitive impairment},
journal = {NeuroImage},
volume = {55},
number = {3},
pages = {856-867},
year = {2011},
issn = {1053-8119},
doi = {https://doi.org/10.1016/j.neuroimage.2011.01.008},
url = {https://www.sciencedirect.com/science/article/pii/S1053811911000267},
author = {Daoqiang Zhang and Yaping Wang and Luping Zhou and Hong Yuan and Dinggang Shen}
}

@article{Huang2020-fusion-review,
  title = {Fusion of medical imaging and electronic health records using deep learning: a systematic review and implementation guidelines},
  volume = {3},
  ISSN = {2398-6352},
  url = {http://dx.doi.org/10.1038/s41746-020-00341-z},
  DOI = {10.1038/s41746-020-00341-z},
  number = {1},
  journal = {npj Digital Medicine},
  publisher = {Springer Science and Business Media LLC},
  author = {Huang,  Shih-Cheng and Pareek,  Anuj and Seyyedi,  Saeed and Banerjee,  Imon and Lungren,  Matthew P.},
  year = {2020},
  month = {10} 
}

@article{zhang2021-imagefusion-survey,
title = {Deep multimodal fusion for semantic image segmentation: A survey},
journal = {Image and Vision Computing},
volume = {105},
pages = {104042},
year = {2021},
issn = {0262-8856},
doi = {https://doi.org/10.1016/j.imavis.2020.104042},
url = {https://www.sciencedirect.com/science/article/pii/S0262885620301748},
author = {Yifei Zhang and Désiré Sidibé and Olivier Morel and Fabrice Mériaudeau},
}

@article{Couprie2013IndoorSS,
  title={Indoor Semantic Segmentation using depth information},
  author={Camille Couprie and Cl{\'e}ment Farabet and Laurent Najman and Yann LeCun},
  journal={arXiv: Computer Vision and Pattern Recognition},
  year={2013},
  url={https://api.semanticscholar.org/CorpusID:6681692}
}

@article{yi2022-multimodal-autunomous,
author = {Xiao, Yi and Codevilla, Felipe and Gurram, Akhil and Urfalioglu, Onay and L\'{o}pez, Antonio M.},
title = {Multimodal End-to-End Autonomous Driving},
year = {2022},
issue_date = {Jan. 2022},
publisher = {IEEE Press},
volume = {23},
number = {1},
issn = {1524-9050},
url = {https://doi.org/10.1109/TITS.2020.3013234},
doi = {10.1109/TITS.2020.3013234},
journal = {Trans. Intell. Transport. Sys.},
month = jan,
pages = {537–547},
numpages = {11}
}

@article{Qiu2018-latefusoin-alzheimer,
  title = {Fusion of deep learning models of MRI scans,  Mini–Mental State Examination,  and logical memory test enhances diagnosis of mild cognitive impairment},
  volume = {10},
  ISSN = {2352-8729},
  url = {http://dx.doi.org/10.1016/j.dadm.2018.08.013},
  DOI = {10.1016/j.dadm.2018.08.013},
  number = {1},
  journal = {Alzheimers Dement (Amst)},
  publisher = {Wiley},
  author = {Qiu,  Shangran and Chang,  Gary H. and Panagia,  Marcello and Gopal,  Deepa M. and Au,  Rhoda and Kolachalama,  Vijaya B.},
  year = {2018},
  month = {1},
  pages = {737–749}
}

@article{Reda2018-latefusion-cancer,
  title = {Deep Learning Role in Early Diagnosis of Prostate Cancer},
  volume = {17},
  ISSN = {1533-0338},
  url = {http://dx.doi.org/10.1177/1533034618775530},
  DOI = {10.1177/1533034618775530},
  journal = {Technology in Cancer Research Treatment},
  publisher = {SAGE Publications},
  author = {Reda,  Islam and Khalil,  Ashraf and Elmogy,  Mohammed and Abou El-Fetouh,  Ahmed and Shalaby,  Ahmed and Abou El-Ghar,  Mohamed and Elmaghraby,  Adel and Ghazal,  Mohammed and El-Baz,  Ayman},
  year = {2018},
  month = {1}
}

@article{Yoo2019-latefusion-lesion,
author = {Youngjin Yoo and Lisa Y. W. Tang and David K. B. Li and Luanne Metz and Shannon Kolind and Anthony L. Traboulsee and Roger C. Tam},

title = {Deep learning of brain lesion patterns and user-defined clinical and MRI features for predicting conversion to multiple sclerosis from clinically isolated syndrome},
journal = {Computer Methods in Biomechanics and Biomedical Engineering: Imaging \& Visualization},
volume = {7},
number = {3},
pages = {250--259},
year = {2019},
publisher = {Taylor \& Francis},
doi = {10.1080/21681163.2017.1356750},
URL = {https://doi.org/10.1080/21681163.2017.1356750},
}

@article{ding2022-coop,
author = {Daisy Yi Ding  and Shuangning Li  and Balasubramanian Narasimhan  and Robert Tibshirani },
title = {Cooperative learning for multiview analysis},
journal = {Proceedings of the National Academy of Sciences},
volume = {119},
number = {38},
pages = {e2202113119},
year = {2022},
doi = {10.1073/pnas.2202113119},
URL = {https://www.pnas.org/doi/abs/10.1073/pnas.2202113119},
eprint = {https://www.pnas.org/doi/pdf/10.1073/pnas.2202113119}}

@article{Yala2019-jointfusion-breastcancer,
  title = {A Deep Learning Mammography-based Model for Improved Breast Cancer                    Risk Prediction},
  volume = {292},
  ISSN = {1527-1315},
  url = {http://dx.doi.org/10.1148/radiol.2019182716},
  DOI = {10.1148/radiol.2019182716},
  number = {1},
  journal = {Radiology},
  publisher = {Radiological Society of North America (RSNA)},
  author = {Yala,  Adam and Lehman,  Constance and Schuster,  Tal and Portnoi,  Tally and Barzilay,  Regina},
  year = {2019},
  month = {7},
  pages = {60–66}
}

@article{Nie2019-jointfusion-braintumor,
  title = {Multi-Channel 3D Deep Feature Learning for Survival Time Prediction of Brain Tumor Patients Using Multi-Modal Neuroimages},
  volume = {9},
  ISSN = {2045-2322},
  url = {http://dx.doi.org/10.1038/s41598-018-37387-9},
  DOI = {10.1038/s41598-018-37387-9},
  number = {1},
  journal = {Scientific Reports},
  publisher = {Springer Science and Business Media LLC},
  author = {Nie,  Dong and Lu,  Junfeng and Zhang,  Han and Adeli,  Ehsan and Wang,  Jun and Yu,  Zhengda and Liu,  LuYan and Wang,  Qian and Wu,  Jinsong and Shen,  Dinggang},
  year = {2019},
  month = {1}
}

@inproceedings{Tan2022-jointfusion-socialmedia,
author = {Tan, YunPeng and Liu, Fangyu and Li, BoWei and Zhang, Zheng and Zhang, Bo},
title = {An Efficient Multi-View Multimodal Data Processing Framework for Social Media Popularity Prediction},
year = {2022},
isbn = {9781450392037},
publisher = {Association for Computing Machinery},
address = {New York, NY, USA},
url = {https://doi.org/10.1145/3503161.3551607},
doi = {10.1145/3503161.3551607},
booktitle = {Proceedings of the 30th ACM International Conference on Multimedia},
pages = {7200–7204},
numpages = {5},
location = {Lisboa, Portugal},
series = {MM '22}
}

@inproceedings{SrivastavaS2012_multimodal-DBM,
 author = {Srivastava, Nitish and Salakhutdinov, Russ R},
 booktitle = {Advances in Neural Information Processing Systems},
 editor = {F. Pereira and C.J. Burges and L. Bottou and K.Q. Weinberger},
 pages = {},
 publisher = {Curran Associates, Inc.},
 title = {Multimodal Learning with Deep Boltzmann Machines},
 url = {https://proceedings.neurips.cc/paper_files/paper/2012/file/af21d0c97db2e27e13572cbf59eb343d-Paper.pdf},
 volume = {25},
 year = {2012}
}

@inproceedings{sutter2024-unity,
title={Unity by Diversity: Improved Representation Learning in Multimodal {VAE}s},
author={Thomas M. Sutter and Yang Meng and Norbert Fortin and Julia E Vogt and Babak Shahbaba and Stephan Mandt},
booktitle={Sixth Symposium on Advances in Approximate Bayesian Inference - Non Archival Track},
year={2024},
url={https://openreview.net/forum?id=ZgGIM0MXWA}
}

@inproceedings{Wu2018-multimodalVAE,
author = {Wu, Mike and Goodman, Noah},
title = {Multimodal generative models for scalable weakly-supervised learning},
year = {2018},
publisher = {Curran Associates Inc.},
address = {Red Hook, NY, USA},
pages = {5580–5590},
numpages = {11},
location = {Montr\'{e}al, Canada},
series = {NIPS'18}
}

@article{tibshirani1996-lasso,
 ISSN = {00359246},
 URL = {http://www.jstor.org/stable/2346178},
 author = {Robert Tibshirani},
 journal = {Journal of the Royal Statistical Society. Series B (Methodological)},
 number = {1},
 pages = {267--288},
 publisher = {[Royal Statistical Society, Oxford University Press]},
 title = {Regression Shrinkage and Selection via the Lasso},
 volume = {58},
 year = {1996}
}

@article{Breiman1996-bagging,
  title = {Bagging predictors},
  volume = {24},
  ISSN = {1573-0565},
  url = {http://dx.doi.org/10.1007/BF00058655},
  DOI = {10.1007/bf00058655},
  number = {2},
  journal = {Machine Learning},
  publisher = {Springer Science and Business Media LLC},
  author = {Breiman,  Leo},
  year = {1996},
  month = {8},
  pages = {123–140}
}

@article{He2015DeepRL,
  title={Deep Residual Learning for Image Recognition},
  author={Kaiming He and X. Zhang and Shaoqing Ren and Jian Sun},
  journal={2016 IEEE Conference on Computer Vision and Pattern Recognition (CVPR)},
  year={2015},
  pages={770-778},
  url={https://api.semanticscholar.org/CorpusID:206594692}
}

@inproceedings{Devlin2019BERTPO,
  title={BERT: Pre-training of Deep Bidirectional Transformers for Language Understanding},
  author={Jacob Devlin and Ming-Wei Chang and Kenton Lee and Kristina Toutanova},
  booktitle={North American Chapter of the Association for Computational Linguistics},
  year={2019},
  url={https://api.semanticscholar.org/CorpusID:52967399}
}

@article{wood2023diversity,
author = {Wood, Danny and Mu, Tingting and Webb, Andrew M. and Reeve, Henry W. J. and Luj\'{a}n, Mikel and Brown, Gavin},
title = {A unified theory of diversity in ensemble learning},
year = {2024},
issue_date = {January 2023},
publisher = {JMLR.org},
volume = {24},
number = {1},
issn = {1532-4435},
journal = {J. Mach. Learn. Res.},
month = {3},
articleno = {359},
numpages = {49}
}

@article{kumar2023fsp-ensemble,
title = {A review of feature set partitioning methods for multi-view ensemble learning},
journal = {Information Fusion},
volume = {100},
pages = {101959},
year = {2023},
issn = {1566-2535},
doi = {https://doi.org/10.1016/j.inffus.2023.101959},
url = {https://www.sciencedirect.com/science/article/pii/S1566253523002750},
author = {Aditya Kumar and Jainath Yadav},
}

@article{Hinton2015DistillingTK,
  title={Distilling the Knowledge in a Neural Network},
  author={Geoffrey E. Hinton and Oriol Vinyals and Jeffrey Dean},
  journal={ArXiv},
  year={2015},
  volume={abs/1503.02531},
}

@inproceedings{bucilua2006model-compression,
author = {Bucilu\u{a}, Cristian and Caruana, Rich and Niculescu-Mizil, Alexandru},
title = {Model compression},
year = {2006},
isbn = {1595933395},
publisher = {Association for Computing Machinery},
address = {New York, NY, USA},
url = {https://doi.org/10.1145/1150402.1150464},
doi = {10.1145/1150402.1150464},
booktitle = {Proceedings of the 12th ACM SIGKDD International Conference on Knowledge Discovery and Data Mining},
pages = {535–541},
numpages = {7},
location = {Philadelphia, PA, USA},
series = {KDD '06}
}

@article{rousseeuw1987silhouettes,
title = {Silhouettes: A graphical aid to the interpretation and validation of cluster analysis},
journal = {Journal of Computational and Applied Mathematics},
volume = {20},
pages = {53-65},
year = {1987},
issn = {0377-0427},
doi = {https://doi.org/10.1016/0377-0427(87)90125-7},
url = {https://www.sciencedirect.com/science/article/pii/0377042787901257},
author = {Peter J. Rousseeuw},
}

@INPROCEEDINGS{marutho2018elbow,
  author={Marutho, Dhendra and Hendra Handaka, Sunarna and Wijaya, Ekaprana and Muljono},
  booktitle={2018 International Seminar on Application for Technology of Information and Communication}, 
  title={The Determination of Cluster Number at k-Mean Using Elbow Method and Purity Evaluation on Headline News}, 
  year={2018},
  volume={},
  number={},
  pages={533-538},
  doi={10.1109/ISEMANTIC.2018.8549751}}

@article{li-2024-ml-multimodal-recommender,
title = {Graph neural networks with deep mutual learning for designing multi-modal recommendation systems},
journal = {Information Sciences},
volume = {654},
pages = {119815},
year = {2024},
issn = {0020-0255},
doi = {https://doi.org/10.1016/j.ins.2023.119815},
url = {https://www.sciencedirect.com/science/article/pii/S0020025523014007},
author = {Jianing Li and Chaoqun Yang and Guanhua Ye and Quoc Viet Hung Nguyen},
}

@inproceedings{zhang-2021-ml-image-segmentation,
author = {Zhang, Yao and Yang, Jiawei and Tian, Jiang and Shi, Zhongchao and Zhong, Cheng and Zhang, Yang and He, Zhiqiang},
title = {Modality-Aware Mutual Learning for Multi-modal Medical Image Segmentation},
year = {2021},
isbn = {978-3-030-87192-5},
publisher = {Springer-Verlag},
address = {Berlin, Heidelberg},
url = {https://doi.org/10.1007/978-3-030-87193-2_56},
doi = {10.1007/978-3-030-87193-2_56},
booktitle = {Medical Image Computing and Computer Assisted Intervention – MICCAI 2021: 24th International Conference, Strasbourg, France, September 27–October 1, 2021, Proceedings, Part I},
pages = {589–599},
numpages = {11},
location = {Strasbourg, France}
}

@article{zhang2023-contrastive-ml-image,
title = {Multi-modal contrastive mutual learning and pseudo-label re-learning for semi-supervised medical image segmentation},
journal = {Medical Image Analysis},
volume = {83},
pages = {102656},
year = {2023},
issn = {1361-8415},
doi = {https://doi.org/10.1016/j.media.2022.102656},
url = {https://www.sciencedirect.com/science/article/pii/S1361841522002845},
author = {Shuo Zhang and Jiaojiao Zhang and Biao Tian and Thomas Lukasiewicz and Zhenghua Xu},
}

@ARTICLE{wang-2022-amnet,
  author={Wang, Jinping and Li, Jun and Shi, Yanli and Lai, Jianhuang and Tan, Xiaojun},
  journal={IEEE Transactions on Circuits and Systems for Video Technology}, 
  title={AM³Net: Adaptive Mutual-Learning-Based Multimodal Data Fusion Network}, 
  year={2022},
  volume={32},
  number={8},
  pages={5411-5426},
  doi={10.1109/TCSVT.2022.3148257}}

@article{Lu-2002-inverse-block-matrix,
title = {Inverses of 2 × 2 block matrices},
journal = {Computers \& Mathematics with Applications},
volume = {43},
number = {1},
pages = {119-129},
year = {2002},
issn = {0898-1221},
doi = {https://doi.org/10.1016/S0898-1221(01)00278-4},
url = {https://www.sciencedirect.com/science/article/pii/S0898122101002784},
author = {Tzon-Tzer Lu and Sheng-Hua Shiou}
}
